%% file: neurips_2026.tex
\documentclass{article}

    \PassOptionsToPackage{numbers, compress}{natbib}

\usepackage[main, final]{neurips_2026}

\usepackage[utf8]{inputenc}%
\usepackage[T1]{fontenc}%
\usepackage{hyperref}%
\usepackage{url}%
\usepackage{booktabs}%
\usepackage{amsfonts}%
\usepackage{nicefrac}%
\usepackage{microtype}%
\usepackage[dvipsnames]{xcolor}
\usepackage{graphicx}
\usepackage{amsthm}
\usepackage{amsmath}
\newtheorem{example}{Example}
\newtheorem{proposition}{Proposition}
\newtheorem{theorem}{Theorem}
\newtheorem{lemma}{Lemma}
\newtheorem{corollary}{Corollary}
\newcommand{\medcirc}{\raisebox{1pt}{\scalebox{0.8}{$ \bigcirc$}}}
\usepackage{appendix}
\usepackage{algorithm}
\usepackage{algpseudocode}
\usepackage{amssymb}
\usepackage{subcaption}
\usepackage{threeparttable}
\usepackage[table]{xcolor}
\usepackage{colortbl}
\usepackage{bbding}
\usepackage{wrapfig}
\usepackage{bm}
\usepackage{multirow}
\usepackage{paralist}
\usepackage{enumitem}
\usepackage{needspace}

\newcommand{\bfX}{\mathbf{X}}
\newcommand{\bfY}{\mathbf{Y}}

\newcommand{\bfU}{\mathbf{U}}
\newcommand{\bfV}{\mathbf{V}}

\newcommand{\bfx}{\mathbf{x}}
\newcommand{\bfz}{\mathbf{z}}
\newcommand{\bfy}{\mathbf{y}}
\newcommand{\bfg}{\mathbf{g}}

\newcommand{\bfa}{\mathbf{a}}
\newcommand{\bfu}{\mathbf{u}}

\newcommand{\know}{\Pi}%
\newcommand{\layer}{\textsc{Casper}}
\newcommand{\argmin}{\arg\!\min}
\newcommand{\argmax}{\arg\!\max}

\newcommand{\twotick}{\textcolor{ForestGreen}\Checkmark\kern-0.73em\textcolor{ForestGreen}\Checkmark}
\newcommand{\tick}{\textcolor{ForestGreen}\Checkmark}
\newcommand{\redx}{\textcolor{BrickRed}\XSolidBrush}
\newcommand\msmall[1]{\mbox{\scriptsize\ensuremath{#1}}}

\newcommand{\clX}{\mathcal{X}}
\newcommand{\clY}{\mathcal{Y}}

\newcommand{\clO}{\mathcal{O}}
\newcommand{\clS}{\mathcal{S}}
\newcommand{\clB}{\mathcal{B}}
\newcommand{\clC}{\mathcal{C}}
\newcommand{\clP}{\mathcal{P}}
\newcommand{\clA}{\mathcal{A}}
\newcommand{\clJ}{\mathcal{J}}
\newcommand{\clL}{\mathcal{L}}
\newcommand{\clM}{\mathcal{M}}
\newcommand{\clN}{\mathcal{N}}
\newcommand{\clZ}{\mathcal{Z}}
\newcommand{\clF}{\mathcal{F}}

\newcommand{\clQ}{\mathcal{Q}}
\newcommand{\clR}{\mathcal{R}}
\newcommand{\clG}{\mathcal{G}}

    \definecolor{exactshade}{RGB}{252, 228, 230}%
    \definecolor{approxshade}{RGB}{223, 237, 247}%
    \definecolor{ablationshade}{RGB}{232, 251, 232}%

\title{\layer: A Projection-Based Neurosymbolic Layer for Scalable \& Guaranteed Constraint Satisfaction}

\author{
  Lohith Konathala \\
  Imperial College London \\
  \texttt{l.konathala25@imperial.ac.uk} \\
  \And
  Luca Andolfi \\
  Imperial College London \\
  \texttt{l.andolfi24@imperial.ac.uk} \\
  \And
  Eleonora Giunchiglia \\
  Imperial College London \\
  \texttt{e.giunchiglia@imperial.ac.uk} \\
}

\begin{document}

\maketitle

\begin{abstract}
Many real-world problems demand exact constraint satisfaction to be meaningfully
solved, yet existing neurosymbolic methods face a stark trade-off: those that
guarantee satisfaction fail to scale, while those that scale must resort to
approximation. Meanwhile, traditional deep learning approaches offer no satisfaction
guarantees at all. In this paper, we propose \layer{}, a neurosymbolic layer that simultaneously
(i) guarantees constraint satisfaction, (ii) scales at inference time, (iii) is fully automated, requiring no problem-specific
engineering, and (iv) is fully differentiable, making it composable with any neural predictor both at training and at inference time. \layer{} casts constraint satisfaction as a closed-form Euclidean
projection onto the constraint-feasible region, computable in a single forward
pass. Experiments across MNIST arithmetic (up to 1024 digits), Sudoku solving, and Warcraft pathfinding show that \layer{} is the only method that produces guaranteed-valid outputs at scale: on 30×30 Warcraft grids, exact baselines time out and approximate ones produce valid paths less than 7\% of the time, while \layer{} produces them 100\% of the time.
\end{abstract}

\section{Introduction}\label{sec:intro}

Many structured prediction problems require outputs that are guaranteed to satisfy logical constraints. A clinical decision support system recommending combinations of medications must never prescribe pairs with known contraindications; a hierarchical multi-label classifier must respect that any predicted sub-class implies its parent. In such settings, a single constraint violation invalidates the entire prediction, regardless of how accurate the underlying predictor is on average. Standard neural architectures provide no guarantee that individual predictions satisfy the constraints at inference time.

Neurosymbolic learning addresses this gap by combining neural predictors with symbolic representations of background knowledge, but existing methods exhibit a stark trade-off between guarantee and scalability. Methods that enforce constraints by construction --- Semantic 
Probabilistic Layers~\citep{ahmed2022semantic} and 
CCN+~\citep{giunchiglia2024ccn+} --- either compile the constraint 
into a circuit and renormalize the predictor against it for every 
input (SPL), or encode each $h$-valued categorical variable as $h$ 
binary indicators (CCN+). Neither approach scales to the structured-prediction 
regimes our experiments target.
Methods that scale---Scallop~\citep{huang2021scallop}, DeepSoftLog~\citep{maene2023soft}, A-NeSI~\citep{van2023nesi}, and CTSketch~\citep{choi2025ctsketch}---replace exact symbolic inference with approximations such as top-$k$ proof enumeration, neural surrogates for weighted model counting, or low-rank tensor sketches, but in doing so forfeit the satisfaction guarantee.

We present \layer{} (\textbf{C}ategorical \textbf{A}rgmax \textbf{S}atisfaction via \textbf{P}roj\textbf{E}ction and \textbf{R}esolution) \footnote{Code is available at \url{https://github.com/lkonathala25/casper}.}, a differentiable neurosymbolic layer that enforces logical constraint satisfaction at scale. Given a neural predictor $m_\theta$ and a set of constraints $\Pi$ over categorical variables, \layer{} composes with $m_\theta$ as $\tilde m_\theta = \layer \circ m_\theta$, and processes the variables in sequence projecting each 
unconstrained output onto the constraint-feasible region of the 
probability simplex. The variable assignments induced by the argmax of $\tilde m_\theta$ are guaranteed to satisfy $\Pi$ by construction. \layer{} achieves scalability through two ideas: an offline compilation step that uses the signed resolution rule to reduce multi-variable satisfaction to a sequence of single-variable problems, and a closed-form surrogate of the per-variable projection that requires quadratically fewer sequential steps on a GPU than the exact solution while preserving the satisfaction guarantee.  To further exploit GPU parallelism, \layer{} parallelizes the computation over different variables whenever possible using the same intuition as in~\cite{giunchiglia2024ccn+}, making the fully sequential case 
the worst-case bound. Table~\ref{tab:methods_comparison} summarises the contrast: \layer{} is the only method that simultaneously (i) guarantees constraint satisfaction, (ii) scales at inference time, and (iii) requires no problem-specific engineering beyond the constraint specification.

Our experiments confirm these claims across three benchmarks. On Warcraft pathfinding, \layer{} is the only method that produces guaranteed valid paths on the $30 \times 30$ grid, where exact baselines time out and approximate methods produce invalid predictions. On MNIST-Arithmetic, \layer{} is the only exact-satisfaction method that scales to sums over more than a thousand digits, with inference time that closes the gap to approximate methods as the problem grows. On Sudoku solving, \layer{} reaches $66\%$ post-layer board accuracy, against $25\%$ for the strongest scalable baseline unconstrained.

\begin{table}[t]
    \centering
    \caption{Comparison of \layer{} against baselines on three properties.
    \emph{Guaranteed Satisfaction} (Sat): method provably produces constraint-compliant
    outputs. \emph{Scalable} (Scale): method manages to scale over most over the tasks in our experimental analysis, \emph{Automatic} (Auto): no problem-specific
    engineering required.}%
    \setlength{\tabcolsep}{4pt}
    \begin{tabular}{lccccccc}
    \toprule
    Method & SPL~\citep{ahmed2022semantic} & CCN+~\citep{giunchiglia2024ccn+} & DSL~\citep{maene2023soft} & Scallop~\citep{huang2021scallop} & A-NeSI~\citep{van2023nesi} & CTSketch~\citep{choi2025ctsketch} & \layer \\
    \midrule
    Sat & \tick & \tick & \redx & \redx & \redx & \redx & \tick \\
    Scale               & \redx & \redx & \tick & \tick & \tick & \tick & \tick \\
    Auto               & \tick & \tick & \tick & \tick & \tick & \redx & \tick \\
    \bottomrule
    \end{tabular}
    \label{tab:methods_comparison}
\end{table}

\section{Problem Definition}\label{Sec:problem_def}

\textbf{Notation.}
We denote scalar constants by lower-case letters, e.g., $x$, random variables by upper-case letters, e.g., $X$, and ordered sets of constants and random variables by bold letters, e.g., $\bfx$ and $\bfX$, respectively. We write $[n]$ for the set $\{1,\ldots,n\}$.

\textbf{Setting.}
We consider a learning setting with input space $\clX$ and structured output space $\clY = [h_1] \times [h_2] \times \cdots \times [h_n]$, i.e., the structured output is given by $n$ categorical variables
$\bfY = (Y_1,\ldots,Y_n),$
where each $Y_i$ takes values in $[h_i]$.
Let $\know$ be a set of constraints defining the {\sl feasible assignments} $\clY_\know = \{\bfa \in \clY \mid \bfa \models \know\}$ of the problem, $\bfa \models \Pi$ indicating that the assignment $\bfa \in \clY$ satisfies $\know$.
For ease of exposition, we represent background knowledge using signed clauses with empty order~\citep{beckert2000_signed_CNF}. A {\sl signed clause} is an expression of the form: 
\begin{equation}\label{eq:signed_clause}
    \clS_1 : Y_1 \vee \clS_2 : Y_2 \vee \ldots \vee \clS_m : Y_m,
\end{equation}
where, with a slight abuse of notation, $Y_i$ indicates a multiple-valued logic variable (corresponding to the respective random variable), $\clS_i \subseteq [h_i]$ represents a set of truth values, and $\clS_i : Y_i$ is a {\sl signed literal}.%
Without loss of generality, from now on we assume that every variable appears at most once in every clause, as any clause of form $\clS_1 : Y_i \lor \ldots \lor \clS_k : Y_i \lor \varphi$ can always be rewritten as $(\clS_1 \cup \ldots \cup \clS_k): Y_i \lor \varphi$.
An assignment $\bfa$ {\sl satisfies} the signed clause in Equation~ \ref{eq:signed_clause} iff for some $i = 1, \ldots, m$ it assigns a value in $\clS_i$ to $Y_i$. An assignment $\bfa$ {\sl satisfies} a conjunction (or a set) of signed clauses if it satisfies all of them. Notice that this is purely a presentational choice: in our setting, signed clauses are in one-to-one correspondence with propositional logic formulas~\citep{Beckert1999_equivalence_propo&signed}. We adopt this representation only because it yields cleaner notation and simplifies the theoretical development in the remainder of the paper.

\begin{example}\label{example:problem_specification}
    Suppose we have an image classification task where we have to decide (i) if the shape shown in the image is a square, circle or triangle, (ii) if it is yellow, red or blue and (iii) if the shape is big ($\uparrow$) or small ($\downarrow$). Suppose also we know that shapes that are either big or yellow are always circles. In this problem, our structured output space will be: 
    \begin{equation*}
    \clY = \clY_1 \times \clY_2 \times \clY_3 \qquad \clY_1 = \{{\color{yellow}\blacksquare}, {\color{red}\blacksquare}, {\color{blue}\blacksquare}\} \qquad \clY_2 = \{\medcirc, \square, \triangle\}   \qquad \clY_3 = \{\downarrow, \uparrow\}
\end{equation*}
    and our knowledge can be expressed using two signed clauses:
    \begin{equation*}
   \{\medcirc\} : Y_2 \vee \{\downarrow\}: Y_3 \qquad\qquad  \{\medcirc\} : Y_2 \vee \{ {\color{red}\blacksquare}, {\color{blue}\blacksquare} \} : Y_1
    \end{equation*}
\end{example}

We study models of the form $m_\theta : \clX \to \prod_{i=1}^n \Delta([h_i]),$
where $\Delta([h_i])$ denotes the probability simplex over $[h_i]$. For any input $\bfx \in \clX$, we write
$m_\theta(\bfx) = \big(m_\theta^{1}(\bfx), \ldots, m_\theta^{n}(\bfx)\big),$
where $m_\theta^{i}(\bfx) \in \Delta([h_i])$ is the predicted categorical distribution for the $i$-th output variable, and we say that $\bfa \in \clY$ is the assignment induced by the output $m_\theta(\bfx)$ if for every $i = 1, \ldots n$, $a_i =  \argmax_{j \in [h_i]} m_\theta^{i}(\bfx)_j$.\footnote{In presence of multiple identical maximal (resp. minimal) values, we assume that $\argmax$ (resp. ($\argmin$)) deterministically returns a single index, with both operators following the same fixed tie-breaking policy.} We say that $m_\theta$ {\sl is compliant with } $\know$ if, for every $\bfx \in \clX$, the assignment $\bfa \in \clY$ induced by $m_\theta(\bfx)$
satisfies $\know$.

\textbf{Goal.} Given any network $m_\theta: \clX \to \prod_{i=1}^n \Delta([h_i])$ and constraints $\know$, our goal is to design a layer (i) that is guaranteed to be compliant with $\know$, (ii) that is scalable, and (iii) requires no problem-specific engineering beyond
the constraint specification.

\section{\layer}\label{sec:casper}

Since \layer{} reduces any multi-variable problem to a sequence of single-variable ones via an offline compilation step, we present the single-variable case first and then the general case.

\subsection{Single Variable Case}\label{subsec:single_variable}

Let $\clY = [h]$ and suppose that background knowledge trivially expresses that, for every input $\bfx$, the assignment $\bfa \in [h]$ induced by the output $m_\theta(\bfx)$ must always belong to the subset $\clS \subseteq [h]$, i.e., $\know = \{\clS : Y\}$.  

Given a model $m_\theta: \clX \to \Delta([h])$ and an input $\bfx$, let $\hat \bfy = m_\theta(\bfx)$. To ensure that the final predictor $\tilde m_\theta = \layer \circ m_\theta$ is compliant with $\know$, $\layer$ must map $\hat\bfy$ to the closest point in the simplex whose argmax lies in $\clS$. We measure closeness in squared Euclidean distance: as we show in the rest of this section, this choice yields a closed-form fixed-point characterization of the projection (Equation~\eqref{eq:fixed_point_system}), an exact decomposition of its surrogate error (Theorem~\ref{thm:exact_error}), and a principled gradient analysis (Proposition~\ref{prop:gradient_attenuation}). Concretely, we would like to solve:
\begin{equation}\label{eq:optimization_problem}
    {\argmin}_{\bfz \in \Delta([h])} \|\bfz - \hat\bfy\|_2^2 \qquad \text{ subject to } 
        \arg\!\max_{j \in [h]}  \bfz \in \clS.
\end{equation}
 If $|\clS| =1$, then the optimization problem is convex, while if $|\clS| > 1$ the feasible region can be decomposed in $|\clS|$ convex regions, each corresponding to a candidate maximizer $t\in \clS$. To ensure the feasible set is closed and the projection is well-defined, we find a {\sl $\epsilon$-margin solution} to the problem, which for every $t \in \clS$, ensures $z_t \ge z_j + \epsilon, \forall j \in [h]\setminus t$. 
 The $\epsilon$-margin solution for each of the convex problems is characterized by the fixed-point system described in Equation~\eqref{eq:fixed_point_system}. 

 \begin{theorem} [Fixed point solution]\label{prop:closed-form update}
     Let $\hat \bfy \in \Delta([h])$ and $\clS \subseteq [h]$, the $\epsilon$-margin solution to the optimization problem in Equation~\eqref{eq:optimization_problem} is equal to $\bfz^{t^\star}$ with $t^\star$ equal to:
     \begin{equation}
    t^\star = \argmin_{t\in \clS}\|\bfz^t - \hat \bfy\|_2^2,   
\end{equation}
where for every $t \in \clS$, $\bfz^t$ is the solution of the convex minimization problem obtained by fixing $t$ as the unique maximizer and $\epsilon > 0$ as an arbitrarily small constant:
\begin{equation}\label{eq:fixed_point_system}
    z^t_j = \begin{cases}
        \hat y_j & \text{if } j \not\in (\clM^t \cup \{t\})\\
        z_t^t - \epsilon & \text{if }  j \in \clM^t \qquad \qquad \quad \clM^t = \{j \in [h] \setminus \{t\} : \hat y_j \ge z^t_t - \epsilon\} \\
        \frac{
\hat y_t + \sum_{j\in \mathcal M^t} \hat y_j + |\mathcal M^t|\epsilon
}{
1+|\mathcal M^{t}|} & \text{if } j = t
    \end{cases}
\end{equation}
 \end{theorem}
The proof can be found in Appendix~\ref{app:proof_minimization_problem_solution_fixed_point}.
The solution to the fixed-point system~\eqref{eq:fixed_point_system} can be computed in
$\mathcal{O}(h \log h)$ sequential operations and in $\mathcal{O}(\log^2 h)$ GPU operations via parallel reductions (see bottom of Appendix~\ref{app:proof_minimization_problem_solution_fixed_point}). To maximise the scalability of our approach, we work with an over-approximation of $\clM^t$ that
is available in closed form directly from $\hat{\mathbf{y}}$. We define
\begin{equation}\label{eq:A_def}
\mathcal{A}^t \;=\; \bigl\{ j \in [h] \setminus \{t\} \,:\, \hat y_j \geq \hat y_t - \epsilon \bigr\},
\end{equation}
which is computable in $\mathcal{O}(h)$ time and requires a single thresholding operation on a GPU.
\begin{lemma}\label{lem:overapproximation}
For every $t \in [h]$ and every $\hat{\mathbf{y}} \in \Delta([h])$, $\mathcal{M}^t \subseteq \mathcal{A}^t.$
\end{lemma}
Proof can be found in Appendix~\ref{app:proof_overapproximation_lemma}. Intuitively, this entails that by replacing $\mathcal{M}^t$ with $\mathcal{A}^t$ in Equation~\eqref{eq:fixed_point_system} we obtain a {\sl surrogate projection} $\tilde\bfz^t$ in which the $t$th element is guaranteed to be the maximum, but for which we might have over-corrected some other elements.
\begin{theorem}[Constraint satisfaction]\label{th:constraint_satisfaction_basic_case}
        Let $m_\theta:\clX \to \Delta([h])$, $\bfx \in \clX$ and $\hat \bfy = m_\theta(\bfx)$. Let $\Pi = \{\clS : Y\}$ for some $\emptyset \subset \clS \subseteq [h]$. If 
    \begin{equation}\label{eq:closed_form_solution_approximated}
        \layer(\hat \bfy) = \tilde \bfz^{t^\star}, \qquad t^\star = \argmin_{t\in \clS}\|\tilde \bfz^t - \hat \bfy\|_2^2, 
    \end{equation}
    where for every $t \in \clS$
    \begin{equation}
        \tilde z^t_j = \begin{cases}
        \hat y_j & \text{if } j \not\in (\clA^t \cup \{t\})\\
        \tilde z_t^t  - \epsilon & \text{if }  j \in \clA^t \qquad \qquad \quad \clA^t = \{j \in [h] \setminus \{t\} : \hat y_j \ge \hat y_t - \epsilon\}, \\
        \frac{
\hat y_t + \sum_{j\in \mathcal A^t} \hat y_j + |\mathcal A^t|\epsilon
}{
1+|\mathcal A^{t}|} & \text{if } j = t
\end{cases}
    \end{equation}
then for $\epsilon \to 0^+$, the assignment $\bfa$ induced by $\layer(\hat \bfy)$ satisfies $\Pi$, i.e., $\bfa \models \Pi$.
\end{theorem}
Proof can be found in Appendix~\ref{app:proof_constraint_satisfaction_basic_case}. 
Thus, for every $t \in \clS$, \layer{} uses the surrogate projection
$\tilde\bfz^t$ instead of the exact fixed-point solution $\bfz^t$. The surrogate avoids the sorting step required by the exact solution, reducing the number of sequential GPU operations per candidate from $\mathcal{O}(\log^2 h)$ to $\mathcal{O}(\log h)$. Since candidates are processed in parallel, the full single-variable inference step retains $\mathcal{O}(\log h)$ sequential operations. However, replacing $\mathbf z^t$ with $\tilde{\mathbf z}^t$ does not change which candidate is selected.

\begin{theorem}[Candidate selection]\label{thm:argmax}
Let $\hat{\mathbf y}\in\Delta([h])$ and $\emptyset\subset S\subseteq[h]$, and
let $\mathbf z^t$ and $\tilde{\mathbf z}^t$ be defined as in Equations~(4)
and~(7). Then, for $\epsilon\to0^+$,
\[
\argmin_{t\in S}\|\mathbf z^t-\hat{\mathbf y}\|_2^2
=\argmin_{t\in S}\|\tilde{\mathbf z}^t-\hat{\mathbf y}\|_2^2
=\argmax_{t\in S}\hat y_t.
\]
\end{theorem}

Proof can be found in Appendix~\ref{app:thm:argmax}. Hence the surrogate affects only the
continuous output of \layer{}, not the value it selects; we now characterise
the resulting continuous error exactly.

\begin{example}(Cont.'ed Ex.~\ref{example:problem_specification})
Consider the simplified case where $\know = \{{\color{yellow}\blacksquare}\} : Y$ (i.e., we know that the shape is always yellow), and suppose that the network outputs:
\begin{equation*}
    m_\theta(\mathbf{x}) = ({\color{yellow}\blacksquare}: 0.3, \; {\color{red}\blacksquare}: 0.5, \; {\color{blue}\blacksquare}: 0.2), \qquad \qquad \argmax \left(m_\theta(\mathbf{x})\right) = {\color{red}\blacksquare}
\end{equation*}
Let $t = {\color{yellow}\blacksquare}$ and (for simplicity) $\epsilon = 0.01$, then:
\begin{equation*}
    \mathcal{A}^{{\color{yellow}\blacksquare}} = \{j \in \{{\color{red}\blacksquare},{\color{blue}\blacksquare}\}: \hat y_{{\color{yellow}\blacksquare}} < \hat y_j + \epsilon\} = \{{\color{red}\blacksquare}\}
\end{equation*}
\begin{equation*}
\layer(\hat \bfy) = \left({\color{yellow}\blacksquare} : \frac{\hat y_{{\color{yellow}\blacksquare}}+\hat y_{{\color{red}\blacksquare}}+\epsilon}{2}, {\color{red}\blacksquare} : \tilde z_{{\color{yellow}\blacksquare}}^{{\color{yellow}\blacksquare}}-\epsilon, {\color{blue}\blacksquare} : \hat y_{{\color{blue}\blacksquare}} \right) = \left({\color{yellow}\blacksquare}: 0.405, {\color{red}\blacksquare}: 0.395, {\color{blue}\blacksquare} : 0.2 \right)
\end{equation*}
\end{example}

We now give an exact characterization of the distance between $\tilde \bfz^t$ and $\bfz^t$. The error decomposes into contributions from each spuriously included
index, weighted by its \emph{sub-threshold gap} $g_j = z_t^t - \epsilon - \hat y_j$.
\begin{theorem}[Surrogate fidelity]\label{thm:exact_error}
Let $\mathbf{z}^t$ be the solution of the fixed point system \eqref{eq:fixed_point_system}, $\tilde{\mathbf{z}}^t$ defined as in Equation~\eqref{eq:closed_form_solution_approximated}, and $\mathcal{W}^t = \mathcal{A}^t \setminus \mathcal{M}^t$. For each $j \in \mathcal{W}^t$, let $g_j = z_t^t - \epsilon - \hat y_j$.
Then
\begin{equation}\label{eq:exact_identity}
\bigl\| \tilde{\mathbf{z}}^t - \mathbf{z}^t \bigr\|_2^2
= \sum_{j \in \mathcal{W}^t} g_j^{2}
- \frac{1}{1 + |\mathcal{A}^t|}\Bigl(\sum_{j \in \mathcal{W}^t} g_j\Bigr)^{\!2}.
\end{equation}
\end{theorem}

The proof can be found in Appendix~\ref{app:proof_exact_error}. The exact identity in Equation~\eqref{eq:exact_identity} can be bounded 
by an
$\ell_2$ regularizer on the distance between $\hat{\mathbf{y}}$ and $\mathbf{z}^t$.

\begin{corollary}\label{cor:L2_bound}
Under the hypotheses of Theorem~\ref{thm:exact_error},
$\bigl\| \tilde{\mathbf{z}}^t - \mathbf{z}^t \bigr\|_2 \leq  \bigl\| \tilde\bfz^t - \hat{\mathbf{y}} \bigr\|_2.$
\end{corollary}

Corollary~\ref{cor:L2_bound} motivates adding an $\ell_2$ regularizer
$\|\tilde\bfz^t - \hat{\mathbf{y}}\|_2^2$ to the training objective: driving this
quantity to zero drives 
$\tilde{\mathbf{z}}^t \to \mathbf{z}^t$. 
Beyond controlling the surrogate fidelity, the $\ell_2$ regularizer serves a
complementary role during training: it counteracts the rank deficiency of the layer's Jacobian. This rank deficiency might cause problems during training as it results in a structurally identifiable subspace of upstream gradients being mapped to zero during back-propagation.

\begin{proposition}[Gradient attenuation]\label{prop:gradient_attenuation}
Let $\hat{\mathbf y} \in \Delta([h])$ be such that $t^\star$ and $\mathcal A^{t^\star}$
in~\eqref{eq:closed_form_solution_approximated} are locally constant at $\hat{\mathbf y}$. Let $\layer(\hat \bfy) = \tilde \bfz^{t^\star}$ and%
$V = \bigl\{\mathbf v \in \mathbb R^{h} : v_j = v_{t^\star},\ j \in \mathcal A^{t^\star}\bigr\}.$
Then the Jacobian $J = \partial\tilde{\mathbf z}^{t^\star}/\partial\hat{\mathbf y}$ equals the orthogonal projector
$P_V$. Hence, for any differentiable loss $\mathcal L_{\textnormal{task}}$,
\[
\nabla_{\hat{\mathbf y}}\bigl[\mathcal L_{\textnormal{task}} \circ \tilde{\mathbf z}^{t^\star}\bigr]
\;=\; P_V\,\nabla_{\tilde{\mathbf z}^{t^\star}}\mathcal L_{\textnormal{task}}\,;
\]
the back-propagated gradient is the orthogonal projection onto
$V$ of $\nabla_{\tilde{\mathbf z}^{t^\star}}\mathcal L_{\textnormal{task}}$, with the $|\mathcal A^{t^\star}|$-dimensional subspace $V^\perp$ mapped to zero by $J$.
 \end{proposition}
The proof can be found in Appendix~\ref{app:grad_attenuation}. The $\ell_2$ regularizer solves this problem by restoring the gradients  along the dimensions that are mapped to zero. 

\begin{proposition}[Gradient restoration]\label{prop:l2_restores_gradient}
Under the assumptions of Proposition~\ref{prop:gradient_attenuation}, consider the
composite training objective
$$
\mathcal L_{\textnormal{total}}(\hat{\mathbf y})
= \mathcal L_{\textnormal{task}}\bigl(\tilde{\mathbf z}^{t}(\hat{\mathbf y})\bigr)
+ \tfrac{\lambda}{2}\,\bigl\|\tilde{\mathbf z}^{t}(\hat{\mathbf y}) - \hat{\mathbf y}\bigr\|_2^2,
\qquad \lambda > 0.
$$
Then
$$
\nabla_{\hat{\mathbf y}}\mathcal L_{\textnormal{total}}
\;=\; \underbrace{P_V\,\nabla_{\tilde{\mathbf z}}\mathcal L_{\textnormal{task}}}_{\in\,V}
\;+\; \underbrace{\lambda\,\bigl(\hat{\mathbf y} - \tilde{\mathbf z}^{t}\bigr)}_{\in\,V^\perp}.
$$
\end{proposition}
The proof can be found in Appendix~\ref{app:gradient_restoration}. 
Propositions~\ref{prop:gradient_attenuation} and \ref{prop:l2_restores_gradient} together explain
why the $\ell_2$ regularizer plays a double role in our framework:
it both tightens the approximation bound (Corollary~\ref{cor:L2_bound}) and
stabilizes optimization by restoring the gradient along the  dimensions originally mapped to zero.

\subsection{General Case}

Extending \layer{} to multiple variables requires reasoning about how the assignment of one variable restricts the feasible values of the others. We process variables in a user-specified order $Y_1; \ldots; Y_n$ and, at each step, derive the set of values for $Y_i$ consistent with the constraints under the assignments already fixed for $Y_1, \ldots, Y_{i-1}$. The key technical tool is an inference rule over signed clauses, analogous to propositional resolution, that combines two clauses sharing a literal on the same variable into a single entailed clause. Iterating this rule lets us compile $\Pi$ into a per-variable form that exposes exactly the constraints needed at each step of inference, and reduces multi-variable satisfaction to a sequence of single-variable projections of the kind solved in Section~\ref{subsec:single_variable}.

We make use of the \emph{signed resolution} rule, a standard inference rule from the multiple-valued logic literature~\citep{hanle1994_cnf_in_finitely_valued_logics,beckert2000_signed_CNF}. Given two signed clauses both containing a term on variable $Y_i$: $(\clS : Y_i) \lor \varphi$ and  $(\clS' : Y_i) \lor \varphi'$,
where $\varphi$ and $\varphi'$ denote signed clauses over variables other than $Y_i$, the rule derives a new clause entailed by the premises:
\begin{equation}\label{eq:intersection}
    \frac{(\clS : Y_i) \lor \varphi \qquad (\clS' : Y_i) \lor \varphi'}{\left(\clS \cap \clS': Y_i \right) \lor \varphi \lor \varphi'}.
\end{equation}
Soundness of the rule is a classical result~\citep{hanle1994_cnf_in_finitely_valued_logics}.
Given an user-defined ordering over the variables $Y_1;Y_2;\ldots;Y_n$, we can use the intersection rule to compute the set of constraints necessary to determine the feasible assignment for each variable $Y_i$ knowing the initial assignment given by the model $m_\theta$ and the final assignments for the variables $Y_1, \ldots, Y_{i-1}$. The final set of constraints will then be used to build $\layer{}$. We describe the offline compilation below, and we provide the pseudocode for both the compilation and the inference steps in Algorithm~\ref{alg:casper}.

Starting from the end of the ordering: 
\begin{itemize}
    \item Let $\know_n = \know$,
    \item For every $i=n, \ldots, 1$, let $\bar \know_i =\left\{(\clS_j : Y_i) \lor \varphi_j \right\}_{j=1}^m \subseteq \know_i$ be the subset of clauses of $\know_i$ where $Y_i$ appears, which allows us to iteratively compute the set $\know_{i-1}$ as:
    \begin{equation}\label{eq:pi_i-1}
        \know_{i-1} = \left(\know_i\setminus \bar\know_{i}\right) \cup \text{Confl}_i\left(\bar\know_i\right), 
    \end{equation}
\end{itemize}
in which $\text{Confl}_i(\bar\know_i)$ is the {\sl conflict clauses set of $\bar\know_i$}:
\begin{equation}\label{eq:conflict_set}
    \mathrm{Confl}_i(\bar{\Pi}_i) = \left\{\bigvee_{j \in \clJ}\varphi_j \mid  \bigcap_{j \in \mathcal{J}}\clS_j = \emptyset, \emptyset \subset \mathcal{J} \subseteq [m] \right\}.
\end{equation}
The conflict clauses set represents the set of clauses that need to be added to the layer below the $i$th to avoid ending up in an inconsistent state, i.e., avoiding having $(\emptyset : Y_i) \in \know_i$. The conflict clauses set can equivalently be characterized as the closure of $\bar\know_i$
  under the intersection rule restricted to unsatisfiable combinations of signed terms.

\begin{lemma}[Equisatisfiability]\label{lem:equisat}
Let $\know_i$ be a finite and satisfiable set of signed clauses in the variables $Y_1, \ldots, Y_i$. Then $\know_i$ and $\know_{i-1}$ are equisatisfiable, and each assignment to $Y_1, \ldots, Y_{i-1}$ satisfying $\know_{i-1}$ can be extended to satisfy $\know_i$.
\end{lemma}
The proof is given in Appendix~\ref{app:equisat-proof}. 
We now show that the complexity of the \textsc{Compile} routine is governed by how many variables a conflict clauses set involves.
For a signed clause $\varphi$, let $\texttt{scope}(\varphi) = \{Y \mid \exists \ell = (\clS_j : Y_j) \in \varphi .\, Y = Y_j \}$, i.e., the scope of $\varphi$ is the set of variables occurring in some of its literals.
We define the \emph{interaction graph} of a set of signed clauses $\Pi$ as the undirected graph $\clG_{\Pi} = (\bfY, E_{\Pi})$, where $E_{\Pi} = \{ (Y_j,Y_z) \mid \exists \varphi \in \Pi.\, \{Y_j,Y_z\} \subseteq \texttt{scope}(\varphi), j \neq z \}$. Intuitively, the interaction graph of $\Pi$ has an edge between each pair of variables co-occurring in a clause in $\Pi$.
Given an ordering $d = Y_1 ; \ldots ; Y_n$, the \emph{induced} interaction graph $\clG_{\Pi}^{\star}(d)$ of $\Pi$ along $d$ connects all the neighbours of a variable $Y_i$ that precede it in $d$, recursively from $Y_n$ to $Y_1$ (cf. ~\cite[Def.~3.7]{Dechter2013}). 
Formally, $\clG_{\Pi}^{\star}(d) = (\bfY, E_0)$, where $E_n = E_{\Pi}$ and for $i = n, \ldots, 1$ we have:
\begin{equation}\label{eq:ei-1}
    E_{i-1} = E_i \cup \{ (Y_j, Y_z) \mid \{(Y_j, Y_i), (Y_i, Y_z)\} \subseteq E_i, \; j \neq z, \;\textit{and}\; j,z < i \}.
\end{equation}
Our bound is controlled by the maximum number of variables preceding some $Y_i$ in $d$ that are adjacent to it in $\clG_{\Pi}^{\star}(d)$. 
We call this number the \emph{conflict width}. For $i \in [n]$, the set $\clN_d(Y_i) = \{Y_j \mid j < i, (Y_j, Y_i) \in E_0 \}$
denotes the neighbours of $Y_i$ in $\clG_{\Pi}^{\star}(d)$ that precede it in $d$; the conflict width of $\Pi$ along $d$ is 
$w^{\star}(d) = \max_{i \in [n]}|\clN_d(Y_i)|$.
\begin{theorem}[Computational Complexity of \textsc{Compile}]\label{thm:complexity}
    Let $\Pi$ be a finite set of $r$ signed clauses over the variables $Y_1;\ldots;Y_n$, $k = \max_{i \in [n]}h_i$ and
    $w^{\star}(d)$ the conflict width of $\Pi$ along the ordering $d=Y_1;\ldots;Y_n$.  \textsc{Compile}$(\Pi, Y_1;\ldots;Y_n)$ 
    in Algorithm~\ref{alg:casper} can be implemented to run in
    \[ \textup{time}\; \clO((r+n) \cdot k^{w^{\star}(d)+1}) \qquad \textup{and} \qquad \textup{space}\; \clO(n \cdot k^{w^{\star}(d)}). \]
\end{theorem}
Proof in Appendix~\ref{app:complexity-proof}. Theorem~\ref{thm:complexity} shows that the cost of compilation grows exponentially with the 
conflict width $w^{\star}(d)$: this is the same worst-case as adaptive consistency and directional resolution~\citep{DECHTER199941}.

\begin{algorithm}[t]
\caption{Compilation \& Inference Step with \layer}
\label{alg:casper}
\begin{minipage}[t]{0.44\textwidth}
\begin{algorithmic}[1]
\Function{Compile}{$\know, Y_1; \ldots; Y_n$}
    \State $\know_n \leftarrow \know$
    \For{$i \leftarrow n$ \textbf{down to} $1$}
        \State Compute $\bar\know_i$, $\mathrm{Confl}_i(\bar{\Pi}_i)$
        \State $\Pi_{i-1} \leftarrow (\know_{i} \setminus \bar\know_{i}) \cup \mathrm{Confl}_i(\bar{\Pi}_i)$
    \EndFor
    \If{$\know_0$ is unsatisfiable}
        \State \Return UNSAT\_FLAG
    \Else
        \State \Return $\bar\know_1, \ldots, \bar\know_n$
    \EndIf
\EndFunction
\end{algorithmic}
\end{minipage}
\hfill
\begin{minipage}[t]{0.55\textwidth}
\begin{algorithmic}[1]
\Function{Infer}{$\hat\bfy, \bar\Pi_1, \ldots, \bar\Pi_n$}
    \For{$i \leftarrow 1$ \textbf{to} $n$}
        \State \!Compute $\clS_i(\bfa_{<i}) \subseteq [h_i]$ from $\bar\Pi_i$ under $\bfa_{<i}$
        \If{$\argmax \hat\bfy_i \in \clS_i(\bfa_{<i})$}
            \State \!$\tilde \bfz^{t^\star}_i \gets \hat\bfy_i$
        \Else
            \State \!Compute $\tilde \bfz^{t^\star}_i$ using Eq.~\eqref{eq:closed_form_solution_approximated} with input $\hat \bfy_i$
            \Statex \qquad \qquad \quad and constraint $\{\clS_i(\bfa_{<i}) : Y_i\}$
        \EndIf
        \State \!Compute $a_i = \argmax \tilde \bfz^{t^\star}_i$
    \EndFor
    \State \Return $\layer(\hat \bfy) = \left(\tilde\bfz^{t^\star}_1,  \ldots, \tilde\bfz^{t^\star}_n\right)$
\EndFunction
\end{algorithmic}
\end{minipage}

\end{algorithm}

Once the clause sets $\Pi_1,\ldots,\Pi_n$ have been computed, \layer{} is evaluated sequentially according to the variable ordering $Y_1,\ldots,Y_n$ as described below and in the inference step of Algorithm~\ref{alg:casper}.
Let $\hat{\bfy} = m_\theta(\bfx)$. We process the variables in the order 
$Y_1; \ldots; Y_n$. At step $i$, the assignments 
$\bfa_{<i} = [a_1, \ldots, a_{i-1}]$ for the preceding variables have already 
been determined by steps $1, \ldots, i-1$. These assignments induce a feasible 
set $\clS_i(\bfa_{<i}) \subseteq [h_i]$ consisting of all values of $Y_i$ 
consistent with the clauses in $\bar{\know}_i$ under $\bfa_{<i}$. By construction of the compilation procedure, every clause in $\Pi_1$ mentions only $Y_1$, and it can always be reduced to $\Pi_1 = \{\clS_1 : Y_1\}$ for some $\clS_1 \subseteq [h_1]$. Hence for $i = 1$, 
$\clS_1(\bfa_{<1}) = \clS_1$. Substituting $\bfa_{<i}$ into $\bar{\know}_i$ 
yields a set of constraints on $Y_i$ equivalent to the single signed clause 
$(\clS_i(\bfa_{<i}) : Y_i)$. If $\arg\max \hat{\bfy}_i \in \clS_i(\bfa_{<i})$, the prediction already satisfies the constraints and no correction is needed:
$a_i = \arg \max \hat{\bfy}_i$ and $\tilde{\bfz}^{t^\star}_i = \hat{\bfy}_i$. Otherwise, we compute the refined prediction 
$\tilde{\bfz}^{t^\star}_i$ for $Y_i$ using the closed-form update of 
Equation~\eqref{eq:closed_form_solution_approximated} with input $\hat{\bfy}_i$ and constraints $\{\clS_i(\bfa_{<i}) : Y_i\}$, and 
set $a_i = \arg\max \tilde{\bfz}^{t^\star}_i$. Iterating for $i = 1, \ldots, n$ yields 
the final refined output $\layer(\hat{\bfy}) = (\tilde{\bfz}^{t^\star}_1, \ldots, 
\tilde{\bfz}^{t^\star}_n)$. Notice that, at step $i$, computing $\clS_i(\bfa_{<i})$ from  the precomputed $\bar{\know}_i$ 
runs in $\mathcal{O}(\log |\bar \know_i|)$ sequential GPU operations.  
Hence, each 
step requires $\mathcal{O}(\log(|\bar \Pi_i| \cdot h_i))$ sequential GPU 
operations. 
The output of $\layer$ satisfies $\Pi$ as proven in 
Appendix~\ref{app:correctness-proof}.

\begin{theorem}[Correctness]\label{thm:correctness}
Let $\Pi$ be a finite and satisfiable set of signed clauses. For any datapoint $\bf{x}$, model $m_\theta$ and variable ordering, the corresponding output $\layer(m_\theta(\bf{x}))$ satisfies $\Pi$ for $\epsilon \to 0^+$.
\end{theorem}

Notice that the feasible set $\clS_i(\bfa_{<i})$ depends only on the
preceding discrete assignments, rather than on the continuous values produced
by the preceding projections. Hence, since the surrogate and exact fixed-point
projections select the same candidate at every step
(Theorem~\ref{thm:argmax}), an early continuous over-correction by the
surrogate cannot restrict the feasible sets available to later variables:

\begin{corollary}[Sequential assignment equivalence]\label{cor:snowball_effect}
Consider a finite and satisfiable set of signed clauses $\Pi$ and a variable ordering $Y_1;\dots;Y_n$. 
Given $\hat\bfy=m_\theta(\bf{x})$ for some $\bf{x}\in\mathcal X$, let $\tilde\bfa$
be the assignment induced by $\layer(\hat\bfy)$, $\bfa$ be the assignment
induced by the same procedure when each surrogate projection is replaced by
the exact fixed-point projection, and let $\bfa^\star$ be the sequential
feasible argmax assignment,
$a^\star_i=\argmax_{t\in\clS_i(\bfa^\star_{<i})}\hat y_{i,t}$.
Then, for $\epsilon\to0^+$:
$$
  \text{(i) }  \bfa=\tilde\bfa=\bfa^\star; \qquad \text{and} \qquad
  \text{(ii) } \clS_i(\bfa_{<i})=\clS_i(\tilde\bfa_{<i})=\clS_i(\bfa^\star_{<i}),
  \forall i\in[n].
$$
\end{corollary}

Proof in Appendix~\ref{app:cor:snowball_effect}. Longer sequences therefore only accumulate local continuous discrepancies, characterised at each step by Theorem~\ref{thm:exact_error}, which do not compound through the feasible sets.

\section{Experimental Analysis}\label{sec:experiments}

We conduct our analysis over three tasks: the warcraft pathfinding task~\cite{poganvcic2019differentiation}, 11 variations of the standard MNIST Arithmetic task~\cite{manhaeve2018deepproblog} considering different number of digits to sum and different ways of summing them, and sudoku solving~\cite{wang2019satnet}. For each task  we apply a timeout (indicated as T/O) of 5s per training-example (consistent with \citep{choi2025ctsketch}). The experiments for warcraft and MNIST were run on a workstation equipped with 3 NVIDIA RTX 4500 each with 26GB VRAM, while the sudoku solving task was run on a workstation equipped with 1 NVIDIA H200.

\begin{table}[t]
    \centering
    \caption{Warcraft Pathfinding Task.}
    \label{tab:warcraft}
        \begin{tabular}{l cc cc}
        \toprule
        & \multicolumn{2}{c}{\textbf{12$\times$12}} & \multicolumn{2}{c}{\textbf{30$\times$30}} \\
        \cmidrule(lr){2-3} \cmidrule(lr){4-5}
        \textbf{Method} & {Optimal(\%)} & {Consistent (\%)} & {Optimal(\%)} & {Consistent (\%)} \\
        \midrule
        ResNet-18                 & $59.7 \msmall{\pm 1.2}$ & $62.5 \msmall{\pm 1.0}$ & $5.1 \msmall{\pm 0.6}$ & $6.4 \msmall{\pm 1.1}$ \\
        \midrule
        \addlinespace[2pt]
        \rowcolor{exactshade}\multicolumn{5}{l}{\textit{Exact Methods}} \\
        \addlinespace[2pt]
        DPL                       & T/O & T/O & T/O & T/O \\
        CCN+                      & $74.2 \msmall{\pm 0.6}$ & $\mathbf{100.0} \msmall{\pm\mathbf{0.0}}$ & T/O & T/O \\
        SPL                       & T/O & T/O & T/O & T/O \\
        \addlinespace[6pt]
        \rowcolor{approxshade}\multicolumn{5}{l}{\textit{Approximate Methods}} \\
        \addlinespace[2pt]
        Scallop                   & $63.6 \msmall{\pm 5.3}$ & $66.1 \msmall{\pm 5.7}$ & T/O & T/O \\
        DSL                       & $65.1 \msmall{\pm 1.3}$ & $68.3 \msmall{\pm 1.8}$ & $4.8 \msmall{\pm 0.5}$ & $5.8 \msmall{\pm 0.7}$ \\
        CTSketch                  & $60.4 \msmall{\pm 1.1}$ & $62.8 \msmall{\pm 1.1}$ & $5.1 \msmall{\pm 0.3}$ & $6.1 \msmall{\pm 0.6}$ \\
        \midrule
        \rowcolor{ablationshade}\multicolumn{5}{l}{\textit{Ablation Studies}} \\
        \addlinespace[2pt]
        \layer{} (Train Only)     & $57.8 \msmall{\pm 0.9}$ & $60.5 \msmall{\pm 0.7}$ & $5.7 \msmall{\pm 0.2}$ & $9.4 \msmall{\pm 1.6}$ \\
        \layer{} (Inference Only) & $78.3 \msmall{\pm 0.8}$ & $\mathbf{100.0} \msmall{\pm\mathbf{0.0}}$ & $11.5 \msmall{\pm 0.8}$ & $\mathbf{100.0} \msmall{\pm\mathbf{0.0}}$ \\
        \layer{} (w/o $\ell_2$)   & $77.5 \msmall{\pm 0.6}$ & $\mathbf{100.0} \msmall{\pm\mathbf{0.0}}$ & $11.0 \msmall{\pm 1.4}$ & $\mathbf{100.0} \msmall{\pm\mathbf{0.0}}$ \\
        \midrule
        {\layer}                  & $78.7 \msmall{\pm 0.3}$ & $\mathbf{100.0} \msmall{\pm\mathbf{0.0}}$ & $\mathbf{12.1} \msmall{\pm\mathbf{1.1}}$ & $\mathbf{100.0} \msmall{\pm\mathbf{0.0}}$ \\
        \bottomrule
    \end{tabular}%
    \vspace{-12pt}
\end{table}

\subsection{Warcraft Pathfinding}

\vspace{-0.1cm}
\begin{wrapfigure}{r}{0.31\textwidth}
  \centering
    \vspace{-0.8cm}
    \includegraphics[
        width=0.24\textwidth,
        trim={0 0 215 0},
        clip
    ]{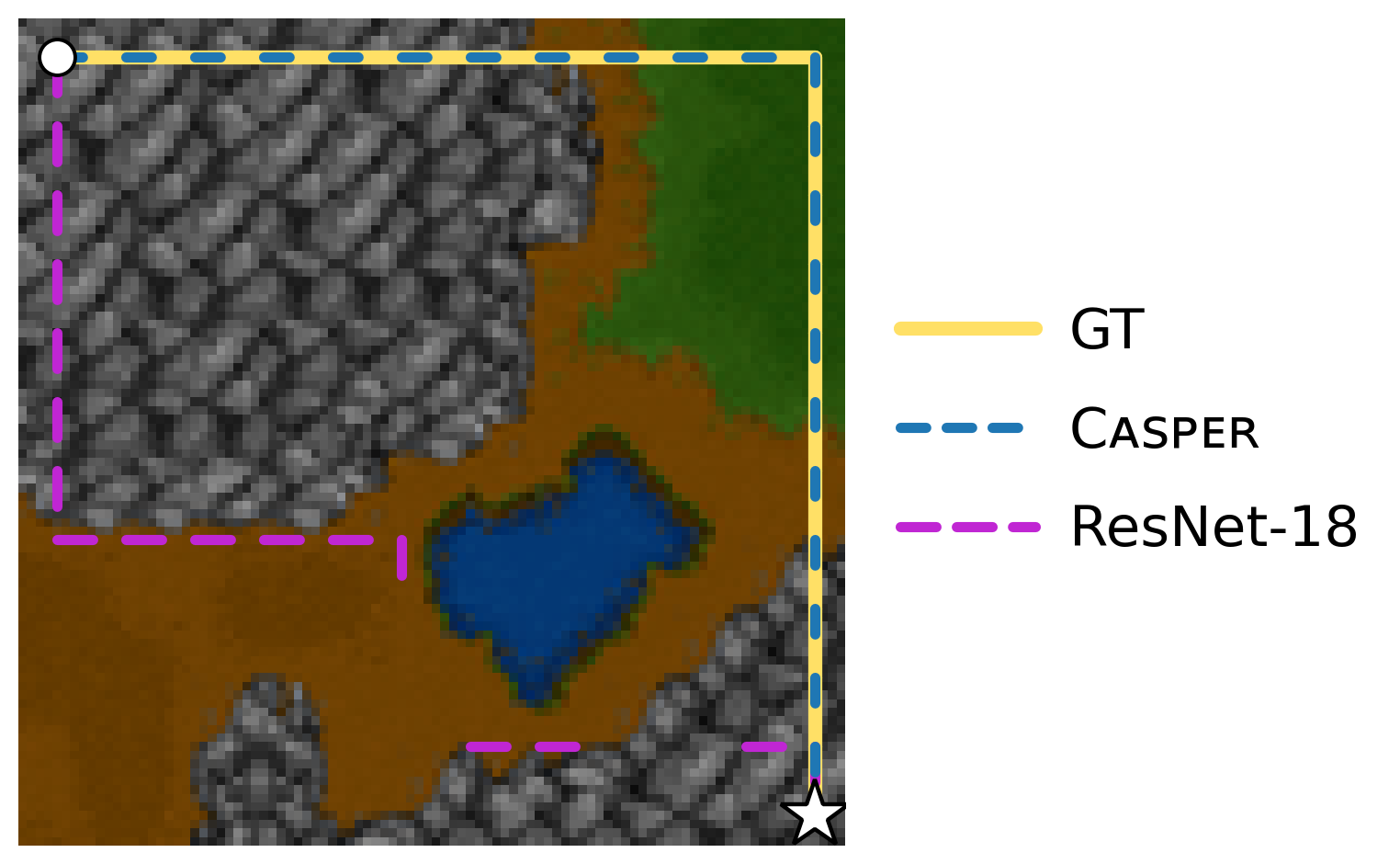}

    \vspace{0.0em}

    \begin{tabular}{ccc}
        \includegraphics[
            width=0.07 \textwidth,
            trim={380 190 80 105},
            clip
        ]{warcraft_12x12_intraining_vs_inference_only.png}
        \includegraphics[
            width=0.11\textwidth,
            trim={380 150 40 140},
            clip
        ]{warcraft_12x12_intraining_vs_inference_only.png}
        \includegraphics[
            width=0.13\textwidth,
            trim={380 112 14 185},
            clip
        ]{warcraft_12x12_intraining_vs_inference_only.png}
    \end{tabular}
    \caption{Qualitative example for Warcraft Pathfinding. GT stands for ground truth.}
    \label{fig:warcraft-ablation}

    \vspace{-0.6cm}
\end{wrapfigure}

The goal  is predicting the minimum-cost path from the top-left to the bottom-right corner 
of an $N \times N$ grid, given an image rendered from Warcraft II terrain tiles where each tile has an associated traversal cost. We consider grid-sizes $N \in \{12, 30\}$ where a path is classed as valid if it is connected, contained within the grid, and terminates at the goal. To exemplify the  task, we show on qualitative example in Figure~\ref{fig:warcraft-ablation}.
In Table~\ref{tab:warcraft}, we report the percentage of test instances for which each method predicts an optimal path, i.e., a path with minimum cost, and a consistent path, i.e., a path satisfying the task constraints. These are denoted by {Optimal}(\%) and {Consistent}(\%), respectively. To complement these metrics, we also report a cumulative cost-optimality analysis: for each tolerance factor $\alpha \geq 1$, we measure the fraction of predictions whose cost is within a factor $\alpha$ of the optimal path cost in Figure~\ref{fig:warcraft_cost_analysis} (Appendix~\ref{app:experiments_warcraft}). This provides a finer-grained view of how close non-optimal predictions are to the optimum.
We compare \layer{} against both exact (SPL~\citep{ahmed2022semantic}, DPL~\citep{manhaeve2018deepproblog}, CCN+~\citep{giunchiglia2024ccn+}) and approximate baselines (DSL~\citep{maene2023soft}, Scallop~\citep{huang2021scallop}, CTSketch~\citep{choi2025ctsketch}). We evaluate different variants of CTSketch and 
report the best-performing configuration; the full ablation is given in Appendix~\ref{app:experiments_warcraft}. 
We do not include Dijkstra-based methods such as A-NeSI~\citep{van2023nesi}
and I-MLE~\citep{niepert2021imle}, which predict per-tile costs and recover the path with a shortest-path solver. These are complementary approaches that address a different formulation of the task: the solver performs the combinatorial optimization, while the network learns only the tile costs. In our setting, as in SPL~\citep{ahmed2022semantic}, the network instead directly predicts the structured output, with no shortest-path solver, and constraints are enforced on this prediction. All models use the standard ResNet-18~\citep{he2016deep} backbone. \layer{} and all baselines except CTSketch work with the step-wise formulation of the path constraint (Appendix~\ref{sec:trimmed-steps}), which constrains a time-unrolled walk over grid cells and not the path itself. Its feasible set contains every simple path but also walks that revisit a cell, thus for all methods all metrics are computed on the loop-erased path (Appendix~\ref{app:experiments_warcraft}).

We further conduct an ablation study by testing the application of \layer{} only at inference time (Inference Only), by training \layer{} without the $\ell_2$ regularizer (w/o $\ell_2$), and by training with \layer{} and then removing it at inference time (Train Only). Table~\ref{tab:warcraft} shows three findings. First, \layer{} is the only method  that produces $100\%$ valid paths on both grid sizes.%
Among exact methods, only \layer{} runs to completion on $30 \times 30$; SPL, DPL, and CCN+ all time out, while CTSketch's optimal-path rate ($5.1\%$) does not exceed the unconstrained ResNet-18 baseline. \layer{} reaches $12.1\%$ optimal on $30 \times 30$, more than doubling CTSketch.
Second, the ablation study isolates the contribution of each component of \layer{}. Applying \layer{} only at inference time already reaches $78.3\%$ optimal on $12 \times 12$ and produces valid paths $100\%$ of the times. This demonstrates that \layer{} can be deployed as a post-hoc constraint-enforcement step on existing predictors with no retraining. Conversely, training with \layer{} but removing it at inference (Train Only) performs similarly to the unconstrained ResNet-18, showing that the gains stem from enforcing the constraints at inference.
Third, the $\ell_2$ regularizer provides a consistent improvement on both grid sizes: $+1.2$ points on $12 \times 12$ and $+1.1$ points on $30 \times 30$, in line with Section~\ref{subsec:single_variable}.

\begin{figure}[hbt]
    \centering
    \begin{subfigure}[b]{\linewidth}
        \centering
        \includegraphics[width=0.75\linewidth]{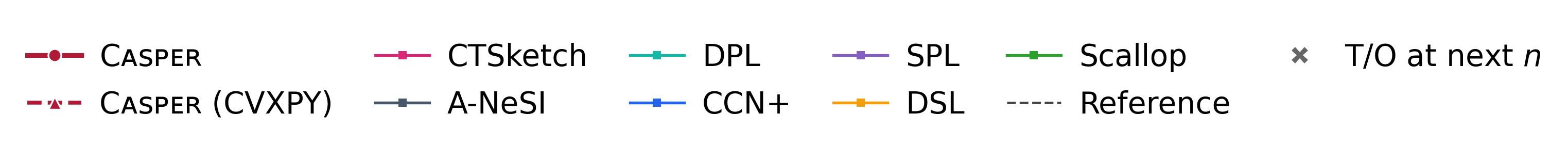}
    \end{subfigure}
    \begin{subfigure}[b]{0.49\linewidth}
        \centering
        \includegraphics[width=\linewidth]{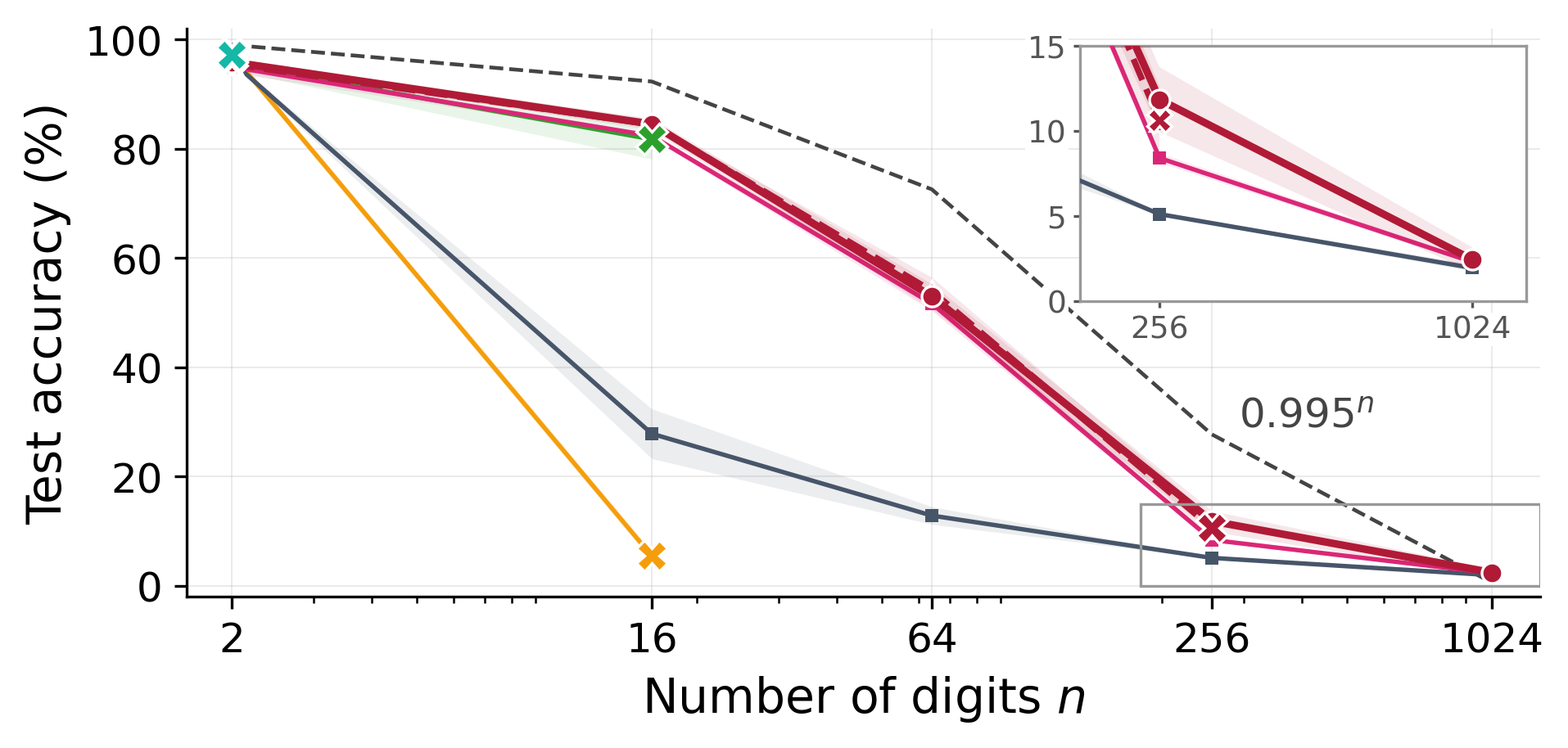}
        \caption{MNIST-Sum}
        \label{fig:mnist-sum-scaling}
    \end{subfigure}%
    \hfill%
    \begin{subfigure}[b]{0.49\linewidth}
        \centering
        \includegraphics[width=\linewidth]{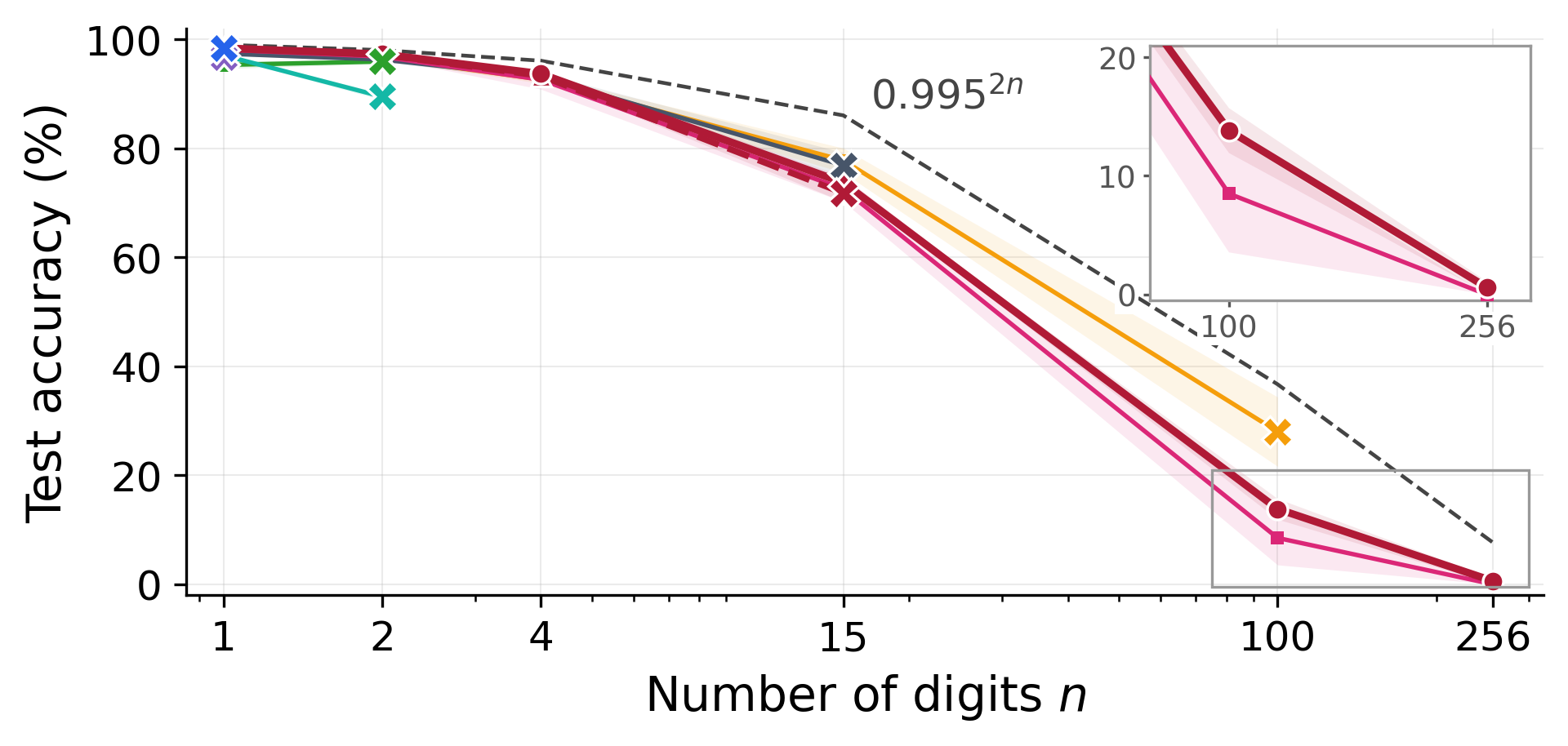}
        \caption{MNIST-Add}
        \label{fig:mnist-add-scaling}
    \end{subfigure}
    
    \caption{Test accuracy as a function of problem size $n$ on MNIST (a) Sum and (b) Add. 
    $\times$ marks the largest $n$ each method completed before timing
    out or failing.}
    \label{fig:mnist-scaling}
    \vspace{-12pt}
\end{figure}

\subsection{MNIST Arithmetic}
We next investigate the scalability of \layer{} by considering two arithmetic reasoning tasks where the constraint structure grows with the input size: MNIST-Sum and MNIST-Add. 
For MNIST-Sum, we consider the problem of computing the sum of $n$ handwritten digits sampled from the MNIST dataset~\citep{lecun2002gradient} with $n = 2^k$ for $k \in \{1, 4, 6, 8, 10\}$.%
For MNIST-Add, the objective is to compute the sum of two $n$-digit numbers represented as sequences of handwritten digits.  As in \cite{choi2025ctsketch}, we test using $n \in \{2, 15, 100, 256\}$.%
In both tasks, the constraint requires the final output to follow the sum rules and we train under full  supervision.
We define a reference accuracy of $0.995^n$ for MNIST-Sum and   $0.995^{2n}$ for MNIST-Add, representing the expected  accuracy of a MNIST classifier with $99.5\%$ per-digit accuracy. In addition to the previous task baselines, we include A-NeSI~\citep{van2023nesi}, and we evaluate
\begin{wraptable}{r}{0.56\textwidth}
    \centering
    \vspace{-0.3cm}
    \caption{Time (s) per epoch on MNIST-Sum.}
    \label{tab:mnist_sum_time}
    \setlength{\tabcolsep}{3pt}
    \definecolor{exactshade}{RGB}{252, 228, 230}
    \definecolor{approxshade}{RGB}{223, 237, 247}
    \begin{tabular}{l ccccc}
        \toprule
        Method / $n$ & $2$ & $16$ & $64$ & $256$ & $1024$ \\
        \midrule
        \addlinespace[2pt]
        \rowcolor{exactshade}\multicolumn{6}{l}{\textit{Exact Methods}} \\
        \addlinespace[2pt]
        DPL & 58.4  & T/O   & T/O   & T/O    & T/O  \\
                CCN+      & 26.9  & T/O   & T/O   & T/O    & T/O \\
        SPL       & 4.1   & T/O   & T/O   & T/O    & T/O \\
        \addlinespace[6pt]
        \rowcolor{approxshade}\multicolumn{6}{l}{\textit{Approximate Methods}} \\
        \addlinespace[2pt]

        A-NeSI    & 11.8 & 24.7 & 66.4 & 232.3 & 995.8 \\
        Scallop$^{\dagger}$   & 3100  & 3400  & T/O   & T/O    & T/O \\
        DSL       & 33.3 & 386.6 & T/O   & T/O    & T/O \\
        CTSketch  & 0.62  & 4.2  & 14.9 & 60.9  & 239.2 \\
         \midrule
         \layer{} (\textit{cvxpy)} & 33.7 & 94.8 & 257.6 & 815.2 & T/O \\
        \layer{}  & 1.5 & 5.9 & 17.5 & 62.3 & 240.3 \\
        \bottomrule
    \end{tabular}%
    \\[2pt]
    {\footnotesize $^{\dagger}$Scallop runs on CPU.}
    \vspace{-0.4cm}
\end{wraptable}
 a second variant of \layer{} that for every variable 
solves the optimization problem in Equation~\eqref{eq:optimization_problem} using 
\textit{cvxpy}~\citep{diamond2016cvxpy,agrawal2018_cvxpy} with default solver SCS \citep{ocpb:16}. All methods use a Le-Net~5 backbone \citep{lecun2002gradient}.%

We report the results obtained for every dimension in Figure~\ref{fig:mnist-scaling} and the time (in seconds) to complete an epoch for each method in Table~\ref{tab:mnist_sum_time}. Figure~\ref{fig:mnist-scaling} shows that of the exact methods only \layer{} is able to scale from the basic case; DPL, SPL and CCN+ time out for all larger problem sizes.  Regarding the approximate methods, on MNIST-Sum we can see that Scallop and DSL fail to scale beyond $n=16$, with Table~\ref{tab:mnist_sum_time} showing that Scallop in particular, being CPU bound, incurs a significant increase in training time. Additionally, on MNIST-Sum, for every $n \geq 16$, \layer{} outperforms both CTSketch and A-NeSI in terms of accuracy. 
Regarding MNIST-Add, \layer{} is competitive with all approximate methods at all scales, with the exception of DSL and A-NeSI at $n = 15$, and DSL leading at $n = 100$ ($28.0\%$ vs \layer{} at $13.8\%$ and CTSketch at $8.5\%$). However, DSL times out for $n=256$ where \layer{} marginally outperforms CTSketch. 

Regarding the comparison with \layer{} using \textit{cvxpy}, we can see that
the \textit{cvxpy} variant performs comparably to the standard \layer{} on
both MNIST-Sum and MNIST-Add, but times out at $n = 1024$ and $n = 100$,
respectively. This confirms the need for the closed-form surrogate: 
solving the fixed-point system exactly is tractable only at small scale.

\subsection{Sudoku Solving}

We consider the SatNet dataset~\citep{wang2019satnet} composed of $10{,}000$ nine-by-nine Sudoku boards constructed with MNIST digits, 
whose an average number of hints is equal to $36{.}22$ (min. $31$, max. $42$). The task is to output the completed final board.
The pipeline we adopt is the same as~\cite{cornelio2023learning}: 
(i) a CNN classifies MNIST hints, 
(ii) a board solver completes the board and it is followed by 
(iii) a masking network who blanks out unreliable cells. Finally, 
(iv) a Prolog solver returns the final board. For (ii) we test two different models: the pretrained neuro-solver from~\cite{cornelio2023learning} and the LLM 
Qwen2.5-14B-Instruct~\citep{qwen2025}, which we finetuned over the training SatNet boards. We use the same implementation of \layer{} interchangeably on top of both solvers, projecting the cells distribution wrt. the $27$ groups, over $9$ rows, $9$ columns, and $9$, three-by-three, boxes. \layer{} enforces the
non-duplicate constraint individually across each. Table~\ref{tab:sudoku}, reports the output accuracy of the intermediate solver (\textit{NeuroAcc}), the percentage of CNN errors it can recover from (\textit{RecRate}) and the final Prolog module (\textit{PipeAcc}). On top of improving the final accuracy, \layer{} more than doubles the solver performance, while consistently recovering through different solvers almost $95\%$ of the CNN perception errors.

\section{Related Work}

\layer{} belongs to a broader effort in neurosymbolic AI~\cite{third_wave_survey_garcez,Marra2024_survey} to integrate symbolic background 
\begin{wraptable}{r}{0.55\textwidth}
    \centering
    \caption{Sudoku solving test results.}
    \label{tab:sudoku}
    \setlength{\tabcolsep}{4pt}

    \definecolor{neuroshade}{RGB}{223, 237, 247}
    \definecolor{llmshade}{RGB}{232, 245, 233}
\vspace{-0.2cm}
    \begin{tabular}{lccc}
        \toprule
        \textbf{Method} & \textit{NeuroAcc} & \textit{RecRate} & \textit{PipeAcc} \\
        \midrule

        \addlinespace[2pt]
        \rowcolor{neuroshade}\multicolumn{4}{l}{\textit{Board solver} from \citep{cornelio2023learning}} \\
        \addlinespace[2pt]

        A-NesI           & $18.9 \msmall{\pm7.3}$ & $89.5 \msmall{\pm1.3}$ & $70.0 \msmall{\pm24.7}$ \\
        CTSketch         & $25.2 \msmall{\pm0.3}$ & $88.2 \msmall{\pm2.3}$ & $81.0 \msmall{\pm1.7}$ \\
        \textbf{\layer}  & $\textbf{66.0} \msmall{\pm\textbf{0.0}}$ & $\textbf{94.2} \msmall{\pm\textbf{1.1}}$ & $\textbf{84.7} \msmall{\pm\textbf{0.7}}$ \\

        \addlinespace[6pt]
        \rowcolor{llmshade}\multicolumn{4}{l}{\textit{LLM board solver (Qwen2.5-14B-Instruct finetuned)}} \\
        \addlinespace[2pt]

        A-NesI           & $25.6 \msmall{\pm0.2}$ & $94.7 \msmall{\pm0.7}$ & $68.3 \msmall{\pm2.4}$ \\
        CTSketch         & $25.4 \msmall{\pm 0.0}$ & $88.6 \msmall{\pm 0.4}$ & $80.0 \msmall{\pm0.0}$ \\
        \textbf{\layer}  & $\textbf{65.3} \msmall{\pm\textbf{0.5}}$ & $\textbf{95.3} \msmall{\pm\textbf{0.6}}$ & $\textbf{81.8} \msmall{\pm\textbf{1.0}}$ \\
        \bottomrule
        \vspace{-0.8cm}
\end{tabular}
\end{wraptable}knowledge into neural predictors.%
The earliest neurosymbolic methods incorporate constraints as differentiable training signals~\cite{LTN,badreddine2022_LTN,diligenti2017sbrnn,krieken2022_analyzing,scassola2025_diffusion_constraints,xu2018semantic,ahmed_2022_entropy_regularization,pryor2023NeuPSL}.  These methods incur no inference-time overhead but cannot guarantee that test predictions satisfy the constraint.
A second family treats the symbolic component as a probabilistic program.
DeepProbLog~\cite{manhaeve2018deepproblog} and NeurASP~\cite{yang2020neurasp}
perform exact weighted model counting (WMC), but inference is \#P-hard. Scallop~\cite{huang2021scallop} retains only the
top-$k$ proofs, so discarded proofs contribute neither probability mass nor
gradient; DeepSoftLog~\cite{maene2023soft} adds soft unification and, on
multi-digit addition, retains exact inference through a hand-designed
digit-wise carry recursion, outperforming \layer{} on MNIST-Add at $n=100$,
although this factorisation does not directly apply to MNIST-Sum, where it
times out beyond 16 digits; DeepStochLog~\cite{winters2022_deepstochlog} replaces probabilistic with stochastic semantics. A complementary line bypasses
logic-programming languages: A-NeSI~\cite{van2023nesi} amortises WMC with a
learned neural surrogate, introducing approximation and generalisation error;
CTSketch~\cite{choi2025ctsketch} encodes the program as a tensor compressed
via tensor-train decomposition;
Softened Symbol Grounding~\cite{li2023_softened} uses MCMC to sample symbol assignments
consistent with the constraint. Across this family, scalability is bought at
the cost of exactness or, in the case of DeepSoftLog, generality, and
constraint satisfaction cannot be guaranteed at test time.
The closest family to ours injects constraints at the predictive layer, guaranteeing that every output is feasible. C-HMCNN~\cite{giunchiglia2020}, CCN~\cite{giunchiglia2021} and CCN+~\cite{giunchiglia2024ccn+} clamp predictions into coherence intervals; SPL~\cite{ahmed2022semantic} multiplies a predictive probabilistic circuit by a compiled constraint circuit and renormalises; MultiplexNet~\cite{hoernle2022multiplexnet} parameterises outputs as mixtures over feasible regions. The same principle has recently been applied to controlled language generation, e.g., Ctrl-G~\cite{zhang2024_ctrl-G}.%
\layer{} belongs to this final family but breaks its scalability bottleneck.
\layer{} natively handles categorical variables via a closed-form
Euclidean projection, generalising the simplex projection of Duchi et
al.~\cite{duchi2008_projections} to argmax constraints, that is computable in
a single forward pass with parallel depth bounded by the number of variables,
independent of constraint complexity. Conceptually, \layer{} is also close to
differentiable convex-optimisation layers~\cite{amos2017_optnet,agrawal2019_cvxpy},
but differently from them it is specialised to categorical structured
prediction.

\section{Conclusions}
In this paper we presented \layer, a differentiable neurosymbolic layer that enforces hard logical constraints over categorical variables while scaling to thousands of variables. In the single-variable case, constraint satisfaction reduces to a Euclidean projection onto the argmax-feasible region of the simplex, which we solve via a closed-form surrogate with exactly characterised error and gradient behaviour. The general case is handled by an offline compilation step based on signed resolution that reduces multi-variable satisfaction to a sequence of such single-variable projections, parallelizable across variables which are independent according to the constraints.
Across our benchmarks, \layer{} is the only method that simultaneously guarantees satisfaction and scales. Indeed, it predicts valid paths on 100\% of 30×30 Warcraft instances, where all exact baselines time out and the strongest approximate baseline falls below 7\%. Further, it is the only exact method that scales to MNIST sums of more than a thousand digits. Finally, it more than doubles the post-layer accuracy of the strongest, unconstrained scalable baseline on Sudoku.%

\section*{Acknowledgements}
This work was supported by the AI for Math Fund, which is managed by Renaissance Philanthropy in partnership with founding donor XTX Markets.  Lohith Konathala is funded by the Engineering and Physical Sciences Research Council (EPSRC) and IceBrook through an Industrial Doctoral Landscape Award (IDLA) studentship.
The authors acknowledge the use of resources provided by the Isambard-AI National AI
Research Resource (AIRR). Isambard-AI is operated by the University of Bristol and is funded by
the UK Government’s Department for Science, Innovation and Technology (DSIT) via UK Research
and Innovation; and the Science and Technology Facilities Council [ST/AIRR/I-A-I/1023]. We thank the anonymous
reviewers and area chairs for their helpful comments and feedback.

\bibliographystyle{plainnat}
\bibliography{references}

\newpage

\appendix

\section{Limitations}
\label{app:limitations}

\layer{} targets a specific corner of the neurosymbolic problem space —
guaranteed satisfaction of hard logical constraints over discrete categorical
variables — and the design choices that make this corner tractable at scale
carry their own limitations, which we discuss here.

The simplex projection at the heart of \layer{}{} is built for finite-domain
categorical variables and signed clauses. Constraints over continuous outputs \cite{stoian2025,Wang2023_linsatnet,smet2023_deepseaproblog,donti2021_dc3}, soft
constraints encoding preferences, and constraints mixing discrete and
continuous quantities all fall outside its scope: handling them would require
redesigning the projection geometry rather than re-instantiating it. Within the categorical setting, our approach also relies on a closed-form
over-approximation $\clA^t$ of the exact active set $\clM^t$
(Lemma~\ref{lem:overapproximation}) in order to retain $\mathcal{O}(\log h)$
parallel depth. This approximation affects only the continuous output of
\layer{}: the surrogate selects the same value as the exact projection at
every step (Theorem~\ref{thm:argmax} and
Corollary~\ref{cor:snowball_effect}), and Theorem~\ref{thm:exact_error}
characterises the resulting gap exactly, with Corollary~\ref{cor:L2_bound}
bounding it for every input. The $\ell_2$ regulariser
(Proposition~\ref{prop:l2_restores_gradient}) shrinks this bound during
training, but only on average over the training distribution, not on every
test input. Inference-only deployments, which never see the
regulariser, therefore incur a small but measurable degradation, as the
(Inference-Only) ablation in Table~\ref{tab:warcraft} shows.

Finally, several active areas of neurosymbolic learning are complementary to
ours rather than addressed by it: reasoning shortcuts
\citep{marconato2023_not_all_neurosymbolic, Marra2024_survey}, modelling dependencies between
predicted symbols \citep{krieken2025_nesy_diffusion}, and concept-based
interpretability
\citep{koh2020cbm, giannini2024_interpretable, barbiero2023_interpretable}.
\layer{} composes naturally with methods targeting these axes — for instance,
the categorical output of a concept bottleneck model can itself serve as the
variables on which \layer{} enforces constraints — but a thorough empirical
study of such combinations is beyond the scope of this paper.

\section{Proof of Theorem~\ref{prop:closed-form update}\label{app:proof_minimization_problem_solution_fixed_point}}

\paragraph{Statement.} {\sl Let $\hat \bfy \in \Delta([h])$ and $\clS \subseteq [h]$, the $\epsilon$-margin solution to the optimization problem in
\begin{equation}\label{eq:optimization_probelm_app}
     {\argmin}_{\bfz \in \Delta([h])} \|\bfz - \hat\bfy\|_2^2 \qquad \text{ subject to } 
        \arg\!\max_{j \in [h]}  \bfz \in \clS.
\end{equation}
is given by $\bfz^{t^\star}$ with $t^\star$ equal to
     \begin{equation}\label{eq:app_solution_optimization_problem}
     t^\star \in \argmin_{t\in \clS}\|\bfz^t - \hat \bfy\|_2^2,
\end{equation}
where for every $t \in \clS$, $\bfz^t$ is the solution of the convex minimization problem obtained by fixing $t$ as the unique maximizer: 
\begin{equation}\label{app:eq:zt_def_in_statement}
    z^t_j = \begin{cases}
        \hat y_j & \text{if } j \not\in (\clM^t \cup \{t\})\\
        z_t^t - \epsilon & \text{if }  j \in \clM^t\\
        \frac{
\hat y_t + \sum_{j\in \mathcal M^t} \hat y_j + |\mathcal M^t|\epsilon
}{
1+|\mathcal M^{t}|} & \text{if } j = t
    \end{cases},
\end{equation}
 $\clM^t = \{j \in [h] \setminus \{t\} : \hat y_j \ge z_t^t - \epsilon\}$ represents the {\sl set of active categories} given $t$, and $\epsilon > 0$ is an arbitrarily small constant.}

\begin{proof}

The $\epsilon$-margin feasible space of Equation~\eqref{eq:optimization_probelm_app} decomposes as
$$
\{z \in \Delta([h]) : \exists\, t \in \mathcal{S},\; z_t \geq z_j + \epsilon\;\forall j \neq t\} = \bigcup_{t \in \mathcal{S}} C^t, \quad C^t = \{z : z_t \geq z_j + \epsilon\}.
$$
Since every $C^t$ is an intersection of halfspaces with the simplex, it is closed and convex, and the objective $\|\bfz^t - \hat \bfy\|_2^2$ is convex and continuous. Hence, 
$$
\min_{\bfz \in \bigcup_{t\in \clS}C^t} \|\bfz - \hat \bfy\|_2^2 = \min_{t \in \clS}\min_{\bfz \in C^t}\|\bfz - \hat \bfy\|_2^2.
$$
This proves Equation~\eqref{eq:app_solution_optimization_problem} is the $\epsilon$-margin solution of the optimization problem in Equation~\eqref{eq:optimization_probelm_app} provided that $\bfz^t$ is equal to Equation~\eqref{app:eq:zt_def_in_statement}.

Consider $t \in \clS$.  
We solve
\begin{equation*}
    \min_\bfz \|\bfz - \hat \bfy\|_2^2
\end{equation*}
subject to 
\begin{equation}\label{eq:primal_feasibility_cond_app}
    z_t \ge z_j + \epsilon \quad \forall j \in [h] \setminus \{t\}, \qquad \sum_{j=1}^{h} z_j = 1, \qquad z_j \geq 0 \quad \forall j \in [h]. 
\end{equation}
This yields a well-posed convex problem enforcing uniqueness of the maximizer with margin $\epsilon$.%

The Lagrangian is: 
\begin{equation*}
    \clL = \sum_{j=1}^h (z_j - \hat y_j)^2 + \nu \left(\sum_{j=1}^h z_j - 1 \right) - \sum_{j \in [h] \setminus \{t\}} \lambda_j (z_t - z_j - \epsilon) - \sum_{j=1}^h \mu_j z_j.
\end{equation*}
The KKT conditions (which include the primal feasibility in Equation~\eqref{eq:primal_feasibility_cond_app}) give us: 
\begin{equation*}
    \frac{\partial \clL }{\partial z_j} = 2(z_j - \hat y_j) + \nu + \lambda_j - \mu_j = 0, \qquad \forall j \in [h] \setminus t  \qquad \text{(Stationarity (1))}
 \end{equation*}
 \begin{equation*}
    \frac{\partial \clL }{\partial z_t} = 2(z_t - \hat y_t) + \nu - \sum_{j \in [h] \setminus \{t\}} \lambda_j - \mu_t = 0 \qquad \text{(Stationarity (2))}
 \end{equation*}
\begin{equation*}
    \lambda_j \ge 0, \qquad \mu_j \ge 0 \qquad \forall j \in [h] \qquad \text{(Dual Feasibility)}
\end{equation*}
\begin{equation*}
       \lambda_j (z_t - z_j - \epsilon) = 0 \qquad \forall j \in [h] \setminus \{t\} \qquad \mu_j z_j = 0 \qquad \forall j \in [h] \qquad \text{(Complementary Slackness)}
\end{equation*}

Let
\begin{itemize}
\item $\clZ^t = \{j \in [h] : \mu_j \not = 0\} $ be the set of categories for which the zero constraint is active,
    \item $\clM^t = \{j \in [h] \setminus t : \lambda_j \not = 0\}$ be the set of categories for which the margin constraint is active,  and
    \item $\clF^t = [h] \setminus (\clM^t \cup \clZ^t \cup \{t\})$ the free set where neither of the constraints is active.
\end{itemize}
  Notice that these sets, together with $\{t\}$, form a partition of $[h]$. 
  
  Indeed, the primal feasibility conditions impose that  $\bfz \in \Delta([h])$ and $z_t - z_j \ge \epsilon$ for all $j \neq t$. If $z_t = \epsilon$, then for all $j \in [h] \setminus \{t\}$ we would have
\[
z_j \le z_t - \epsilon = 0.
\]
Since $z_j \ge 0$ for all $j \in [h]$ and $\epsilon < 1$, it follows that $z_j = 0$ for all $j \neq t$. 
Therefore,
\[
\sum_{j=1}^h z_j = z_t = \epsilon \neq 1,
\]
which contradicts the simplex constraint $\bfz \in \Delta$. 
Hence $z_t > \epsilon$, and in particular $z_t - \epsilon > 0$.

It follows that no index $j \neq t$ can simultaneously satisfy
\[
z_j = 0
\quad\text{and}\quad
z_j = z_t - \epsilon,
\]
and hence $\clZ^t \cap \clM^t = \emptyset$.

Then: 
\begin{itemize}
    \item if $j \in \clZ^t$ then $z_j=0$. We know that  $\lambda_j =0$ and  we can use the  stationarity condition together with the dual feasibility to identify the coordinates belonging to this set as: 
    \begin{equation}\label{app:tau_ge_zero}
        \mu_j = \nu  - 2\hat y_j \ge 0 \qquad   \hat y_j \le \tau,
    \end{equation}
    where $\tau = \frac{\nu}{2}$.
        \item if $j \in \clF^t$ then, following the stationarity condition, $z_j = \hat y_j - \frac{\nu}{2}$.
    \item if $j \in \clM^t$ then $z_j = z_t - \epsilon$. We know that $\mu_j =0$, and we can use the stationarity condition to get: 
     \begin{equation*}
        \lambda_j = 2 (\hat y_j - z_j) - \nu = 2 (\hat y_j - z_t + \epsilon) - \nu = 2(\hat y_j - z_t + \epsilon - \tau)
    \end{equation*}
\end{itemize}
Since we know that for all $j \in \clZ^t \cup \clF^t$, $\lambda_j=0$ we can rewrite the first stationarity condition as:
    \begin{equation*}
        2(z_t - \hat y_t) + \nu - \sum_{j \in \clM^t} \lambda_j = 0
    \end{equation*}
    \begin{equation*}
        2(z_t - \hat y_t) + \nu - \sum_{j \in \clM^t} 2(\hat y_j - z_t + \epsilon - \tau)  = 0
    \end{equation*}

Let $m = |\clM^t|$,
    \begin{equation*}
        (z_t - \hat y_t) + \tau - \sum_{j \in \clM^t} \hat y_j + m z_t - m\epsilon + m\tau  = 0
    \end{equation*}
    \begin{equation}\label{app:stationarity_zt}
        (m+1)z_t = {\hat y_t + \sum_{j \in \clM^t} \hat y_j + m\epsilon} - (m+1)\tau  
    \end{equation}

To determine $\tau$, we impose the simplex constraint $\sum_{j=1}^h z_j = 1$ and decompose the sum according to the partition $\{\mathcal Z^t, \mathcal F^t, \mathcal M^t\}$ of $[h]\setminus\{t\}$. By construction,
\[
z_j =
\begin{cases}
0, & j \in \mathcal Z^t,\\
\hat y_j - \tau, & j \in \mathcal F^t,\\
z_t - \epsilon, & j \in \mathcal M^t,
\end{cases}
\qquad \text{and} \qquad z_t \text{ for } j = t.
\]
Hence,
\[
1
=
z_t
+
\sum_{j \in \mathcal F^t} (\hat y_j - \tau)
+
\sum_{j \in \mathcal M^t} (z_t - \epsilon).
\]

Let $f = |\mathcal F^t|$ and $m = |\mathcal M^t|$. Then
\begin{equation}\label{app:application_of_simplex_constraint}
    1
=
(m+1) z_t
+
\sum_{j \in \mathcal F^t} \hat y_j
-
f \tau
-
m \epsilon \qquad \qquad (m+1) z_t = 1 - \sum_{j \in \mathcal F^t} \hat y_j + f\tau + m\epsilon
\end{equation}

By setting \ref{app:stationarity_zt} equal to \ref{app:application_of_simplex_constraint} we obtain 
\begin{equation*}
{\hat y_t + \sum_{j \in \clM^t} \hat y_j + m\epsilon} - (m+1)\tau = 1 - \sum_{j \in \mathcal F^t} \hat y_j + f\tau + m\epsilon
\end{equation*}
\begin{equation}
    \sum_{j \in \clZ^t}\hat y_j = - (h-|\clZ^t|)\tau
\end{equation}
Since by definition $\sum_{j\in \clZ^t} \hat y_j \geq 0$ and $h - |\clZ^t| \geq 1$ (as $t \notin \clZ^t$), Equation~(20) gives $\tau \leq 0$.
However, according to Equation~\eqref{app:tau_ge_zero}, we have that for all $j \in \clZ^t$, $0 \le \hat y_j \le \tau $. Hence, for every $j \in \clZ^t$, $\hat y_j = 0$, and  $\tau =\nu =0$.

Since $\tau = 0$, we have that for all $j \not \in \clM^t \cup \{t\}$, $z_j = \hat y_j$, hence the sets $\clF^t$ and $\clZ^t$ are indistinguishable. Hence, we have that: 
\begin{equation}\label{eq:final_zt_app}
    z_j = \begin{cases}
        \hat y_j, &\quad j \not \in \clM^t \cup \{t\}, \\
        z_t - \epsilon, & \quad j \in \clM^t, \\
        \frac{\hat y_t + \sum_{j \in \clM^t} \hat y_j + m\epsilon}{m+1} & \quad j = t
    \end{cases}
\end{equation}

Notice that Equations~\eqref{app:eq:active_margin_set}-\eqref{eq:final_zt_app} define a fixed-point system. Indeed, the set $\mathcal{M}^t$:
\begin{equation}\label{app:eq:active_margin_set}
    \clM^t = \{j \in [h]\setminus\{t\} \,:\, \hat y_j \ge z_t^t - \epsilon \}
\end{equation}
is defined through $z_t^t$, which in turn depends on
$\mathcal{M}^t$. We now show that this system admits a unique solution,
which can be found by a one-dimensional sweep over the sorted values of
$\hat{\mathbf{y}}$.

Let $\hat y^{(1)} \geq \hat y^{(2)} \geq \cdots \geq \hat y^{(h-1)}$ denote the
values $\{\hat y_j : j \neq t\}$ sorted in decreasing order, with the convention
$\hat y^{(0)} := +\infty$ and $\hat y^{(h)} := -\infty$. For $a \in \{0, 1, \ldots, h-1\}$,
define the \emph{candidate threshold}
\begin{equation}\label{eq:theta_def}
\theta_a \;=\; \frac{\hat y_t + \sum_{k=1}^{a} \hat y^{(k)} - \epsilon}{1 + a}.
\end{equation}
By construction, $\theta_a = z_t^t - \epsilon$ whenever $|\mathcal{M}^t| = a$ and
$\mathcal{M}^t$ consists of the top-$a$ indices.

To prove the uniqueness of the solution, we prove that there exists a unique integer $a^\star \in \{0, 1, \ldots, h-1\}$ such that
\begin{equation}\label{eq:selection}
\hat y^{(a^\star)} \,>\, \theta_{a^\star} \qquad \text{and} \qquad \hat y^{(a^\star+1)} \,\leq\, \theta_{a^\star},
\end{equation}
and that setting $\mathcal{M}^t$ to be the top-$a^\star$ indices of $\{\hat y_j : j \neq t\}$
and $z_t^t = \theta_{a^\star} + \epsilon$ yields the unique solution of the
fixed-point system \eqref{app:eq:active_margin_set}–\eqref{eq:final_zt_app}.

Indeed, from \eqref{eq:theta_def}:
\begin{equation}\label{eq:theta_recurrence}
\theta_{a+1} - \theta_a \;=\; \frac{\hat y^{(a+1)} - \theta_a}{2 + a},
\end{equation}
which implies the equivalence
\begin{equation}\label{eq:monotonicity}
\theta_{a+1} \,>\, \theta_a \quad \Longleftrightarrow \quad \hat y^{(a+1)} \,>\, \theta_a,
\end{equation}
and similarly for the reverse inequality.

\emph{Existence.} Define
\[
a^\star \; =\; \max\bigl\{\, a \in \{0, 1, \ldots, h-1\} \,:\, \hat y^{(a)} > \theta_{a-1} \,\bigr\},
\]
where the set is nonempty because $\hat y^{(0)} = +\infty > \theta_{-1}$ with the
convention $\theta_{-1} = -\infty$. By definition, $\hat y^{(a^\star)} > \theta_{a^\star - 1}$,
and applying \eqref{eq:monotonicity} at index $a^\star - 1$ yields $\theta_{a^\star} > \theta_{a^\star - 1}$.
Rearranging \eqref{eq:theta_recurrence} gives
\[
\theta_{a^\star} \;=\; \frac{a^\star \theta_{a^\star - 1} + \hat y_{(a^\star)}}{1 + a^\star},
\]
which is a convex combination of $\theta_{a^\star - 1}$ and $\hat y_{(a^\star)}$; since
$\theta_{a^\star - 1} < \hat y^{(a^\star)}$, we obtain%
 $\hat y^{(a^\star)} > \theta_{a^\star}$, the first part of \eqref{eq:selection}.
If $a^\star = h-1$, the second part of \eqref{eq:selection} holds vacuously via
$\hat y^{(h)} = -\infty$; otherwise, maximality of $a^\star$ gives $\hat y^{(a^\star + 1)} \leq \theta_{a^\star}$.

\emph{Uniqueness.} Suppose $a_1 < a_2$ both satisfy~\eqref{eq:selection}. Hence, $\hat y^{(a_1+1)} \le \theta_{a_1}$. 
We show by induction that
\begin{equation}
  \hat y^{(b)} \le \theta_b \qquad \text{for every } b \ge a_1 + 1,
  \label{eq:uniqueness-claim}
\end{equation}
which contradicts the strict inequality $\hat y^{(a_2)} > \theta_{a_2}$ required 
by~\eqref{eq:selection} at $a_2$.

Rearranging~\eqref{eq:theta_recurrence} yields the identity
\begin{equation}
  \theta_{b+1} - \hat y^{(b+1)} 
  \;=\; \frac{1+b}{2+b}\,\bigl(\theta_b - \hat y^{(b+1)}\bigr).
  \label{eq:theta-yhat-gap}
\end{equation}

\emph{Base case} ($b = a_1 + 1$). Applying~\eqref{eq:theta-yhat-gap} at index $a_1$,
\[
  \theta_{a_1+1} - \hat y^{(a_1+1)} 
  \;=\; \frac{1+a_1}{2+a_1}\bigl(\theta_{a_1} - \hat y^{(a_1+1)}\bigr) \;\ge\; 0
\]
by the hypothesis $\hat y^{(a_1+1)} \le \theta_{a_1}$.

\emph{Inductive step.} Assume $\theta_b \ge \hat y^{(b)}$. Since $\hat y^{(\cdot)}$ 
is non-increasing, $\hat y^{(b+1)} \le \hat y^{(b)} \le \theta_b$, so the 
right-hand side of~\eqref{eq:theta-yhat-gap} is non-negative, giving 
$\theta_{b+1} \ge \hat y^{(b+1)}$.

Iterating up to $b = a_2$ gives $\hat y^{(a_2)} \le \theta_{a_2}$, the desired 
contradiction.

\emph{Consistency with the fixed-point system.} With $a^\star$ as above, set
$\mathcal{M}^t$ to be the top-$a^\star$ indices and $z_t^t = \theta_{a^\star} + \epsilon$.
Then \eqref{eq:final_zt_app} holds by construction of $\theta_{a^\star}$, and
\eqref{app:eq:active_margin_set} holds by \eqref{eq:selection}: every index in
$\mathcal{M}^t$ has $\hat y_j \geq \hat y_{(a^\star)} > \theta_{a^\star} = z_t^t - \epsilon$,
and every index outside has $\hat y_j \leq \hat y_{(a^\star + 1)} \leq \theta_{a^\star} = z_t^t - \epsilon$. 
\end{proof}

\paragraph{Computing $\bfz^t$.} Given an unconstrained output $\hat \bfy$ and a constraint stating that the final assignment must belong to $\clS$, for every $t \in \clS$ we can compute $\bfz^t$ by modifying the sorting method used to project onto the probabilistic simplex~\cite{held1974_proj_simplex,Gafni1984TWOMETRICPM,duchi2008_projections}:
\begin{enumerate}
  \item Sort $\{\hat y_j : j \neq t\}$ in decreasing order as $\hat y_{1} \geq \cdots \geq \hat y_{h-1}$.
  \item Compute the cumulative sums and evaluate $\theta_a$ for $a = 0, 1, \ldots, h-1$ via \eqref{eq:theta_def} in $\mathcal{O}(h)$ time.
  \item Find the unique $a^\star$ satisfying \eqref{eq:selection} (equivalently, the largest $a$ with $\hat y_{(a)} > \theta_{a-1}$).
  \item Set $\mathcal{M}^t$ to be the corresponding top-$a^\star$ indices and return $\mathbf{z}^t$ via \eqref{eq:fixed_point_system} with $z_t^t = \theta_{a^\star} + \epsilon$.
\end{enumerate}
This algorithm needs to perform $\mathcal{O}(h \log h)$ sequential operations, 
dominated by the sort, while it requires only $\mathcal{O}(\log^2 h)$ 
sequential GPU operations thanks to parallelization. Alternatively, the 
convex problem can be solved with a generic 
solver (like~\citep{diamond2016cvxpy}), applying an iterative 
interior-point or ADMM method.

\section{Proof of Lemma~\ref{lem:overapproximation}}\label{app:proof_overapproximation_lemma}

\paragraph{Statement.}
{\sl For every $t \in [h]$ and every $\hat{\mathbf{y}} \in \Delta([h])$, $\mathcal{M}^t \subseteq \mathcal{A}^t.$}

\begin{proof}
Let $j \in \mathcal{M}^t$. From $\mathcal{M}^t$ definition we know that 
$\hat y_j \ge z_t^t - \epsilon$. To show that $\mathcal{M}^t \subseteq \mathcal{A}^t.$  we need to demonstrate that $z_t^t \geq \hat y_t$.

Using the closed form \eqref{eq:final_zt_app} and writing $m = |\mathcal{M}^t|$,
\begin{equation}\label{eq:zt_minus_yt}
z_t^t - \hat y_t
\;=\; \frac{\hat y_t + \sum_{k \in \mathcal{M}^t} \hat y_k + m\,\epsilon}{1 + m} - \hat y_t
\;=\; \frac{\sum_{k \in \mathcal{M}^t} \hat y_k - m\,\hat y_t + m\,\epsilon}{1 + m}.
\end{equation}
If $m = 0$, the right-hand side vanishes and $z_t^t = \hat y_t$. 

If $m \geq 1$,
each $k \in \mathcal{M}^t$ satisfies $\hat y_k \ge z_t^t - \epsilon$ yielding $\sum_{k \in \mathcal{M}^t} \hat y_k \ge m\,(z_t^t - \epsilon)$. Substituting into
\eqref{eq:zt_minus_yt},
\[
z_t^t - \hat y_t
\ge \frac{m\,(z_t^t - \epsilon) - m\,\hat y_t + m\,\epsilon}{1 + m}
\ge \frac{m}{1+m}\,(z_t^t - \hat y_t),
\]
which rearranges to \ $z_t^t \ge \hat y_t$.

\end{proof}

\section{Proof of Theorem~\ref{th:constraint_satisfaction_basic_case}}\label{app:proof_constraint_satisfaction_basic_case}

\paragraph{Statement.} {\sl Let $m_\theta:\clX \to \Delta([h])$, $\bfx \in \clX$ and $\hat \bfy = m_\theta(\bfx)$. Let $\Pi = \{\clS : Y\}$ for some $\emptyset \subset \clS \subseteq [h]$. If 
    \begin{equation}\label{eq:closed_form_solution_approximated_app}
        \layer(\hat \bfy) = \tilde \bfz^{t^\star}, \qquad t^\star = \argmin_{t\in \clS}\|\tilde \bfz^t - \hat \bfy\|_2^2, 
    \end{equation}
    where for every $t \in \clS$
    \begin{equation}
        \tilde z^t_j = \begin{cases}
        \hat y_j & \text{if } j \not\in (\clA^t \cup \{t\})\\
        \tilde z_t^t  - \epsilon & \text{if }  j \in \clA^t \qquad \qquad \quad \clA^t = \{j \in [h] \setminus \{t\} : \hat y_j \ge \hat y_t - \epsilon\}, \\
        \frac{
\hat y_t + \sum_{j\in \mathcal A^t} \hat y_j + |\mathcal A^t|\epsilon
}{
1+|\mathcal A^{t}|} & \text{if } j = t
\end{cases}
    \end{equation}
then for $\epsilon \to 0^+$, the assignment $\bfa$ induced by $\layer(\hat \bfy)$ satisfies $\Pi$, i.e., $\bfa \models \Pi$.}

\begin{proof}

We show that for every $t \in \clS$, the surrogate $\tilde\bfz^t$ defined by
Eq.~(\ref{eq:closed_form_solution_approximated_app})               lies in the simplex
$\Delta([h])$ and has $t$ as its unique argmax. Applied to $t = t^\star$,
this gives $\bfa = \argmax_j \tilde z^{t^\star}_j = t^\star \in \clS$,
hence $\bfa \models \Pi$.

Let $t \in \clS$. We first show that $\tilde \bfz^t \in \Delta([h])$:
$$
\begin{aligned}
    \sum_{j \in [h]}\tilde z_j^t & = \sum_{j \not\in \clA^t \cup \{t\}} \hat y_j + \sum_{j \in \clA^t}(\tilde z_t^t - \epsilon) + \tilde z_t^t \\
    & = \sum_{j \not\in \clA^t \cup \{t\}} \hat y_j +(|\clA^t| +1) \frac{
\hat y_t + \sum_{j\in \mathcal A^t} \hat y_j + |\mathcal A^t|\epsilon
}{
1+|\mathcal A^{t}|} - |\clA^t|\epsilon \\
 & = \sum_{j \in [h]} \hat y_j =  1,
\end{aligned}
$$
Non-negativity of $\tilde z^t_j$ follows from $\hat\bfy \in \Delta([h])$ for the
unmodified entries, from $\tilde z^t_t$ being a weighted average of non-negative
quantities, and from $\tilde z^t_t - \epsilon \geq \hat y_t - \epsilon \geq 0$ on
$\clA^t$ provided $\epsilon \leq \hat y_t$, which holds for $\epsilon$ sufficiently small.
Hence $\tilde\bfz^t \in \Delta([h]).$ 

To prove $t = \argmax_j \tilde z^t_j$, we compare $\tilde z^t_t$ against the elements in both $\clA^t$ and
$[h] \setminus (\clA^t \cup \{t\})$.

For $j \in \clA^t$: $\tilde z^t_j = \tilde z^t_t - \epsilon < \tilde z^t_t$ for any $\epsilon > 0$.

For $j \notin \clA^t \cup \{t\}$: $\tilde z^t_j = \hat y_j < \hat y_t - \epsilon$ by
definition of $\clA^t$. Since every $k \in \clA^t$ satisfies $\hat y_k \geq \hat y_t - \epsilon$,
$\sum_{k \in \clA^t} \hat y_k \geq |\clA^t|(\hat y_t - \epsilon)$, and substituting into
the closed form for $\tilde z^t_t$ yields $\tilde z^t_t \geq \hat y_t$. Combining,
$\tilde z^t_t \geq \hat y_t > \hat y_t - \epsilon > \hat y_j = \tilde z^t_j$.
\end{proof}

\section{Proof of Theorem~\ref{thm:argmax}}
\label{app:thm:argmax}

\paragraph{Statement.} {\sl Let $\hat{\mathbf y}\in\Delta([h])$, $\bfz^t $ be the solution of the fixed point system \eqref{eq:fixed_point_system} and $\tilde\bfz^t$ its surrogate solution as defined in Equation~\eqref{eq:closed_form_solution_approximated}. Then, for
$\epsilon \to 0^+$,  
$$
\argmin_{t\in \clS}
\left\|
\mathbf z^t-\hat{\mathbf y}
\right\|_2^2
=
\argmin_{t\in \clS}
\left\|
\widetilde{\mathbf z}^{\,t}-\hat{\mathbf y}
\right\|_2^2
=
\argmax_{t\in \clS}\hat y_t.
$$}

\begin{proof}
Notice that if $|\clS| =1$ the thesis is trivially satisfied. We hence assume $|\clS| > 1$. We prove separately that the surrogate and exact projection criteria
are strictly improved whenever a candidate $t$ is replaced by a
candidate $t'$ satisfying $\hat y_{t'}>\hat y_t$; the case $\hat y_{t'} = \hat y_t$ is
treated at the end of the proof.

\paragraph{Exact fixed-point solution.}
Suppose by contradiction that
$t = \argmin_{s\in\clS} \|\bfz^s - \hat{\bfy}\|_2^2$ satisfies
$\hat y_t < \max_{s\in\clS} \hat y_s$. Then there exists $t' \in \clS$
such that $\hat y_{t'} > \hat y_t$.

Given $s \in \clS$, let $C^s = \{\bfz \in \Delta([h]) : z_s \geq z_j + \epsilon,\ \forall j \neq s\}$, be the {\sl $s$-selection region}, i.e., the subset of $\Delta([h])$ in which coordinate $s$ is maximal with margin at least $\epsilon$.

Consider an alternative candidate solution to the fixed point system $\overline{\mathbf z}^{\,t'}\in \clC^{t'}$ constructed from $\bfz^t$ by swapping the values at the coordinates $t$ and $t'$:
\[
\overline z^{\,t'}_j
=
\begin{cases}
z^t_{t'},
&
j=t,
\\
z^t_t,
&
j=t',
\\
z^t_j,
&
j\notin\{t,t'\}.
\end{cases}
\]
We know that $\overline{\mathbf z}^{\,t'}\in \clC^{t'}$ because by definition of $\bfz^t$,
$
z^t_t\geq z^t_j+\epsilon,
\text{for every }j\neq t.
$
For every $j\not\in\{ t', t\}$, this implies
\[
\overline z^{\,t'}_{t'}
=
z^t_t
\geq
z^t_j+\epsilon
=
\overline z^{\,t'}_j+\epsilon, \qquad \text{and}\qquad 
\overline z^{\,t'}_{t'}
=
z^t_t
\geq
z^t_{t'}+\epsilon
=
\overline z^{\,t'}_t+\epsilon.
\]

Since only coordinates $t$ and $t'$ differ between
$\overline{\mathbf z}^{\,t'}$ and $\mathbf z^t$:
\begin{align}
&
\|
\overline{\mathbf z}^{\,t'}-\hat{\mathbf y}
\|_2^2
-
\|
\mathbf z^t-\hat{\mathbf y}
\|_2^2
\nonumber\\
&=
\left(z^t_{t'}-\hat y_t\right)^2
+
\left(z^t_t-\hat y_{t'}\right)^2
-
\left(z^t_t-\hat y_t\right)^2
-
\left(z^t_{t'}-\hat y_{t'}\right)^2
\nonumber\\
&=
2\left(z^t_t-z^t_{t'}\right)
\left(\hat y_t-\hat y_{t'}\right).
\label{eq:fixed-point-swap-distance}
\end{align}
Because
\[
z^t_t-z^t_{t'}\geq\epsilon>0
\]
and
\[
\hat y_t-\hat y_{t'}<0,
\]
Equation~\eqref{eq:fixed-point-swap-distance} implies
\[
\|
\overline{\mathbf z}^{\,t'}-\hat{\mathbf y}
\|_2^2
<
\|
\mathbf z^t-\hat{\mathbf y}
\|_2^2.
\]
Let $\mathbf z^{t'}$ be the minimizer over $\mathcal C^{t'}$, since
$\overline{\mathbf z}^{\,t'}\in\mathcal C^{t'}$,
\[
\|
\mathbf z^{t'}-\hat{\mathbf y}
\|_2^2
\leq
\|
\overline{\mathbf z}^{\,t'}-\hat{\mathbf y}
\|_2^2
<
\|
\mathbf z^t-\hat{\mathbf y}
\|_2^2,
\]
contradicting the assumed optimality of $t$.

Thus, no index with a strictly sub-maximal value of $\hat y_t$ can
 minimize the exact projection distance.

\paragraph{Surrogate solution.}
For every $t\in \clS$ and $j\in \clA^t$, let
\[
r_j^t
=
\hat y_j-\hat y_t+\epsilon
\qquad \text{and} \qquad
\alpha_t
=
\tilde z^{t}_t-\hat y_t
=
\frac{1}{1+|\clA^t|}
\sum_{j\in \clA^t}r_j^t,
\]
$r_j^t$ representing the $\epsilon$-{\sl margin violation for coordinate $j$}.
For $j\in \clA^t$, we have
\[
\tilde z^{t}_j-\hat y_j
=
\tilde z^{t}_t-\epsilon-\hat y_j
=
\alpha_t-r_j^t,
\]
while the remaining coordinates are unchanged. Hence,
$$
\begin{aligned}
\|
\tilde{\mathbf z}^{t}-\hat{\mathbf y}
\|_2^2
& =
\alpha_t^2
+
\sum_{j\in \clA^t}
\left(\alpha_t-r_j^t\right)^2 \\
& = \alpha_t^2 + |\clA^t|\alpha^2_t - \sum_{j \in \clA^t} 2\alpha_t r_j^t + \sum_{j \in \clA^t} (r_j^t)^2 \\
& = \frac{1}{1+|\clA^t|}\left(\sum_{j \in \clA^t} r_j^t\right)^2 - \frac{2}{1+|\clA^t|}\left(\sum_{j \in \clA^t} r_j^t\right)^2+ \sum_{j \in \clA^t} (r_j^t)^2 \\
& =
\sum_{j\in \clA^t}(r_j^t)^2
-
\frac{1}{1+|\clA^t|}
\left(
\sum_{j\in \clA^t}r_j^t
\right)^2.
\label{eq:surrogate-distance-residual}
\end{aligned}
$$

The above equation is equivalent to the computational formula of the sum of squares over the multiset $\clQ^t =
 \{r^t_j : j \in \clA^t\} \uplus \{r^t_t\} 
$ with $r_t^t=0$, defined as:
$$
\operatorname{SS}(\clQ^t)
=
\sum_{q\in \clQ^t}(q-\overline q)^2.
$$
$\clQ^t$ can be interpreted as the multiset of displacement terms associated with the coordinates modified by the surrogate (i.e., those in $\clA^t$ and $t$) when removing the shared displacement $\tilde z_t^t - \hat y_t$.

Consider \(t,t'\in \clS\) such that
\[
\hat y_{t'}>\hat y_t,
\qquad
d =\hat y_{t'}-\hat y_t>0.
\]
Since \(\epsilon\to0^+\), we may assume that \(\epsilon<d\). The corresponding active sets satisfy
\[
\clA^{t'}\subseteq \clA^t,
\qquad
t\notin \clA^{t'},
\qquad
t'\in \clA^t.
\]
Additionally, for every \(j\in \clA^{t'}\),
\[
r_j^t
=
\hat y_j-\hat y_t+\epsilon
=
r_j^{t'}+d,
\]
while
\[
r_{t'}^t=d+\epsilon.
\]

Let
\[
\clP =\clQ^{t'}+d
=
\{r_j^{t'}+d:j\in \clA^{t'}\}\uplus\{d\},
\]
where \(\clQ^{t'}+d\) denotes the multiset obtained by adding \(d\)
to each element of \(\clQ^{t'}\). Since the sum of squared deviations
is translation invariant,
\[
\operatorname{SS}(\clP)=\operatorname{SS}(\clQ^{t'}).
\]

We next relate \(\clP\) to \(\clQ^t\) by introducing a new multiset $\clR$ obtained by removing the element \(d\in \clP\)
and adding the two elements \(0\) and \(d+\epsilon\):
$$
\clR=(\clP\setminus\{d\})\uplus \{0,d+\epsilon\}.
$$
The element \(0\) is the residual associated with the candidate
coordinate \(t\), whereas \(d+\epsilon=r_{t'}^t\). The goal is to show that 
$$
\operatorname{SS}(\clQ^t)
\ge
\operatorname{SS}(\clR)
>
\operatorname{SS}(\clP)
=
\operatorname{SS}(\clQ^{t'}).
$$

Let \(n=|\clP|\) and \(s=\sum_{q\in \clP}q\). Since $d \in \clP$  
and all its other elements are at least \(d\), we have \(s\ge nd\).
Using the computational formula of the sum of squares
we obtain
\[
\begin{aligned}
\operatorname{SS}(\clR)-\operatorname{SS}(\clP)
&=
(d+\epsilon)^2 - d^2 
-\frac{(s+\epsilon)^2}{n+1}
+\frac{s^2}{n} \\
&=
\frac{
s^2-2n\epsilon s+2n(n+1)d\epsilon+n^2\epsilon^2
}{
n(n+1)
}.
\end{aligned}
\]
The numerator is increasing in $s$ for $s \geq n\epsilon$, and
$s \geq nd > n\epsilon$ since $d > \epsilon$; hence it is minimised at
$s = nd$, giving
\[
\operatorname{SS}(\clR)-\operatorname{SS}(\clP)
\ge
\frac{nd^2+2d\epsilon+n\epsilon^2}{n+1}
>0.
\]

Finally, \(\clQ^t\) is obtained from \(\clR\) by adding the residuals
\[
\{r_j^t:
j\in \clA^t\setminus(\clA^{t'}\cup\{t'\})\}.
\]
Adding an element to a multiset cannot decrease its sum of squared
deviations: if \(\clQ\) has cardinality \(m\) and mean \(\overline q\),
then
\[
\operatorname{SS}(\clQ\uplus\{x\})
=
\operatorname{SS}(\clQ)
+
\frac{m}{m+1}(x-\overline q)^2
\ge
\operatorname{SS}(\clQ).
\]
Consequently,
\[
\operatorname{SS}(\clQ^t)
\ge
\operatorname{SS}(\clR)
>
\operatorname{SS}(\clP)
=
\operatorname{SS}(\clQ^{t'}).
\]
Equivalently,
\[
\|\widetilde{\mathbf z}^{\,t'}-\widehat{\mathbf y}\|_2^2
<
\|\widetilde{\mathbf z}^{\,t}-\widehat{\mathbf y}\|_2^2.
\]
Thus, no candidate \(t\) with a strictly sub-maximal value
\(\hat y_t\) can minimise the surrogate projection distance.

Notice that if $\hat y_t=\hat y_{t'}$, exchanging the coordinates $t$ and $t'$
leaves both projection distances unchanged. Hence,
\[
\|\mathbf z^t-\hat{\mathbf y}\|_2^2
=
\|\mathbf z^{t'}-\hat{\mathbf y}\|_2^2
\quad\text{and}\quad
\|\tilde{\mathbf z}^{\,t}-\hat{\mathbf y}\|_2^2
=
\|\tilde{\mathbf z}^{\,t'}-\hat{\mathbf y}\|_2^2.
\]
The same candidate is therefore selected by all three operators
according to the common deterministic tie-breaking policy specified
in Footnote~1.

Putting our results together we obtain that:
$$
\argmin_{t\in \clS}
\left\|
\mathbf z^t-\hat{\mathbf y}
\right\|_2^2
=
\argmin_{t\in \clS}
\left\|
\tilde{\mathbf z}^t-\hat{\mathbf y}
\right\|_2^2
=
\argmax_{t\in \clS}\hat y_t.
$$

\end{proof}

\section{Proof of Theorem~\ref{thm:exact_error}}\label{app:proof_exact_error}

\paragraph{Statement.}  {\sl Let $\mathbf{z}^t$ be the solution of the fixed point system \eqref{eq:fixed_point_system}, $\tilde{\mathbf{z}}^t$ defined as in Equation~\eqref{eq:closed_form_solution_approximated}, and $\mathcal{W}^t = \mathcal{A}^t \setminus \mathcal{M}^t$. For each $j \in \mathcal{W}^t$, let $g_j = z_t^t - \epsilon - \hat y_j$.
Then
\begin{equation}
\bigl\| \tilde{\mathbf{z}}^t - \mathbf{z}^t \bigr\|_2^2
= \sum_{j \in \mathcal{W}^t} g_j^{2}
- \frac{1}{1 + |\mathcal{A}^t|}\Bigl(\sum_{j \in \mathcal{W}^t} g_j\Bigr)^{\!2}.
\end{equation}}

\begin{proof}
Let $\mathcal{W}^t = \mathcal{A}^t \setminus \mathcal{M}^t$ denote the
spuriously included indices, with cardinality $w = |\mathcal{W}^t|$. Let
$m = |\mathcal{M}^t|$, $a = |\mathcal{A}^t| = m + w$. Write
$M = \sum_{k \in \mathcal{M}^t}\hat y_k$ and $G = \sum_{j \in \mathcal{W}^t} g_j$,
where $g_j = z_t^t - \epsilon - \hat y_j$ for $j \in \mathcal{W}^t$.

The two solutions read
\begin{equation*}
z_t^t = \frac{\hat y_t + M + m\epsilon}{1 + m},
\qquad
\tilde z_t^t = \frac{\hat y_t + M + \sum_{k \in \mathcal{W}^t}\hat y_k + a\epsilon}{1 + a},
\end{equation*} 
$$
z_j^t = \tilde z_j^t = \hat y_j \text {\qquad for } j \notin \mathcal{A}^t \cup \{t\},$$
$$
z_j^t = z_t^t - \epsilon \qquad 
\tilde z_j^t= \tilde z_t^t - \epsilon
\text{ \qquad for } j \in \mathcal{M}^t
$$ 
$$ 
z_j^t = \hat y_j \qquad \tilde z_j^t = \tilde z_t^t - \epsilon \text{\qquad 
for } j \in \mathcal{W}^t.
$$

We now proceed in calculating the difference for each coordinate. 

Let us consider the difference at $j=t$.
Since $\hat y_t + M = (1+m)z_t^t - m\epsilon$ we can simplify the expression for $\tilde z_t^t$ as
\begin{align*}
\tilde z_t^t - z_t^t
&= \frac{(1+m)z_t^t - m\,\epsilon + \sum_{k \in \mathcal{W}^t}\hat y_k + a\,\epsilon}{1+a} - z_t^t \\[2pt]
&= \frac{(1+m)z_t^t - z_t^t(1+a) + \sum_{k \in \mathcal{W}^t}\hat y_k + w\,\epsilon}{1+a} \\
&= \frac{-wz_t^t + \sum_{k \in \mathcal{W}^t}\hat y_k + w\epsilon}{1+a} \\
&= -\frac{\sum_{k \in \mathcal{W}^t}(z_t^t - \epsilon - \hat y_k)}{1+a}
= -\frac{G}{1+a}.
\end{align*}

For $j \notin \mathcal{A}^t \cup \{t\}$ the difference is zero. 

For
$j \in \mathcal{M}^t$,  
\begin{equation*}
\tilde z_j^t - z_j^t
= \bigl[\tilde z_t^t - \epsilon\bigr] - \bigl[z_t^t - \epsilon\bigr]
= \tilde z_t^t - z_t^t
= -\frac{G}{1+a}.
\end{equation*}
For $j \in \mathcal{W}^t$,
\begin{equation*}
\tilde z_j^t - z_j^t
= \tilde z_t^t - \epsilon - \hat y_j
= \bigl[\tilde z_t^t - z_t^t\bigr] + \bigl[z_t^t - \epsilon - \hat y_j\bigr]
= g_j - \frac{G}{1+a}.
\end{equation*}

Summing squared differences 
(over $\{t\}$, $\mathcal{M}^t$, and $\mathcal{W}^t$):
\begin{align*}
\bigl\| \tilde{\mathbf{z}}^t - \mathbf{z}^t \bigr\|_2^2
&= \frac{G^{\,2}}{(1+a)^{\,2}}
+ m \cdot \frac{G^{\,2}}{(1+a)^{\,2}}
+ \sum_{j \in \mathcal{W}^t}\Bigl(g_j - \frac{G}{1+a}\Bigr)^{\!2} \\
&= \frac{(1+m)\,G^{\,2}}{(1+a)^{\,2}}
+ \sum_{j \in \mathcal{W}^t} g_j^{\,2}
- \frac{2\,G}{1+a}\sum_{j \in \mathcal{W}^t} g_j
+ \frac{w\,G^{\,2}}{(1+a)^{\,2}}.
\end{align*}
Using $\sum_{j \in \mathcal{W}^t} g_j = G$ and $m + w = a$. The total simplifies to
\begin{equation*}
\bigl\| \tilde{\mathbf{z}}^t - \mathbf{z}^t \bigr\|_2^2
=\sum_{j \in \mathcal{W}^t} g_j^{\,2}
- \frac{2\,G^{\,2}}{1+a}
+ \frac{G^{\,2}}{1+a}
= \sum_{j \in \mathcal{W}^t} g_j^{\,2}
- \frac{G^{\,2}}{1+a},
\end{equation*}
which is Equation~\eqref{eq:exact_identity}.

\end{proof}

\section{Proof of Corollary~\ref{cor:L2_bound}}

\paragraph{Statement.} {\sl Under the hypotheses of Theorem~\ref{thm:exact_error},
\begin{equation}\label{eq:L2_bound_app}
\bigl\| \tilde{\mathbf{z}}^t - \mathbf{z}^t \bigr\|_2 \leq  \bigl\| \tilde\bfz^t - \hat{\mathbf{y}} \bigr\|_2.
\end{equation}}

\begin{proof}
We first show that $\mathbf z^t$ is the orthogonal projection of $\hat{\mathbf y}$
onto the affine subspace
\[
V_{\mathcal{M}^t} = \bigl\{\mathbf z \in \mathbb R^h : z_j = z_t - \epsilon,\ j \in \mathcal{M}^t\bigr\}.
\]
Substituting the defining constraint into the objective, the projection problem
$\min_{\mathbf z \in V_{\mathcal{M}^t}} \tfrac12\|\mathbf z - \hat{\mathbf y}\|_2^2$
reduces to
\[
\min_{z_t,\,\{z_j\}_{j \notin \mathcal{M}^t \cup \{t\}}}\;
\tfrac12(z_t - \hat y_t)^2
+ \tfrac12\!\sum_{j \in \mathcal{M}^t}\!(z_t - \epsilon - \hat y_j)^2
+ \tfrac12\!\sum_{j \notin \mathcal{M}^t \cup \{t\}}\!(z_j - \hat y_j)^2.
\]
The free coordinates decouple: $z_j = \hat y_j$ for $j \notin \mathcal{M}^t \cup \{t\}$,
and the first-order condition in $z_t$ yields
$z_t = \bigl(\hat y_t + \sum_{j \in \mathcal{M}^t}\hat y_j + |\mathcal{M}^t|\epsilon\bigr)/(1 + |\mathcal{M}^t|)$.
Together with $z_j = z_t - \epsilon$ for $j \in \mathcal{M}^t$, this is exactly~\eqref{app:eq:zt_def_in_statement},
hence $\mathbf z^t$ is the orthogonal projection of $\hat{\mathbf y}$ onto $V_{\mathcal{M}^t}$.

Since $\mathcal{M}^t \subseteq \mathcal{A}^t$, the approximate solution
$\tilde{\mathbf z}^t$ satisfies $\tilde z_j^t = \tilde z_t^t - \epsilon$ for every
$j \in \mathcal{M}^t$, hence $\tilde{\mathbf z}^t \in V_{\mathcal{M}^t}$. The Pythagorean identity for
orthogonal projection onto an affine subspace then yields
\[
\bigl\|\tilde{\mathbf z}^t - \hat{\mathbf y}\bigr\|_2^2
\;=\; \bigl\|\tilde{\mathbf z}^t - \mathbf z^t\bigr\|_2^2
\;+\; \bigl\|\mathbf z^t - \hat{\mathbf y}\bigr\|_2^2
\;\geq\; \bigl\|\tilde{\mathbf z}^t - \mathbf z^t\bigr\|_2^2,
\]
and taking square roots gives~\eqref{eq:L2_bound_app}.
\end{proof}

\section{Proof of Corollary~\ref{cor:snowball_effect}}
\label{app:cor:snowball_effect}
\textbf{Statement.} {\sl Consider a finite and satisfiable set of signed clauses $\Pi$ and a variable ordering $Y_1;\dots;Y_n$. 
Given $\hat\bfy=m_\theta(\bf{x})$ for some $\bf{x}\in\mathcal X$, let $\tilde\bfa$
be the assignment induced by $\layer(\hat\bfy)$, $\bfa$ be the assignment
induced by the same procedure when each surrogate projection is replaced by
the exact fixed-point projection, and let $\bfa^\star$ be the sequential
feasible argmax assignment,
$a^\star_i=\argmax_{t\in\clS_i(\bfa^\star_{<i})}\hat y_{i,t}$.
Then, for $\epsilon\to0^+$:}
\begin{enumerate}
  \item[(i)]  $\bfa=\tilde\bfa=\bfa^\star$;
  \item[(ii)] $\clS_i(\bfa_{<i})=\clS_i(\tilde\bfa_{<i})=\clS_i(\bfa^\star_{<i})$,
   $\forall i\in[n]$.
\end{enumerate}
\begin{proof}
We prove (i) separately for the exact and the surrogate procedures, showing
in each case that the assignment produced coincides with $\mathbf a^\star$.

\textbf{Exact fixed-point solution.} For $k\in[n]$, let
\[
P(k):\quad a_i=a^\star_i \quad\text{for all } i\le k.
\]
We prove $P(k)$ by induction on $k$, using the following claim.

\emph{Claim.} Let $i\in[n]$. If $\mathbf a_{<i}=\mathbf a^\star_{<i}$, then
$S_i(\mathbf a_{<i})=S_i(\mathbf a^\star_{<i})$ and $a_i=a^\star_i$.

\emph{Proof of the claim.} The feasible set $S_i(\mathbf a_{<i})$ is
obtained by substituting the preceding assignments $\mathbf a_{<i}$ into
$\bar\Pi_i$ (Section~3.2); it depends only on these discrete values and not
on the projected vectors computed for $Y_1,\dots,Y_{i-1}$. Since
$\mathbf a_{<i}=\mathbf a^\star_{<i}$, it follows that
$S_i(\mathbf a_{<i})=S_i(\mathbf a^\star_{<i})=:S_i$, which is non-empty by
Lemma~2 (cf.\ the proof of Theorem~\ref{thm:correctness}). Moreover, the projection for $Y_i$
receives the unmodified input $\hat{\mathbf y}_i$, since the projections for
earlier variables do not alter it. Two cases arise.
\begin{itemize}
  \item If $\argmax_{j\in[h_i]}\hat y_{i,j}\in S_i$, no correction is applied
  (Section~3.2), so
  $a_i=\argmax_{j\in[h_i]}\hat y_{i,j}=\argmax_{t\in S_i}\hat y_{i,t}=a^\star_i$.
  \item Otherwise, by Theorem~\ref{thm:argmax} applied to input
  $\hat{\mathbf y}_i$ and constraint $\{S_i : Y_i\}$, the candidate selected
  in Eq.~(3) is $t^\star_i=\argmax_{t\in S_i}\hat y_{i,t}$. By Theorem~1,
  $t^\star_i$ is the unique maximiser of $\mathbf z^{t^\star_i}$, so
  $a_i=t^\star_i=a^\star_i$.
\end{itemize}
This proves the claim.

\emph{Base case ($k=1$).} No variables precede $Y_1$, so
$\mathbf a_{<1}=\mathbf a^\star_{<1}$ holds trivially. By the claim with
$i=1$, $a_1=a^\star_1$. Hence $P(1)$ holds.

\emph{Inductive step.} Assume $P(k)$ holds for some $k\in[n-1]$, i.e.,
$\mathbf a_{<k+1}=\mathbf a^\star_{<k+1}$. By the claim with $i=k+1$,
$a_{k+1}=a^\star_{k+1}$. Together with $P(k)$, this gives $P(k+1)$.

\emph{Conclusion.} By induction, $P(n)$ holds, i.e., $\mathbf a=\mathbf a^\star$.

\textbf{Surrogate solution.} Let $\tilde P(k)$ be defined as $P(k)$ with
$\tilde a_i$ in place of $a_i$. The argument above applies verbatim with
$\tilde{\mathbf a}$ in place of $\mathbf a$, Eq.~(6) in place of Eq.~(3),
$\tilde{\mathbf z}^{t^\star_i}$ in place of $\mathbf z^{t^\star_i}$, and
Theorem~2 in place of Theorem~1. Theorem~\ref{thm:argmax} again gives
$t^\star_i=\argmax_{t\in S_i}\hat y_{i,t}$, now for $\epsilon\to0^+$, and
Theorem~2 guarantees that $t^\star_i$ is the unique maximiser of
$\tilde{\mathbf z}^{t^\star_i}$. Hence $\tilde{\mathbf a}=\mathbf a^\star$.

Combining the two cases gives $\mathbf a=\mathbf a^\star=\tilde{\mathbf a}$,
which is (i). For (ii), fix $i\in[n]$. By (i),
$\mathbf a_{<i}=\tilde{\mathbf a}_{<i}=\mathbf a^\star_{<i}$, and since
$S_i(\cdot)$ depends only on the preceding assignments,
$S_i(\mathbf a_{<i})=S_i(\tilde{\mathbf a}_{<i})=S_i(\mathbf a^\star_{<i})$.
\end{proof}

\section{Proof of Proposition~\ref{prop:gradient_attenuation}}\label{app:grad_attenuation}

\paragraph{Statement.} {\sl Let $\hat{\mathbf y} \in \Delta([h])$ be such that $t^\star$ and $\mathcal A^{t^\star}$
in~\eqref{eq:closed_form_solution_approximated} are locally constant in a neighbourhood
of $\hat{\mathbf y}$. Let $\layer(\hat \bfy) = \tilde \bfz^t$ and%
$V = \bigl\{\mathbf v \in \mathbb R^{h} : v_j = v_t,\ j \in \mathcal A^t\bigr\}.$
Then the Jacobian $J = \partial\tilde{\mathbf z}^{t}/\partial\hat{\mathbf y}$ is the
orthogonal projector $P_V$ onto $V$. Consequently, for any differentiable downstream
loss $\mathcal L_{\textnormal{task}}$,
\[
\nabla_{\hat{\mathbf y}}\bigl[\mathcal L_{\textnormal{task}} \circ \tilde{\mathbf z}^{t}\bigr]
\;=\; P_V\,\nabla_{\tilde{\mathbf z}}\mathcal L_{\textnormal{task}},
\qquad
\bigl\|P_V\,\nabla_{\tilde{\mathbf z}}\mathcal L_{\textnormal{task}}\bigr\|_2
\;\leq\; \bigl\|\nabla_{\tilde{\mathbf z}}\mathcal L_{\textnormal{task}}\bigr\|_2,
\]
 with the $a$-dimensional subspace $V^\perp$ annihilated entirely. }

\begin{proof}
We can follow the same steps as in the proof of Corollary~\ref{cor:L2_bound} to show that  $\tilde{\mathbf z}^t$ is the
orthogonal projection of $\hat{\mathbf y}$ onto the affine subspace
$V_{\mathcal A^t} = \mathbf p_0 + V$ for any $\mathbf p_0 \in V_{\mathcal A^t}$. Hence
$\tilde{\mathbf z}^t(\hat{\mathbf y}) = \mathbf p_0 + P_V(\hat{\mathbf y} - \mathbf p_0)$,
and differentiating gives $J = P_V$.

The chain-rule expression $\nabla_{\hat{\mathbf y}}[\mathcal L_{\textnormal{task}} \circ \tilde{\mathbf z}^t]
= J^\top \nabla_{\tilde{\mathbf z}^t}\mathcal L_{\textnormal{task}} = P_V \nabla_{\tilde{\mathbf z}^t}\mathcal L_{\textnormal{task}}$
follows since $P_V$ is symmetric, and the contraction
$\|P_V \mathbf g\|_2 \leq \|\mathbf g\|_2$ holds for any $\bfg \in \mathbb{R}^{[h]}$ and orthogonal projector. 

The
coordinate-wise formula comes from the explicit form
$(P_V)_{jk} = \tfrac{1}{1+a}\mathbf 1_{\{j,k \in \mathcal A^t \cup \{t\}\}} + \delta_{jk}\mathbf 1_{\{j \notin \mathcal A^t \cup \{t\}\}}$: 
\[
\frac{\partial\mathcal L_{\textnormal{task}}}{\partial \hat y_k}
=
\begin{cases}
\partial\mathcal L_{\textnormal{task}}/\partial\tilde z_k^{t} & k \notin \mathcal A^t \cup \{t\},\\
\frac{1}{1 + |\mathcal{A}^t|}\,
\left(\frac{\partial \mathcal{L}_{\text{task}}}{\partial \tilde{z}^{t}_t}
+ \sum_{j \in \mathcal{A}^t} \frac{\partial \mathcal{L}_{\text{task}}}{\partial \tilde{z}^{t}_ j}\right) & k \in \mathcal A^t \cup \{t\}.
\end{cases}
\]
\end{proof}

\section{Proof of Proposition~\ref{prop:l2_restores_gradient}}\label{app:gradient_restoration}

\paragraph{Statement.} {\sl 
Under the assumptions of Proposition~\ref{prop:gradient_attenuation}, consider the
composite training objective
$$
\mathcal L_{\textnormal{total}}(\hat{\mathbf y})
= \mathcal L_{\textnormal{task}}\bigl(\tilde{\mathbf z}^{t}(\hat{\mathbf y})\bigr)
+ \tfrac{\lambda}{2}\,\bigl\|\tilde{\mathbf z}^{t}(\hat{\mathbf y}) - \hat{\mathbf y}\bigr\|_2^2,
\qquad \lambda > 0.
$$
Then
$$
\nabla_{\hat{\mathbf y}}\mathcal L_{\textnormal{total}}
\;=\; \underbrace{P_V\,\nabla_{\tilde{\mathbf z}}\mathcal L_{\textnormal{task}}}_{\in\,V}
\;+\; \underbrace{\lambda\,\bigl(\hat{\mathbf y} - \tilde{\mathbf z}^{t}\bigr)}_{\in\,V^\perp}.
$$}

\begin{proof}
The task contribution is $P_V \nabla_{\tilde{\mathbf z}}\mathcal L_{\textnormal{task}}$
by Proposition~\ref{prop:gradient_attenuation}. 
Let
$\phi(\hat{\mathbf y}) = \tfrac12\|\tilde{\mathbf z}^{t}(\hat{\mathbf y}) - \hat{\mathbf y}\|_2^2$
and apply the chain rule:
$$
\begin{aligned}
\nabla_{\hat{\mathbf y}}\phi  = \frac{\partial \phi}{\partial \tilde \bfz^t}\frac{\partial \tilde \bfz^t}{\partial \hat \bfy} + \frac{\partial \phi}{\partial \hat \bfy}  = (J - I)^\top\bigl(\tilde{\mathbf z}^{t} - \hat{\mathbf y}\bigr)  = (P_V - I)\bigl(\tilde{\mathbf z}^{t} - \hat{\mathbf y}\bigr)  = \hat{\mathbf y} - \tilde{\mathbf z}^{t},
\end{aligned}
$$
where the last equality uses $\hat{\mathbf y} - \tilde{\mathbf z}^t \in V^\perp$
(residual orthogonality of the projection in Proposition~\ref{prop:gradient_attenuation}), so
$P_V(\hat{\mathbf y} - \tilde{\mathbf z}^t) = \mathbf 0$.
\end{proof}

\section{Proof of Lemma~\ref{lem:equisat}}\label{app:equisat-proof}

\paragraph{Statement.}{\sl
Let $\know_i$ be a finite and satisfiable set of signed clauses in the variables $Y_1, \ldots, Y_i$. Then $\know_i$ and $\know_{i-1}$ are equisatisfiable, and each assignment to $Y_1, \ldots, Y_{i-1}$ satisfying $\know_{i-1}$ can be extended to satisfy $\know_i$.}

\begin{proof}
    {\sl Forward direction.} If $\Pi_i$ is satisfiable then $\Pi_{i-1}$ is satisfiable. Indeed, every clause in $\mathrm{Confl}_i(\bar{\Pi}_i)$ is derived from $\bar{\Pi}_i \subseteq \Pi_i$
via the intersection rule (Equation~\eqref{eq:intersection}), hence it is entailed by $\Pi_i$). Combined with $\Pi_i \setminus \bar{\Pi}_i \subseteq \Pi_i$, Equation~\eqref{eq:pi_i-1}
yields $\Pi_i \models \Pi_{i-1}$; in particular, if $\mathbf{a} \models \Pi_i$ then its restriction
$\mathbf{a}_{<i}$ (i.e., the assignment $\bfa$ restricted over the variables $Y_1, \ldots, Y_{i-1}$) satisfies $\Pi_{i-1}$, since the latter is over $Y_1,\dots,Y_{i-1}$.

{\sl Backward direction.}
Let $\bar{\Pi}_i = \{ (\mathcal{S}_j : Y_i) \vee \varphi_j \}_{j=1}^{m}$ collect the clauses
of $\Pi_i$ mentioning $Y_i$, with each $\varphi_j$ a signed clause over $Y_1,\dots,Y_{i-1}$.
Let $\mathbf{a}_{<i} \models \Pi_{i-1}$, and assume for contradiction that no $a_i \in [h_i]$
satisfies $(\mathbf{a}_{<i}, a_i) \models \Pi_i$.

Substituting $\mathbf{a}_{<i}$ into the clauses of $\Pi_i$: clauses in $\Pi_i \setminus \bar{\Pi}_i$
are over $Y_1,\dots,Y_{i-1}$ and lie in $\Pi_{i-1}$, hence are satisfied by $\mathbf{a}_{<i}$;
each clause $(\mathcal{S}_j : Y_i) \vee \varphi_j \in \bar{\Pi}_i$ reduces to $(\mathcal{S}_j : Y_i)$
if $\mathbf{a}_{<i} \not\models \varphi_j$, and to a tautology otherwise. Define
\[
\mathcal{K} = \{ j \in [m] : \mathbf{a}_{<i} \not\models \varphi_j \},
\qquad
\bar{\Pi}_i(\mathbf{a}_{<i}) \;=\; \{ (\mathcal{S}_k : Y_i) : k \in \mathcal{K} \}.
\]
By assumption, no $a_i$ satisfies $\bar{\Pi}_i(\mathbf{a}_{<i})$. If $\mathcal{K} = \emptyset$,
any $a_i \in [h_i]$ would do, contradicting the assumption; hence $\mathcal{K} \neq \emptyset$
and $\bigcap_{k \in \mathcal{K}} \mathcal{S}_k = \emptyset$.

By Equation~\eqref{eq:conflict_set}, $\mathcal{K}$ witnesses a clause
\[
\psi = \bigvee_{k \in \mathcal{K}} \varphi_k \;\in\; \mathrm{Confl}_i(\bar{\Pi}_i)
\;\subseteq\; \Pi_{i-1}.
\]
Every $k \in \mathcal{K}$ has $\mathbf{a}_{<i} \not\models \varphi_k$ by definition of $\mathcal{K}$,
so $\mathbf{a}_{<i} \not\models \psi$, contradicting $\mathbf{a}_{<i} \models \Pi_{i-1}$. Hence
some $a_i \in [h_i]$ satisfies $(\mathbf{a}_{<i}, a_i) \models \Pi_i$, which proves the
extension claim. Equisatisfiability follows.

\end{proof}

\section{Proof of Theorem~\ref{thm:complexity}}\label{app:complexity-proof}

\paragraph{Statement.} {\sl 
Let $\Pi$ be a finite set of $r$ signed clauses over the variables $Y_1,\ldots,Y_n$, $k = \max_{i \in [n]}h_i$ and
$w^{\star}(d)$ the conflict width of $\Pi$ along the ordering $d=Y_1;\ldots;Y_n$.  \textsc{Compile}$(\Pi, Y_1;\ldots;Y_n)$ 
in Algorithm~\ref{alg:casper} can be implemented to run in time $\clO((r+n) \cdot k^{w^{\star}(d)+1})$ and space 
$\clO(n \cdot k^{w^{\star}(d)})$.
}
\begin{proof}

The proof proceeds by reducing \textsc{Compile} to bucket elimination~\citep{DECHTER199941}. This suffices to prove the claim, as the latter has worst case time complexity linear in $r+n$ and space linear in $n$, while both grow exponentially — with base $k$ — in the \emph{induced width} $w^{\star}(d)$ of the variable interaction graph associated to a signed-CNF constraint set $\Pi$, along an ordering $d$. Since in our case $\clG_{\Pi}^{\star}(d)$ is defined as the induced graph of $\clG_{\Pi}$ along $d$ in the same way as~\citep{Dechter2013}, then our conflict width coincides with the induced width. 

In particular, we reduce the compilation step of \layer{} in Equations~\eqref{eq:pi_i-1} and~\eqref{eq:conflict_set} to the bucket elimination algorithm 
for belief updating over Bayesian networks in~\cite[Fig.~4.5]{Dechter2013}. The proof comprises three parts: 
\begin{enumerate}
    \item the encoding of the \textsc{Compile} routine in the bucket elimination framework of~\cite{DECHTER199941};
    \item the reduction; and 
    \item the transfer of the time and space bounds from bucket elimination to \textsc{Compile}.
\end{enumerate}

Together these show that \textsc{Compile} admits the same worst-case time and space complexity bounds as classical problems solvable through bucket elimination, like adaptive consistency and directional resolution~\citep{DECHTER199941}.

\textit{1. Relational Encoding.} Consider a set of variables $\bfU = \{Y_{j_1},\ldots,Y_{j_m}\} \subseteq \bfY$ with $j_1 < \ldots < j_m$ ordered along the chosen $d$; let $\clY_{\bfU} = [h_{j_1}] \times \ldots \times [h_{j_m}]$ and let $\bfu = (u_{j_1},\ldots,u_{j_m}) \in \clY_{\bfU}$ be an assignment of the values in $\clY_{\bfU}$ to variables in $\bfU$. 
For a signed clause $\varphi$ so that $\texttt{scope}(\varphi) \subseteq \bfU$,
we say that $\bfu$ satisfies it, written as $\bfu \models \varphi$, if it exists a literal 
$\ell=(\clS_j : Y_j) \in \varphi$ for which it holds that $u_j \in \clS_j$. We can now encode each signed clause $\varphi$ by the relation 
$\mathrm{R}_{\varphi} = \{\bfu \in \clY_{\texttt{scope}(\varphi)}~\mid~\bfu \models \varphi\}$.
For any two relations $\mathrm{R} \subseteq \clY_{\bfU}$ and 
$\mathrm{R}^{\prime} \subseteq \clY_{\bfU^{\prime}}$, their natural join is the relation 
$\mathrm{R} \bowtie \mathrm{R}^{\prime} = \{\bfu \in \clY_{\bfU \cup \bfU^{\prime}}~\mid~ 
\bfu_{\bfU} \in \mathrm{R}, \bfu_{\bfU^{\prime}} \in \mathrm{R}^{\prime}\}$, where $\bfu_{\bfU}$ 
and $\bfu_{\bfU^{\prime}}$ denote the restriction of $\bfu$ to the variables in 
$\bfU$ and in $\bfU^{\prime}$ respectively. On the other hand, for $\bfV \subseteq \bfU$, 
the projection of $\mathrm{R}$ onto $\bfV$ is $\pi_{\bfV}(\mathrm{R}) = \{ \bfu_{\bfV}~\mid~\bfu \in \mathrm{R} \}$.

\begin{table}[t]
\centering
\caption{Correspondence between \textsc{BE-bel}~\cite[Fig.~4.5]{Dechter2013} and \textsc{Compile} (Algorithm~\ref{alg:casper}) under the relational encoding at step 1 of the proof.}
\label{tab:bebel-compile}
\small
\begin{tabular}{@{}lll@{}}
\toprule
& \textsc{BE-bel} & \textsc{Compile} \\
\midrule
Variables, ordering & $X_1, \ldots, X_n$, \quad $d = X_1; \ldots; X_n$ & $Y_1, \ldots, Y_n$, \quad $d = Y_1; \ldots; Y_n$ \\
Max.\ domain size & $k = \max_p |D_p|$ & $k = \max_{i \in [n]} h_i$ \\
Input functions & conditional probability table $P \in \mathbf{P}_G$ & relations $\mathrm{R}_\varphi$, $\varphi \in \Pi$ \\
Number of inputs & $r = |\mathbf{P}_G|$ & $r = |\Pi|$ \\
Combination & product $\prod$ & natural join $\bowtie$ \\
Marginalization & sum $\sum_{X_p}$ & projection $\pi_{\texttt{scope}(\clB_i)}$ \\
Bucket & $\mathit{bucket}_p$: functions with latest var.\ $X_p$ & $\bar{\Pi}_i$: clauses with latest var.\ $Y_i$ \\
Bucket scope & $S_p$ & $\texttt{scope}(\clB_i)$ \\
Inputs in the bucket & $\psi_p = \prod_{P \in \mathit{bucket}_p} P$ & $\bowtie_{\varphi \in \bar{\Pi}_i \cap \Pi} \mathrm{R}_\varphi$ \\
Messages received & $\lambda_{S_1}, \ldots, \lambda_{S_j}$ & $\mathrm{Confl}_q(\bar{\Pi}_q)$, $q > i$, filed in $\bar{\Pi}_i$ \\
Message computed & $\lambda_p = \sum_{X_p} \psi_p \cdot \prod_{l=1}^{j} \lambda_{S_l}$ & $\pi_{\texttt{scope}(\clB_i)}\big(\bowtie_{\varphi \in \bar{\Pi}_i} \mathrm{R}_\varphi\big) \equiv \mathrm{Confl}_i(\bar{\Pi}_i)$ \\
Message placement & latest variable in $S_p$ & latest variable in $\texttt{scope}(\clB_i)$ (Eq.~\eqref{eq:pi_i-1}) \\
Graph & \emph{moral} graph & \emph{interaction} graph $\clG_\Pi$ \\
Width parameter & \emph{induced} width $w^\ast(d)$ & \emph{conflict} width $w^\star(d)$ \\
\bottomrule
\end{tabular}
\end{table}

\textit{2. Reduction.} Under the relational encoding defined in the previous step, the signed resolution in Equation~\eqref{eq:intersection} corresponds to the belief-updating algorithm 
\textsc{BE-bel}~\cite[Fig.~4.5]{Dechter2013}, instantiated over the Boolean relational semiring rather than the sum-product one: indeed, bucket \emph{combination} can be expressed through the natural join $\bowtie$ (taking the role of the product), while bucket \emph{marginalisation} is carried out through the projection operator $\pi$ (instead of the sum). These operations still define a commutative semiring, so the general bucket-elimination scheme of~\cite{DECHTER199941} applies. Table~\ref{tab:bebel-compile} summarizes the correspondence between \textsc{BE-bel}~\cite[Fig.~4.5]{Dechter2013} and \textsc{Compile} under the relational encoding we build: the message $\lambda_p$ computed in a bucket corresponds to the conflict clauses set $\mathrm{Confl}_i(\bar{\Pi}_i)$, while the messages $\lambda_{S_1}, \ldots, \lambda_{S_j}$ it receives correspond to the conflict clauses produced at earlier elimination steps and re-filed into $\bar{\Pi}_i$ by Equation~\eqref{eq:pi_i-1}.

Along the ordering $d$, the \emph{bucket} of $Y_i$ is the set $\clB_i = \bar{\Pi}_i \subseteq \Pi_i$ of the clauses of $\Pi_i$ in which the variable $Y_i$ appears. 
Since by Equation~\eqref{eq:pi_i-1} the variables $Y_{i+1}, \ldots, Y_{n}$ have already been eliminated, then $Y_i$ is the variable occurring in every clause of $\bar{\Pi}_i$ that comes last along $d$. We call the \emph{scope} of a bucket $\clB_i$ the set of the variables — other than the last one — occurring in it, namely:
\[ \texttt{scope}(\clB_i) = \big ( \bigcup_{\varphi\in\bar{\Pi}_i} \texttt{scope}(\varphi) \big ) \setminus \{Y_i\}. \]

Then the per-bucket message computed by \textsc{BE-bel} in the bucket of $Y_i$ — namely 
$\lambda_i = \sum_{Y_i} \psi_i \prod_l \lambda_{S_l}$ with $\psi_i$ being the product of input functions in the corresponding bucket and — over the Boolean semiring — $\lambda_i(\bfu) = 1$ when some value of $Y_i$ makes every function in the bucket true — becomes the relation $\pi_{\texttt{scope}(\clB_i)}(\bowtie_{\varphi \in \bar{\Pi}_i} \mathrm{R}_{\varphi}) \subseteq \clY_{\texttt{scope}(\clB_i)}$.
We remark that joining the bucket for $\bar{\Pi}_i$ and projecting over its scope rules out the tuples with no consistent value for $Y_i$:
by Equation~\eqref{eq:conflict_set}, these are exactly the tuples falsifying some clause of $\mathrm{Confl}_i(\bar{\Pi}_i)$ (cf.\ the proof of Lemma~\ref{lem:equisat}).
Therefore, the conflict set $\mathrm{Confl}_i(\bar{\Pi}_i)$ from Equation~\eqref{eq:conflict_set} can be equivalently expressed as the set of signed clauses:
\begin{equation}\label{eq:conflict_set_red}
    \{ \bigvee_{Y_j \in \texttt{scope}(\clB_i)}([h_j] \setminus \{u_j\}) : Y_j \mid \bfu \in \clY_{\texttt{scope}(\clB_i)} \setminus \pi_{\texttt{scope}(\clB_i)}(\bowtie_{\varphi\in\bar{\Pi}_i} \mathrm{R}_{\varphi})\},
\end{equation}
containing a maximum of $k^{|\texttt{scope}(\clB_i)|}$ clauses. 
It follows that, \textsc{BE-bel} can compute through its bucket-elimination procedure the conflict set in Equation~\eqref{eq:conflict_set}. The result is then the same computed by \textsc{Compile} in virtue of Equation~\eqref{eq:pi_i-1}. \textsc{BE-bel} carries out the same operation as \textsc{Compile} across the buckets: indeed, (i) its buckets partition the clauses by the variable coming last in $d$, (ii) the elimination proceeds backwards from $Y_n$ to $Y_1$, and (iii) the derived constraints are re-filed into lower bucket. 

Since Lemma~\ref{lem:equisat} — and therefore the correctness of \layer{} — depends on $\Pi_{i-1}$ only through the assignments who satisfy it, 
then \textsc{Compile} can compute the set in Equation~\eqref{eq:conflict_set_red} in place of $\mathrm{Confl}_i(\bar{\Pi}_i)$.

\textit{3. Cost transfer.} Eliminating variable $Y_i$ requires computing the combined bucket 
$\bowtie_{\varphi \in \bar{\Pi}_i} \mathrm{R}_{\varphi}$. This is a relation over the $|\texttt{scope}(\clB_i)| + 1$
variables, hence it contains at most $k^{|\texttt{scope}(\clB_i)| + 1}$ tuples, where $k = \max_{i \in [n]} h_i$
is the maximum size of the domain of each variable. 
Since the complexity of \textsc{BE-bel} depends only on the number of functions combined in each bucket and on the size of the combined bucket,
regardless of the particular semiring of choice, the complexity bound transfers. 
To conclude the proof we need to show that $|\texttt{scope}(\clB_i)| \leq w^{\star}(d)$ for every $i \in [n]$.
To this end, we show that, for every $i = n,\ldots,1$ it holds that,
\[ P(i): \qquad \clG_{\Pi_i} \;\text{is a subgraph of}\; (\bfY, E_i), \]
with $E_i$ defined as in Equation~\eqref{eq:ei-1}, i.e., any two variables appearing together in a clause of 
$\Pi_i$ are adjacent in $E_i$. Notice that $P(i)$ implies $|\texttt{scope}(\clB_i)| \leq w^{\star}(d)$. 
Indeed, for every $Y_j \in \texttt{scope}(\clB_i)$ that appears in a clause of $\bar{\Pi}_i \subseteq \Pi_i$ together with $Y_i$, it holds that $(Y_i,Y_j) \in E_i$ by $P(i)$. 
Observe that $j < i$, since $Y_{i+1}, \ldots, Y_n$ do not appear in $\Pi_i$. As $E_i \subseteq E_0$, we get that $Y_j \in \clN_d(Y_i)$, so $\texttt{scope}(\clB_i) \subseteq \clN_d(Y_i)$ and therefore we obtain the desired bound $|\texttt{scope}(\clB_i)| \leq |\clN_d(Y_i)| \leq w^{\star}(d)$.

We now prove $P(i)$ by induction on $i = n,\ldots,1$. We follow the same argument used in~\cite[Lemma~2]{davis-putnam}, showing that directional resolution never couples variables that are not adjacent in the induced graph.

\textit{Base case ($i = n$).} For $i = n$, we have $\Pi_n = \Pi$ and $E_n = E_{\Pi}$. So $\clG_{\Pi_n} = \clG_{\Pi}$ and 
$P(n)$ holds as $\clG_{\Pi}$ is the graph $(\bfY, E_n)$.

\textit{Inductive step ($i \rightarrow i-1$).} Assume $P(i)$ holds. By Equation~\eqref{eq:pi_i-1} every clause $\varphi \in \Pi_{i-1}$ falls in one of two cases: 
\begin{itemize}
    \item $\varphi \in \Pi_i \setminus \bar{\Pi}_i$ is a carried over clause in $\Pi_{i-1}$: any two variables in $\texttt{scope}(\varphi)$ are 
    neighbours in $E_i$ by $P(i)$, hence in $E_{i-1} \supseteq E_i$;
    \item $\varphi$ is a conflict clause in Equation~\eqref{eq:conflict_set_red}, hence
    so that $\texttt{scope}(\varphi) \subseteq \texttt{scope}(\clB_i)$. Any two distinct $Y_j, Y_l \in \texttt{scope}(\clB_i)$ satisfy $j,l < i$
    and $\{(Y_j, Y_i),(Y_l, Y_i)\} \subseteq E_i$, as shown above. Therefore, Equation~\eqref{eq:ei-1} adds the edge $(Y_j,Y_l)$ to $E_{i-1}$
    and any two variables in $\texttt{scope}(\varphi)$ are adjacent in $E_{i-1}$.
\end{itemize}
It follows that $\clG_{\Pi_{i-1}}$ is a subgraph of $(\bfY, E_{i-1})$ which proves $P(i -1)$.
Since we showed that under the relational encoding we provided in step 1. \textsc{Compile} reduces to the \textsc{BE-bel} algorithm, 
and that $|\texttt{scope}(\clB_i)| \leq w^{\star}(d)$ for every $i \in [n]$, then the proof of~\cite[Thm.~4.5, Sec.~4.8]{Dechter2013} yields the bounds:
\[ \textup{time}\; \clO((r+n) \cdot k^{w^{\star}(d)+1}) \qquad \textup{and} \qquad \textup{space}\; \clO(n \cdot k^{w^{\star}(d)}). \]
\end{proof}
We remark that~\cite{Dechter2013} states the bound as $\clO(r \cdot k^{w^\star(d)+1})$ under the assumption $r > n$, which holds for Bayesian networks but not necessarily for $\Pi$.
Also, our result highlights a (somewhat expected) super-polynomial dependence: since \textsc{Compile} decides the satisfiability of $\Pi$ (Algorithm~\ref{alg:casper}), and satisfiability of signed-CNF formulae is NP-complete~\citep{beckert2000_signed_CNF}, no implementation of \textsc{Compile} runs in time polynomial in $r$, $n$ and $k$ unless $\mathrm{P} = \mathrm{NP}$.

\section{Proof of Theorem~\ref{thm:correctness}}\label{app:correctness-proof}

\paragraph{Statement.} {\sl  
Let $\Pi$ be a finite and satisfiable set of signed clauses. For any datapoint $\bf{x}$, model $m_\theta$ and variable ordering, the corresponding output $\layer(m_\theta(\bf{x}))$ satisfies $\Pi$.
}

\begin{proof}
By induction on the number $n$ of variables appearing in $\Pi$.

\textit{Base case ($n = 0$):} $\Pi$ contains no variables and is trivially satisfied by any assignment.%

\textit{Inductive step ($n > 0$):}
Let $Y_n$ be the variable in $\Pi$ with the highest
index in the ordering, so $\Pi_n = \Pi$. Since $\Pi_{n-1}$ involves $n - 1$ variables
and is satisfiable, the inductive hypothesis gives that the partial assignment
$\mathbf a_{<n}$ induced by $\layer(m_\theta(\mathbf x))$ satisfies $\Pi_{n-1}$.
Substituting $\mathbf a_{<n}$ into $\Pi$ leaves clauses involving only $Y_n$; by
Lemma~\ref{lem:equisat}, this residual is satisfiable and reduces to a single signed
clause
$(\clS^\star : Y_n)$ by replacing each variable in $Y_1, \ldots, Y_{n-1}$ with the value given by the assignment $\bfa_{<n}$ induced by $\layer(m_\theta(\bfx))$. Then, Theorem~\ref{th:constraint_satisfaction_basic_case} ensures that $\arg\max_{j \in [h_n]} \layer(m_\theta(\bfx))_n \in \clS^\star$, and hence $\layer(m_\theta(\bfx))$ satisfies $\Pi_n = \Pi$.
\end{proof}

\section{Additional Experimental Results}\label{app:full-results}

\subsection{Warcraft Pathfinding.}\label{app:experiments_warcraft}
\paragraph{Task Details} 
We consider a shortest-path finding task on Warcraft II terrain tiles as proposed by \citep{poganvcic2019differentiation}, with the dataset generated using the MIT-licensed \texttt{war2edit} toolset\footnote{\url{https://github.com/war2/war2edit}}; the underlying tile assets remain the property of Blizzard Entertainment and are used here for research purposes. The full rendered terrain image 
is $96\!\times\!96$ pixels at the $N{=}12$ version and $240\!\times\!240$ at $N{=}30$ one. Given a 
\begin{wraptable}{r}{0.56\textwidth}
        \caption{Warcraft II Tile Cost Mapping}
        \label{tab:tile_costs}
        \begin{tabular}{llc}
            \toprule
            \textbf{Label} & \textbf{Terrain Type} & \textbf{Cost} \\
            \midrule
            Land                & Grass / Open Ground & 0.8 \\
            Dirt                & Dirt / Coast        & 1.2 \\
            Forest \& Mountains & Forest              & 5.3 \\
            Forest \& Mountains & Mountains           & 7.7 \\
            Water               & Water               & 9.2 \\
            \bottomrule
        \end{tabular}
\end{wraptable}
$N \times N$ grid rendered from Warcraft II terrain tiles, the goal of the task is to predict the minimum-cost paths from the top-left to the bottom-right corner. We test grid sizes, $N \in \{12, 30\}$ and each tile has a traversal cost outlined in Table~\ref{tab:tile_costs} with higher costs corresponding to harder-to-traverse terrain. The constraint is that the predicted path forms a valid route: connected, contained within the grid, starting at the source (0,0) and ending at $(N-1, N-1)$. 

\paragraph{Dijkstra-based methods.} We do not include A-NeSI~\citep{van2023nesi} or I-MLE~\citep{niepert2021imle} among the
baselines. These approaches predict a tile-cost map $c_\theta(\bfx)$ and then
use a task-specific shortest-path solver to return
\[
\argmin_{p\in\mathcal P}\sum_{v\in p}c_{\theta,v}(\bfx),
\]
where $\mathcal P$ is the set of valid paths. The solver therefore performs
not only feasibility enforcement but also the combinatorial optimization that
selects the path, while the network learns the costs rather than directly
predicting the structured output. In our formulation, as in SPL~\citep{ahmed2022semantic}, the network directly predicts the path, and hence must learn from the image which of the feasible paths is optimal.
The two formulations thus differ in their neural output space, optimization
objective, and division of labour between the neural and symbolic components. Dijkstra-based decoding is a strong, task-specific alternative that is complementary to our approach; however, it does not address the setting studied here, in which the network must directly predict the structured output and constraints are enforced on those predictions.

\begin{figure}[t]
    \centering
    \begin{subfigure}[b]{\linewidth}
        \centering
        \includegraphics[width=1\linewidth]{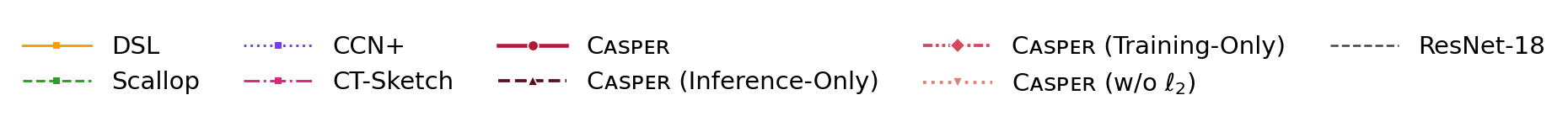}
    \end{subfigure}
    
    \begin{subfigure}[b]{0.49\linewidth}
        \centering
        \includegraphics[width=\linewidth]{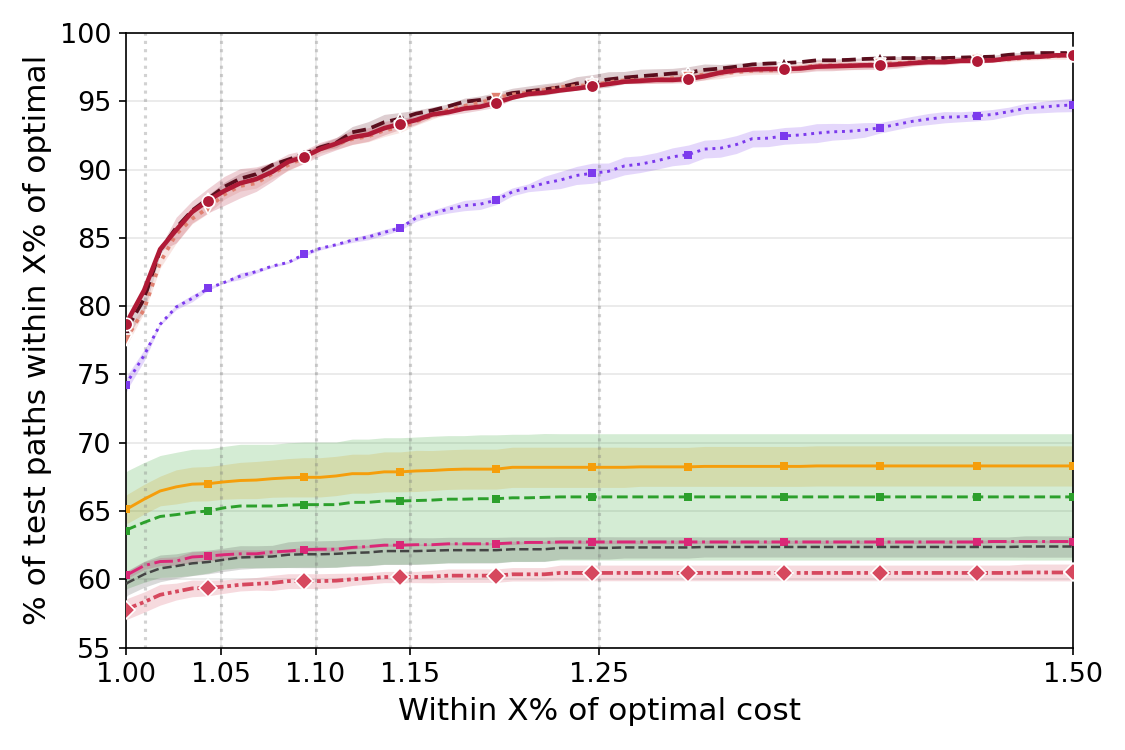}
        \caption{$12\times 12$}
        \label{fig:mnist-sum-scaling}
    \end{subfigure}%
    \hfill%
    \begin{subfigure}[b]{0.49\linewidth}
        \centering
        \includegraphics[width=\linewidth]{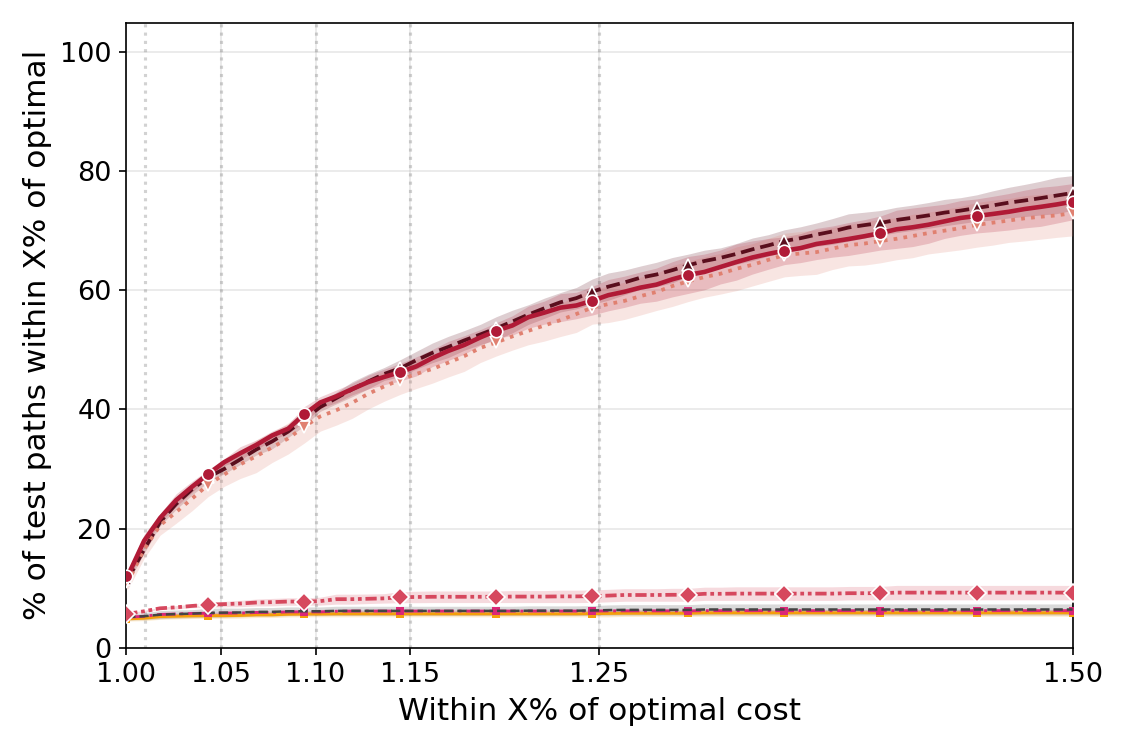}
        \caption{$30 \times 30$}
        \label{fig:mnist-add-scaling}
    \end{subfigure}
    
    \caption{Cumulative path cost-optimality for Warcraft Pathfinding}
    \label{fig:warcraft_cost_analysis}
\end{figure}

\paragraph{Experimental Setup.} For both grid sizes, $N \in \{12, 30\}$, we use a fixed split of 10k training, 1k 
validation, and 1k test examples. The neural backbone is a ResNet-18 \citep{he2016deep} trained from scratch. We train with Adam at learning rate $5\mathrm{e}{-}4$ under a cosine schedule ($\eta_{\min} = 5\mathrm{e}{-}6$), batch size 128, and 300 epochs with early stopping. In addition to the metrics reported in the main text, and for coherence with previous work~\cite{ahmed2022semantic}, we additionally report Hamming accuracy and Exact match. Hamming accuracy is the per-edge accuracy averaged over the grid, and Exact match is the percentage of predicted paths that match the ground-truth shortest path edge-for-edge. We further report Optimal ($\%$), the percentage of predicted paths whose total cost equals the ground-truth cost, along with the fractions within $1\%$, $5\%$, and $10\%$ of the ground-truth cost. For methods that do not always produce consistent paths, these fractions are computed over the full test set (not solely on consistent paths): since inconsistent predictions count as failures regardless of the cost of any partial path they may induce. These metrics are reported in Tables~\ref{tab:warcraft-12} and~\ref{tab:warcraft-30} for the $12\times12$ and $30 \times30$ grids, respectively. To complement the exact-optimality metric, Figure~\ref{fig:warcraft_cost_analysis} (Appendix~\ref{app:full-results}) reports a cumulative cost-optimality analysis: for each tolerance factor $\alpha \geq 1$, we measure the fraction of predictions whose cost is within a factor $\alpha$ of the optimal path cost, providing a finer-grained view of how close non-optimal predictions are to the optimum. Consistency is the percentage of predicted paths that form a valid route: connected, contained within the grid, with correct degree at all nodes.
For methods that predict a walk, all metrics are computed on the loop-erased path. All metrics are reported as mean $\pm$ std over 5 seeds. 

\paragraph{Path encoding.} \layer{} and all structured baselines except
CTSketch share a single step-wise encoding of the path constraint
((Appendix~\ref{sec:trimmed-steps})), compiled once into an artifact from which every method reads
its domains and transitions. For an $N \times N$ grid with cells numbered in
row-major order, source $s = 0$ and target $t = N^2 - 1$, the variables
$X_0, \ldots, X_L$ with $L = N^2 - 1$ denote the cell occupied at each tick.
A cell $v$ belongs to the domain $D_i$ of $X_i$ only if it is reachable from
$s$ by tick $i$ ($d(s,v) \leq i$), has matching parity
($i \equiv d(s,v) \bmod 2$), and can still reach $t$ in time
($d(v,t) \leq L - i$); $X_0 = s$ and $X_L = t$ are fixed. Transitions require
$X_{i+1} \in \mathrm{succ}(X_i) \cap D_{i+1}$, where $\mathrm{succ}(u)$ is the
set of grid neighbours of $u$, except that $t$ can only move to itself. No
all-different constraint is imposed, so walks may revisit cells; loop erasure
turns a predicted walk into a simple path.

\paragraph{Implementation Details.}
We compare \layer{} against the remaining six  neurosymbolic baselines once again spanning exact and approximate methods: DeepProbLog (DPL) \citep{manhaeve2018deepproblog}, Semantic Probabilistic Layers (SPL) \citep{ahmed2022semantic}, CCN+ \citep{giunchiglia2024ccn+}, CTSketch \citep{choi2025ctsketch}, Scallop \citep{huang2021scallop}, and DeepSoftLog (DSL) \citep{maene2023soft}. Both \layer{} and CCN+ enforce path connectivity through a differentiable layer built on the steps encoding. For \layer{}, the per-cell probabilities $P(v \text{ on path})$ are read on the domains $D_i$; a walk selects the highest-scoring allowed successor at each step, and loop erasure turns it into a path. This path sets a target category for each cell, and each cell's distribution is projected with the smallest $\ell_2$ edit that makes its target the argmax. CCN+ instead encodes the steps constraint as propositional clauses over one Boolean atom per step and cell in $D_i$, requiring at least one cell per step and that $X_i = u$ implies one of its allowed successors (the at-most-one constraint is omitted); the clauses are compiled into a PiShield shield fed with the same scores as \layer{}, and prediction is a greedy walk over the shielded scores followed by loop erasure. Scallop and DSL incorporate path validity through their respective probabilistic logic backends during training. Scallop encodes the steps constraint as a differentiable Datalog program over the relations \texttt{at}$(i,v)$, \texttt{dom}$(i,v)$ and \texttt{nxt}$(i,j)$ read from the compiled artifact, with grid edges as probabilistic input facts given by the network's per-node outputs: a walk at cell $u$ at tick $i$ can move along an edge to $v$ if $v \in D_{i+1}$, remains at $t$ once reached, and \texttt{valid\_path} holds when \texttt{at}$(L, t)$ holds. Training minimizes $-\log P(\text{valid\_path})$ alongside the supervised loss. Each proof is one valid walk, whose probability is the product of its edge probabilities; following the default settings, we retain only the top-3 highest-probability walks per fact. This signal is a soft, approximate penalty applied during training rather than a hard constraint. At inference, the prediction is the raw per-cell argmax, without the encoding, so nothing couples these choices into a connected route. The constraint therefore acts on the probability of a valid walk during training, not on this independent decoding, and does not transfer to inference, hence Scallop's low consistency.

SPL compiles the steps constraint into an SDD with one Boolean variable per legal (tick, from, to) move, each taking the probability of the grid edge it crosses, with sink-to-sink moves neutral; prediction uses SPL's greedy decoding through the circuit. CTSketch approximates the per-node path-connectivity constraint via a Tensor-Train~(TT) decomposition of a symbolic one-hot tensor $\phi_{\mathrm{conn}} \in \{0,1\}^{2\times 2\times 2\times 2\times 3}$.
The four binary modes of $\phi_{\mathrm{conn}}$ encode the occupancy of the four incident edges of a node (left, right, up, down in canonical
order), and the final mode indexes three admissible routing classes: degree-zero (off-path node), degree-one (path endpoint), and degree-two
(interior through-node); edge-degree configurations of three or higher receive zero weight in every class, producing a large penalty in
log-space. We deploy CTSketch in two variants that differ in how the terrain image is presented to the backbone and how path-consistency is enforced.

For \layer{}, we set the surrogate threshold $\epsilon = 1\mathrm{e}{-}3$ and add an auxiliary $L_2$ loss between the unprojected and projected per-cell distributions weighted by $\lambda = 1.0$, as motivated by Proposition~\ref{prop:l2_restores_gradient}. The step variables are processed in order, $X_0; \ldots; X_L$. For DSL, we embed the steps constraint in a symbolic functor~\citep{maene2023soft} that runs max-product propagation over the steps encoding, one pass per tick ($L = N^2 - 1$ passes: $143$ at $N = 12$ and $899$ at $N = 30$): after each pass, cells outside $D_i$ are set to zero and $t$ may remain at itself, giving $P(\text{valid}) = \mathrm{reach}_L(t)$. As for Scallop, this term enters only the training loss, and prediction is the raw per-cell argmax. CCN+ and Scallop are run at $N = 12$ only: CCN+'s clause encoding does not scale to $N = 30$, and Scallop exceeds the time-out limit at $N = 30$. For SPL, we run with a configuration consistent with~\citep{ahmed2022semantic}. Lastly, for CTSketch we run two distinct modalities: \emph{CTSketch} and \emph{CTSketch Grid}
(cf.~Table~\ref{tab:warcraft-12},~\ref{tab:warcraft-30}).

\emph{Single modality (CTSketch).} The full rendered terrain image  
is forwarded through the ResNet-18 backbone, yielding independent per-edge 
logits for all $2N(N-1)$ grid edges. The symbolic tensor $\phi_{\mathrm{conn}}$ 
is factorised offline into a chain of TT cores via TT-SVD. At each of the $N^{2}$
grid nodes, the predicted probability $p_e$ of each incident edge is
assembled into a 2-state distribution $[1-p_e,\, p_e]$ (boundary
edges absent from the grid are clamped to $[0, 1]$).  These
distributions are contracted with the shared TT-core chain to produce a per-node
log-score vector over the three routing classes. The symbolic
connectivity loss $\mathcal{L}_{\mathrm{CTS}}$ is defined as the mean
negative log-likelihood of the ground-truth routing class, determined
by the node's degree in the ground-truth shortest path (zero, one, or
two). The full training objective is given by
\(
  \mathcal{L} = \mathcal{L}_{\mathrm{BCE}} + \lambda_{\mathrm{CTS}}\,
  \mathcal{L}_{\mathrm{CTS}},
\)
with $\lambda_{\mathrm{CTS}} = 1.0$.  No post-processing is applied at
inference time; predicted paths are obtained by thresholding the
per-edge sigmoid outputs.

\emph{Grid modality (CTSketch~Grid).}
The terrain image is partitioned into a $3\!\times\!3$ grid of
non-overlapping tiles, each covering $\lfloor N/3\rfloor \times \lfloor
N/3\rfloor$ terrain nodes ($4\!\times\!4$ nodes per tile at $N{=}12$).
The ResNet-18 backbone is applied independently to each tile and
predicts only the edges whose both endpoints lie within that tile.  
The binary cross-entropy loss and the CTS loss $\mathcal{L}_{\mathrm{CTS}}$
are correspondingly restricted to within-tile edges.  An additional
global regulariser $\mathcal{R}_{\mathrm{global}}$, weighted by
$\lambda_{\mathrm{global}} = 1.0$, penalises cross-tile inconsistencies
at evaluation time by scoring boundary-node degree compatibility between
adjacent tiles. At inference time, within-tile edge predictions are thresholded, and a
deterministic stitching procedure fills the missing cross-tile edges:
starting from the top-left tile, the algorithm greedily identifies
pairs of adjacent tiles with degree-one boundary nodes and connects them.  

\paragraph{CTSketch Ablation.}
We study the sensitivity of both CTSketch modalities introduced above 
to the TT approximation rank on the $12\!\times\!12$ grid, evaluating ranks
$r \in \{3, 10, 30\}$ with TT-SVD decomposition across $5$ random
seeds per configuration.

\emph{Single modality.}
The connectivity tensor $\phi_{\mathrm{conn}}$ is a sparse binary array
of shape $(2,2,2,2,3)$ whose 
\begin{table}[t]
    \centering
    \caption{Computational efficiency on Warcraft Shortest Path across grid-sizes.}
    \label{tab:warcraft-efficiency}
    \setlength{\tabcolsep}{4pt}
    \begin{tabular}{l cc cc}
        \toprule
        & \multicolumn{2}{c}{\textbf{Peak Memory (GB)}} & \multicolumn{2}{c}{\textbf{Epoch Time (s)}} \\
        \cmidrule(lr){2-3} \cmidrule(lr){4-5}
        \textbf{Method} & $12{\times}12$ & $30{\times}30$ & $12{\times}12$ & $30{\times}30$ \\
        \midrule
        \addlinespace[2pt]
        \rowcolor{exactshade}\multicolumn{5}{l}{\textit{Exact Methods}} \\
        \addlinespace[2pt]
        DPL             & T/O  & T/O & T/O   & T/O \\
        SPL             & T/O  & T/O & T/O   & T/O \\
        CCN+            & 9.93 & T/O & 202.8 & T/O \\
        \addlinespace[6pt]
        \rowcolor{approxshade}\multicolumn{5}{l}{\textit{Approximate Methods}} \\
        \addlinespace[2pt]
        Scallop         & 0.32 & T/O  & 1682.4 & T/O \\
        DSL             & 0.48 & 2.45 & 616.2  & 1072.5 \\
        CTSketch        & 0.71 & 1.82 & 1.98   & 17.97 \\
        CTSketch (Grid) & 0.76 & 1.93 & 2.03   & 17.67 \\
        \midrule
        {\layer{} (Ours)} & 0.69 & 3.33 & 11.8 & 80.8 \\
        \bottomrule
    \end{tabular}
\end{table}
intrinsic TT-rank is~3: a TT-SVD at
$r{=}3$ already achieves a reconstruction error below $5\!\times\!10^{-7}$
in the $\ell_\infty$ norm, making the factorisation effectively
lossless.  As a direct consequence, all three rank settings yield
nearly indistinguishable training dynamics and final performance.
We adopt $r\!=\!30$ as the final configuration reported in the main
results, as it achieves the best aggregate performance and provides a
large margin above the minimum admissible rank, ensuring numerical
stability of the TT contraction throughout training.

\emph{Grid modality.}
Across all three ranks, CTSketch~(Grid) yields consistently degenerate
path-level outcomes: exact-match and consistency figures remain near
$0.2\%$ irrespective of $r$
(cf.~Table~\ref{tab:warcraft-12},~\ref{tab:warcraft-30}).  Within-tile edge quality is
comparatively higher, as reflected by Hamming accuracies of
$\approx\!91.5\%$, indicating that the backbone learns meaningful
per-tile edge distributions under the tile-local CTS supervision.
The failure to produce valid global paths is therefore attributable
not to the quality of the TT approximation—which is exact at every
tested rank—but to the structural limitation the method exhibits during global reconciliation: 
tile-local policies are optimised independently,
with no training signal governing the entry and exit points at tile
boundaries, so the greedy cross-tile connector cannot reliably assemble
a globally consistent shortest path.

\paragraph{Computational Efficiency}
Table~\ref{tab:warcraft-efficiency} reports peak GPU memory and per-epoch training time for each method on both grid sizes $N \in \{12, 30\}$. \layer{}'s memory usage and epoch-time grow gracefully with increasing grid-size. All other exact methods, which guarantee path consistency, are unable to run at $N=30$ without breaching the 5s per-sample time-out limit. Of the approximate methods, CTSketch is the most memory and time efficient at $N=30$; however, as outlined in Table~\ref{tab:warcraft-30}, this comes at the cost of both optimality ($5.1\%$ cost-optimal paths vs $12.1\%$ for \layer{}) and consistency ($6.1\%$ consistency vs $100\%$ for \layer{}).

\begin{table}[t]
    \centering
    \caption{Warcraft Pathfinding Task (12$\times$12). Each metric is presented as a percentage (\%). EM stands for Exact-Match.}
    \label{tab:warcraft-12}
    \setlength{\tabcolsep}{3pt}
    \definecolor{exactshade}{RGB}{252, 228, 230}%
    \definecolor{approxshade}{RGB}{223, 237, 247}%
    \definecolor{ablationshade}{RGB}{223, 247, 230}
    \begin{tabular}{l c cccc cc}
        \toprule
        \multirow{2}{*}{\textbf{Method}} & \multirow{2}{*}{\textbf{EM}} & \multicolumn{4}{c}{\textbf{Optimal}} & \multirow{2}{*}{\textbf{Hamming}} & \multirow{2}{*}{\textbf{Consistent}} \\
        \cmidrule(lr){3-6}
        & & == & \textbf{$\leq 1\%$} & \textbf{$\leq 5\%$} & \textbf{$\leq 10\%$} & & \\
        \midrule
        ResNet-18                 & $58.5 \msmall{\pm 1.4}$ & $59.7 \msmall{\pm 1.2}$ & $60.4 \msmall{\pm 1.2}$ & $61.4 \msmall{\pm 1.1}$ & $61.8 \msmall{\pm 1.2}$ & $98.3 \msmall{\pm 0.1}$ & $62.5 \msmall{\pm 1.0}$ \\
        \midrule
        \addlinespace[2pt]
        \rowcolor{exactshade}\multicolumn{8}{l}{\textit{Exact Methods}} \\
        \addlinespace[2pt]
        CCN+                      & $68.6 \msmall{\pm 0.8}$ & $74.2 \msmall{\pm 0.6}$ & $76.5 \msmall{\pm 0.5}$ & $81.6 \msmall{\pm 0.1}$ & $84.2 \msmall{\pm 0.2}$ & $97.5 \msmall{\pm 0.0}$ & $\mathbf{100.0} \msmall{\pm 0.0}$ \\
        SPL                       & T/O & T/O & T/O & T/O & T/O & T/O & T/O \\
        DPL                       & T/O & T/O & T/O & T/O & T/O & T/O & T/O \\
        \addlinespace[6pt]
        \rowcolor{approxshade}\multicolumn{8}{l}{\textit{Approximate Methods}} \\
        \addlinespace[2pt]
        CTSketch                  & $59.8 \msmall{\pm 1.1}$ & $60.4 \msmall{\pm 1.1}$ & $61.0 \msmall{\pm 1.2}$ & $61.8 \msmall{\pm 1.2}$ & $62.3 \msmall{\pm 1.1}$ & $98.4 \msmall{\pm 0.1}$ & $62.8 \msmall{\pm 1.1}$ \\
        CTSketch (Grid)           & $0.2 \msmall{\pm 0.1}$ & $0.2 \msmall{\pm 0.1}$ & $0.2 \msmall{\pm 0.1}$ & $0.2 \msmall{\pm 0.1}$ & $0.2 \msmall{\pm 0.1}$ & $91.5 \msmall{\pm 0.1}$ & $0.2 \msmall{\pm 0.1}$ \\
        Scallop                   & $63.0 \msmall{\pm 5.5}$ & $63.6 \msmall{\pm 5.3}$ & $64.2 \msmall{\pm 5.4}$ & $65.1 \msmall{\pm 5.5}$ & $65.5 \msmall{\pm 5.6}$ & $\mathbf{98.5 \msmall{\pm 0.1}}$ & $66.1 \msmall{\pm 5.7}$ \\
        DSL                       & $64.0 \msmall{\pm 1.8}$ & $65.1 \msmall{\pm 1.3}$ & $65.9 \msmall{\pm 1.3}$ & $67.1 \msmall{\pm 1.6}$ & $67.5 \msmall{\pm 1.8}$ & $\mathbf{98.5 \msmall{\pm 0.1}}$ & $68.3 \msmall{\pm 1.8}$ \\
        \addlinespace[6pt]
        \rowcolor{ablationshade}\multicolumn{8}{l}{\textit{Ablation Studies}} \\
        \addlinespace[2pt]
        \layer{} (Train Only)     & $56.6 \msmall{\pm 0.7}$ & $57.8 \msmall{\pm 0.9}$ & $58.3 \msmall{\pm 0.9}$ & $59.4 \msmall{\pm 0.7}$ & $59.9 \msmall{\pm 0.7}$ & $98.3 \msmall{\pm 0.1}$ & $60.5 \msmall{\pm 0.7}$ \\
        \layer{} (Inf. Only)      & $73.6 \msmall{\pm 0.9}$ & $78.3 \msmall{\pm 0.8}$ & $80.7 \msmall{\pm 1.1}$ & $\mathbf{88.4 \msmall{\pm 0.5}}$ & $\mathbf{91.6 \msmall{\pm 0.3}}$ & $98.1 \msmall{\pm 0.0}$ & $\mathbf{100.0} \msmall{\pm 0.0}$ \\
        \layer{} (w/o $\ell_2$)   & $72.6 \msmall{\pm 0.8}$ & $77.5 \msmall{\pm 0.6}$ & $80.0 \msmall{\pm 0.3}$ & $87.8 \msmall{\pm 0.4}$ & $91.3 \msmall{\pm 0.4}$ & $98.1 \msmall{\pm 0.0}$ & $\mathbf{100.0} \msmall{\pm 0.0}$ \\
        \midrule
        \layer{}                  & $\mathbf{73.9 \msmall{\pm 0.3}}$ & $\mathbf{78.7 \msmall{\pm 0.3}}$ & $\mathbf{81.3 \msmall{\pm 0.2}}$ & $88.1 \msmall{\pm 1.3}$ & $91.4 \msmall{\pm 0.5}$ & $98.1 \msmall{\pm 0.0}$ & $\mathbf{100.0} \msmall{\pm 0.0}$ \\
        \bottomrule
    \end{tabular}%
\end{table}

\begin{table}[t]
    \centering
    \caption{Warcraft Pathfinding Task (30$\times$30). Each metric is presented as a percentage (\%). EM stands for Exact-Match.}
    \label{tab:warcraft-30}
    \setlength{\tabcolsep}{3pt}
    \begin{tabular}{l c cccc cc}
        \toprule
        \multirow{2}{*}{\textbf{Method}} & \multirow{2}{*}{\textbf{EM}} & \multicolumn{4}{c}{\textbf{Optimal}} & \multirow{2}{*}{\textbf{Hamming}} & \multirow{2}{*}{\textbf{Consistent}} \\
        \cmidrule(lr){3-6}
        & & == & \textbf{$\leq 1\%$} & \textbf{$\leq 5\%$} & \textbf{$\leq 10\%$} & & \\
        \midrule
        ResNet-18              & $5.1 \msmall{\pm 0.6}$ & $5.1 \msmall{\pm 0.6}$ & $5.3 \msmall{\pm 0.7}$ & $5.9 \msmall{\pm 0.9}$ & $6.1 \msmall{\pm 0.9}$ & $97.5 \msmall{\pm 0.0}$ & $6.4 \msmall{\pm 1.1}$ \\
        \midrule
        \addlinespace[2pt]
        \rowcolor{exactshade}\multicolumn{8}{l}{\textit{Exact Methods}} \\
        \addlinespace[2pt]
        CCN+                   & T/O & T/O & T/O & T/O & T/O & T/O & T/O \\
        SPL                    & T/O & T/O & T/O & T/O & T/O & T/O & T/O \\
        DPL                    & T/O & T/O & T/O & T/O & T/O & T/O & T/O \\
        \addlinespace[6pt]
        \rowcolor{approxshade}\multicolumn{8}{l}{\textit{Approximate Methods}} \\
        \addlinespace[2pt]
        CTSketch               & $5.1 \msmall{\pm 0.3}$ & $5.1 \msmall{\pm 0.3}$ & $5.3 \msmall{\pm 0.3}$ & $5.7 \msmall{\pm 0.5}$ & $5.9 \msmall{\pm 0.6}$ & $\mathbf{97.7 \msmall{\pm 0.4}}$ & $6.1 \msmall{\pm 0.6}$ \\
        CTSketch (Grid)        & $0.0 \msmall{\pm 0.0}$ & $0.0 \msmall{\pm 0.0}$ & $0.0 \msmall{\pm 0.0}$ & $0.0 \msmall{\pm 0.0}$ & $0.0 \msmall{\pm 0.0}$ & $96.7 \msmall{\pm 0.8}$ & $0.0 \msmall{\pm 0.0}$ \\
        Scallop                & T/O & T/O & T/O & T/O & T/O & T/O & T/O \\
        DSL                    & $4.8 \msmall{\pm 0.5}$ & $4.8 \msmall{\pm 0.5}$ & $4.9 \msmall{\pm 0.5}$ & $5.4 \msmall{\pm 0.6}$ & $5.6 \msmall{\pm 0.6}$ & $97.6 \msmall{\pm 0.1}$ & $5.8 \msmall{\pm 0.7}$ \\
        \addlinespace[6pt]
        \rowcolor{ablationshade}\multicolumn{8}{l}{\textit{Ablation Studies}} \\
        \addlinespace[2pt]
        \layer{} (Train Only)  & $5.7 \msmall{\pm 0.2}$ & $5.7 \msmall{\pm 0.2}$ & $6.1 \msmall{\pm 0.2}$ & $7.2 \msmall{\pm 0.5}$ & $7.8 \msmall{\pm 0.8}$ & $97.3 \msmall{\pm 0.0}$ & $9.4 \msmall{\pm 1.6}$ \\
        \layer{} (Inf. Only)   & $7.4 \msmall{\pm 0.4}$ & $11.5 \msmall{\pm 0.8}$ & $16.6 \msmall{\pm 0.8}$ & $29.7 \msmall{\pm 1.0}$ & $39.7 \msmall{\pm 0.9}$ & $97.2 \msmall{\pm 0.3}$ & $\mathbf{100.0 \msmall{\pm 0.0}}$ \\
        \layer{} (w/o $\ell_2$) & $7.0 \msmall{\pm 0.1}$ & $11.0 \msmall{\pm 1.4}$ & $16.7 \msmall{\pm 1.9}$ & $28.6 \msmall{\pm 2.4}$ & $38.3 \msmall{\pm 3.1}$ & $97.0 \msmall{\pm 0.3}$ & $\mathbf{100.0 \msmall{\pm 0.0}}$ \\
        \midrule
        \layer{}               & $\mathbf{7.8 \msmall{\pm 0.6}}$ & $\mathbf{12.1 \msmall{\pm 1.1}}$ & $\mathbf{18.0 \msmall{\pm 1.3}}$ & $\mathbf{30.4 \msmall{\pm 0.6}}$ & $\mathbf{40.5 \msmall{\pm 1.4}}$ & $97.1 \msmall{\pm 0.2}$ & $\mathbf{100.0 \msmall{\pm 0.0}}$ \\
        \bottomrule
    \end{tabular}
\end{table}

\subsection{MNIST-Arithmetic}\label{app:mnist-sum-results}
\paragraph{Task Details}
We consider two fully supervised tasks based on the MNIST Dataset \citep{lecun2002gradient} which is distributed under the CC BY-SA 3.0 license. For MNIST-Sum, we consider the problem of computing the sum of $n$ handwritten digits sampled from the dataset, and test using sums of size $n = 2^k$ for $k \in \{1, 4, 6, 8, 10\}$. Each sample consists of $n$ single-digit images $(x_1, \ldots, x_n)$ with labels $(Y_1, \ldots, Y_n) \in \{0, \ldots, 9\}^n$ and sum $T \in \{0, \ldots, 9n\}$. The enforced constraint is:
\begin{equation*}
    \sum_{i=1}^{n} Y_i = T.
\end{equation*}

For MNIST-Add, we consider the problem of computing the sum of two $n$-digit numbers represented as sequences of handwritten digits. We test using $n \in \{2, 15, 100, 256\}$. Each sample consists of $2n$ single-digit images $(x_1, \ldots, x_{2n})$ encoding two operands, with labels $(A_1, \ldots, A_n) \in \{0, \ldots, 9\}^n$ and $(B_1, \ldots, B_n) \in \{0, \ldots, 9\}^n$, and an $(n+1)$-digit sum with labels $(S_0, S_1, \ldots, S_n) \in \{0, \ldots, 9\}^{n+1}$, where $A_1$, $B_1$, and $S_0$ denote the most significant digits. The constraint requires the predicted $(n+1)$-digit output to equal the sum of the two input numbers:
\begin{equation*}
    \sum_{i=1}^{n} A_i \cdot 10^{n-i} + \sum_{i=1}^{n} B_i \cdot 10^{n-i} = \sum_{i=0}^{n} S_i \cdot 10^{n-i}.
\end{equation*}

\paragraph{Experimental Setup} To ensure a fair comparison with existing baselines we follow the dataset construction of \citep{choi2025ctsketch}: 5k training and 1k test examples for every $n$ on MNIST-Sum, and $\lfloor 60000/2n \rfloor$ training and $\lfloor 10000/2n \rfloor$ test examples on MNIST-Add. To test scalability we sweep  $n \in \{2, 16, 64, 256, 1024\}$ for MNIST-Sum and $N \in \{2, 15, 100, 256\}$ for MNIST-Add. The neural backbone is shared across both tasks: a LeNet-5 digit encoder which we train with Adam \citep{kingma2014adam} at a learning rate of $1\mathrm{e}{-}3$ (reduced to $3\mathrm{e}{-}4$ for MNIST-Add at $n \in \{100, 256\}$) under a cosine schedule ($\eta_{\min} = 1\mathrm{e}{-}5$). We set batch size 16 for MNIST-Sum and 64 for MNIST-Add, and for each task size $n$ we record mean $\pm$ std over 5 seeds for task accuracy (predicted output exactly matches ground truth) and consistency (predicted digits sum to the predicted output). We present the full set of results for MNIST-Sum in Table~\ref{tab:mnist_sum} and for MNIST-Add in Table~\ref{tab:mnist_add}; these results are visualised in Figure~\ref{fig:mnist-scaling}.
Table~\ref{tab:mnist_add_breakdown} further breaks down MNIST-Add accuracy into operand and sum accuracy, reporting both exact-match and per-digit
(macro) accuracy.

\paragraph{Implementation Details} We compare \layer{} against seven neurosymbolic baselines covering both exact and approximate methods: DeepProbLog (DPL) \citep{manhaeve2018deepproblog}, Semantic Probabilistic Layers (SPL) \citep{ahmed2022semantic}, CCN+ \citep{giunchiglia2024ccn+}, CTSketch \citep{choi2025ctsketch}, A-NeSI \citep{van2023nesi}, Scallop \citep{huang2021scallop}, and DeepSoftLog (DSL) \citep{maene2023soft}. All methods are trained and evaluated in a fully-supervised setting but differ in how the per-digit predictions are aggregated to give the final output. For MNIST-Sum, \layer{}, A-NeSI, and CCN+ use a learned MLP sum-head, while SPL reads the sum directly from the joint distribution over digits and sum encoded in its compiled SDD. CTSketch, Scallop, DSL, and DPL take a per-digit argmax at inference and compute the sum deterministically, so their predictions are consistent by construction. For MNIST-Add, \layer{} and CTSketch use a carry-chain decomposition, processing digit positions sequentially and propagating the carry from each position into the next. Consistent with \citep{van2023nesi}, A-NeSI uses an autoregressive MLP head over the multi-digit output. CCN+ uses a MLP head, as its binary one-hot representation does not support carry-chain decomposition. Scallop, DPL, and DSL evaluate the addition relation as a logic program, while SPL compiles it into a SDD over the joint distribution of digits and sum.

For MNIST-Sum, Scallop uses top-3 proofs with learning rate $1\mathrm{e}{-}3$; CTSketch uses tensor-train decomposition with rank 2 and tree depth $1, 4, 6, 8, 10$ for $n = 2, 16, 64, 256, 1024$ respectively; DSL uses neural embeddings for $n=16$ with learning rate $1\mathrm{e}{-}3$ and weight decay $1\mathrm{e}{-}5$; and A-NeSI uses parameters consistent with \citep{van2023nesi}. For MNIST-Add, Scallop again uses top-3 proofs with learning rate $1\mathrm{e}{-}3$; CTSketch uses tensor-train decomposition with rank 2; and A-NeSI and DSL use the hyperparameters from their respective papers. The original CTSketch implementation reconstructs predicted digits via base-10 expansion over int64, we observe overflow for $n \geq 19$ and report results using a fixed arbitrary-precision implementation. A-NeSI's symbolic function overflows int64 for $n \geq 19$; we therefore report A-NeSI only up to $n = 15$.%

Table~\ref{tab:minedit-ablation} reports an ablation conducted on the MNIST-Sum task to isolate the contribution of training with \layer{} versus applying it only at inference: digit accuracy is essentially unchanged across configurations, confirming that \layer{} does not degrade the perception backbone, while inference-only application already recovers most of the sum-accuracy gain.

\paragraph{Computational Efficiency}  Table~\ref{tab:mnist_sum_efficiency} reports peak GPU memory and per-epoch training time on MNIST-Sum across all scales. Among exact methods, only \layer{} scales beyond $n=2$, and it reaches $n=1024$ while remaining competitive with the approximate methods on both memory and time, using substantially less memory than A-NeSI and running considerably faster at the largest scale. CTSketch is the most resource-efficient baseline, but at the cost of accuracy (Table~\ref{tab:mnist_sum}). The cvxpy variant times out at $n=1024$, confirming that the closed-form surrogate is necessary for scaling while retaining the constraint satisfaction guarantee.

\begin{table}[t]
    \centering
    \caption{Test accuracy (\%) on the MNISTSum task (sum$_n$) for $n \in \{2, 16, 64, 256, 1024\}$. T/O = timeout. The reference approximates the accuracy of a neural predictor for MNIST images with 0.995 accuracy, computed as $0.995^n$.}
    \label{tab:mnist_sum}
    \setlength{\tabcolsep}{4pt}
    \definecolor{exactshade}{RGB}{252, 228, 230}%
    \definecolor{approxshade}{RGB}{223, 237, 247}%
    \begin{tabular}{lccccc}
        \toprule
        \textbf{Method} & \textbf{sum$_{2}$} & \textbf{sum$_{16}$} & \textbf{sum$_{64}$} & \textbf{sum$_{256}$} & \textbf{sum$_{1024}$} \\
        \midrule
  
        \addlinespace[2pt]
        \rowcolor{exactshade}\multicolumn{6}{l}{\textit{Exact Methods}} \\
        \addlinespace[2pt]
        DPL              & $97.2 \msmall{\pm 0.5}$ & T/O              & T/O              & T/O              & T/O             \\
        SPL              & $96.7 \msmall{\pm 0.5}$ & T/O              & T/O              & T/O              & T/O             \\
        CCN+             & $\mathbf{97.4 \msmall{\pm 0.7}}$ & T/O              & T/O              & T/O              & T/O             \\
        \addlinespace[6pt]
        \rowcolor{approxshade}\multicolumn{6}{l}{\textit{Approximate Methods}} \\
        \addlinespace[2pt]
        A-NeSI           & $96.0 \msmall{\pm 0.2}$ & $27.8 \msmall{\pm 4.6}$ & $12.9 \msmall{\pm 1.6}$ & $5.1 \msmall{\pm 0.1}$  & $2.0 \msmall{\pm 0.2}$ \\
        Scallop          & $95.5 \msmall{\pm 1.1}$ & $81.7 \msmall{\pm 3.6}$ & T/O              & T/O              & T/O             \\
        DSL              & $97.3 \msmall{\pm 0.8}$ & $5.5 \msmall{\pm 0.9}$  & T/O              & T/O              & T/O             \\
        CTSketch         & $95.0 \msmall{\pm 1.1}$ & $82.4 \msmall{\pm 0.4}$ & $51.7 \msmall{\pm 1.3}$ & $8.4 \msmall{\pm 0.3}$  & $2.3 \msmall{\pm 0.1}$ \\
        \midrule
        {\layer{}} (\textit{cvxpy}) & $95.5 \msmall{\pm 0.6}$ & $84.4 \msmall{\pm 0.5}$ & $\mathbf{54.0 \msmall{\pm 2.5}}$ & $10.6 \msmall{\pm 2.0}$ & T/O \\
        {\layer{} (Ours)}     & $95.9 \msmall{\pm 0.5}$ & $\mathbf{84.5 \msmall{\pm 1.1}}$ & $52.9 \msmall{\pm 2.3}$ & $\mathbf{11.8 \msmall{\pm 1.9}}$ & $\mathbf{2.4 \msmall{\pm 0.7}}$ \\
        \midrule
        Reference      & $99.0$ & $92.3$ & $72.6$ & $27.7$ & $0.6$ \\
        \bottomrule
    \end{tabular}
\end{table}

\begin{table}[t]
    \centering
    \caption{MNIST-Sum Ablation Study.}
    \label{tab:minedit-ablation}
    \setlength{\tabcolsep}{6pt}
    \definecolor{n16shade}{RGB}{252, 228, 230}%
    \definecolor{n64shade}{RGB}{223, 237, 247}%
    \definecolor{n256shade}{RGB}{226, 242, 228}%
    \begin{tabular}{lccc}
        \toprule
        \textbf{Metric} & \textbf{No \layer{}} & \textbf{\layer{} (Inference)} & \textbf{\layer{} (Train + Inference)} \\
        \midrule
        \addlinespace[2pt]
        \rowcolor{n16shade}\multicolumn{4}{l}{\textit{$n = 16$}} \\
        \addlinespace[2pt]
        Digit Acc.\ (\%)    & $98.8\msmall{\pm 0.1}$ & $98.9\msmall{\pm 0.2}$ & $\textbf{99.0\msmall{\pm 0.1}}$ \\
        Sum Acc.\ (\%)      & $4.3\msmall{\pm 0.6}$  & $84.0\msmall{\pm 2.9}$ & $\mathbf{85.8\msmall{\pm 1.6}}$ \\
        Consistency (\%)    & $4.3\msmall{\pm 0.6}$  & $\textbf{100.0\msmall{\pm 0.0}}$       & $\textbf{100.0\msmall{\pm 0.0}}$ \\
        Avg.\ $L_2$ Edit    & N/A            & $0.9$          & $0.3$ \\
        \addlinespace[6pt]
        \rowcolor{n64shade}\multicolumn{4}{l}{\textit{$n = 64$}} \\
        \addlinespace[2pt]
        Digit Acc.\ (\%)    & $98.9\msmall{\pm 0.0}$ & $\textbf{98.8\msmall{\pm 0.1}}$ & $\textbf{98.8\msmall{\pm 0.0}}$ \\
        Sum Acc.\ (\%)      & $1.7\msmall{\pm 0.6}$  & $\mathbf{49.0\msmall{\pm 1.9}}$ & $47.4\msmall{\pm 0.2}$ \\
        Consistency (\%)    & $1.7\msmall{\pm 0.2}$  & $\textbf{100.0\msmall{\pm 0.0}}$       & $\textbf{100.0\msmall{\pm 0.0}}$ \\
        Avg.\ $L_2$ Edit    & N/A            & $0.8$          & $0.4$ \\
        \addlinespace[6pt]
        \rowcolor{n256shade}\multicolumn{4}{l}{\textit{$n = 256$}} \\
        \addlinespace[2pt]
        Digit Acc.\ (\%)    & $\textbf{98.7\msmall{\pm 0.1}}$ & $\textbf{98.7\msmall{\pm 0.0}}$ & $98.5\msmall{\pm 0.4}$ \\
        Sum Acc.\ (\%)      & $0.8\msmall{\pm 0.1}$  & $7.8\msmall{\pm 0.2}$  & $\textbf{10.3\msmall{\pm 1.1}}$ \\
        Consistency (\%)    & $0.7\msmall{\pm 0.4}$  & $\textbf{100.0\msmall{\pm 0.0}}$        & $\textbf{100.0\msmall{\pm 0.0}}$ \\
        Avg.\ $L_2$ Edit    & N/A            & $0.9$          & $0.7$ \\
        \bottomrule
    \end{tabular}%
\end{table}

\begin{table}[t]
    \centering
    \caption{Test accuracy (\%) on the MNISTAdd task (add$_n$). Accuracy
    is the exact-match accuracy of the predicted sum: all $n{+}1$ digits must be correct. T/O = timeout;
    ERR = numerical error. 
    The reference approximates the accuracy of a neural predictor
    with 0.995 accuracy per digit, computed as $0.995^{2n}$.}
    \label{tab:mnist_add}
    \setlength{\tabcolsep}{4pt}
    \definecolor{exactshade}{RGB}{252, 228, 230}%
    \definecolor{approxshade}{RGB}{223, 237, 247}%
    \begin{tabular}{lcccccc}
        \toprule
        \textbf{Method} & \textbf{add$_{1}$} & \textbf{add$_{2}$} & \textbf{add$_{4}$} & \textbf{add$_{15}$} & \textbf{add$_{100}$} & \textbf{add$_{256}$} \\
        \midrule
        \addlinespace[2pt]
        \rowcolor{exactshade}\multicolumn{7}{l}{\textit{Exact Methods}} \\
        \addlinespace[2pt]
        DPL              & $97.1 \msmall{\pm 0.2}$ & $89.5 \msmall{\pm 0.6}$ & T/O & T/O & T/O & T/O \\
        SPL              & $96.7 \msmall{\pm 0.6}$ & T/O & T/O & T/O & T/O & T/O \\
        CCN+             & $98.4 \msmall{\pm 0.2}$ & T/O & T/O & T/O & T/O & T/O \\
        \addlinespace[6pt]
        \rowcolor{approxshade}\multicolumn{7}{l}{\textit{Approximate Methods}} \\
        \addlinespace[2pt]
        A-NeSI           & $97.4 \msmall{\pm 0.2}$ & $96.3 \msmall{\pm 0.4}$ & $93.1 \msmall{\pm 0.7}$ & $76.9 \msmall{\pm 2.3}$ & ERR & ERR \\
        Scallop          & $95.3 \msmall{\pm 0.7}$ & $95.9 \msmall{\pm 0.4}$ & T/O & T/O & T/O & T/O \\
        DSL              & $98.1 \msmall{\pm 0.2}$ & $96.6 \msmall{\pm 0.4}$ & $92.6 \msmall{\pm 0.7}$ & $\mathbf{77.8 \msmall{\pm 2.2}}$ & $\mathbf{28.0 \msmall{\pm 6.4}}$ & T/O \\
        CTSketch         & $98.1 \msmall{\pm 0.2}$ & $96.8 \msmall{\pm 0.2}$ & $92.6 \msmall{\pm 1.8}$ & $72.5 \msmall{\pm 2.3}$ & $8.5 \msmall{\pm 4.9}$ & $0.0 \msmall{\pm 0.0}$ \\
        \midrule
        {\layer{}} (\textit{cvxpy}) & $\mathbf{98.5 \msmall{\pm 0.1}}$ & $97.2 \msmall{\pm 0.2}$ & $93.3 \msmall{\pm 0.3}$ & $71.8 \msmall{\pm 1.6}$ & T/O & T/O \\
        {\layer{} (Ours)} & $98.4 \msmall{\pm 0.1}$ & $\mathbf{97.3 \msmall{\pm 0.2}}$ & $\mathbf{93.7 \msmall{\pm 0.7}}$ & $73.7 \msmall{\pm 0.3}$ & $13.8 \msmall{\pm 1.9}$ & $\mathbf{0.6 \msmall{\pm 0.5}}$ \\
        \midrule
        Reference        & $99.0$ & $98.0$ & $96.1$ & $86.0$ & $36.7$ & $7.7$ \\
        \bottomrule
    \end{tabular}%
\end{table}

\begin{table}[t]
    \centering
    \providecolor{exactbar}{RGB}{226, 120, 135}
    \providecolor{approxbar}{RGB}{110, 160, 210}
    \providecommand{\catbar}[1]{\textcolor{#1}{\rule[-\dp\strutbox]{2pt}{\dimexpr\ht\strutbox+\dp\strutbox\relax}}}
    \providecommand{\exbar}{\catbar{exactbar}}
    \providecommand{\apbar}{\catbar{approxbar}}
    \caption[Operand and sum accuracy on MNISTAdd]{Operand and sum accuracy (\%) on MNISTAdd. \textit{Exact}: no
    digit errors; \textit{Macro}: per-digit accuracy. \textcolor{exactbar}{\rule{2pt}{1.6ex}}~exact,
    \textcolor{approxbar}{\rule{2pt}{1.6ex}}~approximate methods; unlisted methods time out (A-NeSI: ERR for $n \geq 100$).}
    \label{tab:mnist_add_breakdown}
    \setlength{\tabcolsep}{4pt}
    \begin{tabular}{l@{\hspace{6pt}}c@{\hspace{4pt}}lcccccc}
        \toprule
         & & & \multicolumn{2}{c}{\textbf{Operand 1}} & \multicolumn{2}{c}{\textbf{Operand 2}} & \multicolumn{2}{c}{\textbf{Sum}} \\
        \cmidrule(lr){4-5}\cmidrule(lr){6-7}\cmidrule(lr){8-9}
        $\boldsymbol{n}$ & & \textbf{Method} & \textbf{Exact} & \textbf{Macro} & \textbf{Exact} & \textbf{Macro} & \textbf{Exact} & \textbf{Macro} \\
        \midrule
         & \exbar & DPL & $98.4 \msmall{\pm 0.2}$ & $98.4 \msmall{\pm 0.2}$ & $98.5 \msmall{\pm 0.2}$ & $98.5 \msmall{\pm 0.2}$ & $97.1 \msmall{\pm 0.2}$ & $97.9 \msmall{\pm 0.2}$ \\
         & \exbar & SPL & $98.5 \msmall{\pm 0.4}$ & $98.5 \msmall{\pm 0.4}$ & $98.1 \msmall{\pm 0.4}$ & $98.1 \msmall{\pm 0.4}$ & $96.7 \msmall{\pm 0.6}$ & $97.6 \msmall{\pm 0.4}$ \\
         & \exbar & CCN+ & $\mathbf{99.2 \msmall{\pm 0.1}}$ & $\mathbf{99.2 \msmall{\pm 0.1}}$ & $99.0 \msmall{\pm 0.1}$ & $99.0 \msmall{\pm 0.1}$ & $98.4 \msmall{\pm 0.2}$ & $\mathbf{98.9 \msmall{\pm 0.1}}$ \\
         & \apbar & A-NeSI & $98.6 \msmall{\pm 0.2}$ & $98.6 \msmall{\pm 0.2}$ & $98.7 \msmall{\pm 0.2}$ & $98.7 \msmall{\pm 0.2}$ & $97.4 \msmall{\pm 0.2}$ & $98.2 \msmall{\pm 0.2}$ \\
         & \apbar & Scallop & $97.7 \msmall{\pm 0.5}$ & $97.7 \msmall{\pm 0.5}$ & $97.5 \msmall{\pm 0.5}$ & $97.5 \msmall{\pm 0.5}$ & $95.3 \msmall{\pm 0.7}$ & $96.6 \msmall{\pm 0.5}$ \\
         & \apbar & DSL & $99.0 \msmall{\pm 0.1}$ & $99.0 \msmall{\pm 0.1}$ & $99.1 \msmall{\pm 0.1}$ & $99.1 \msmall{\pm 0.1}$ & $98.1 \msmall{\pm 0.2}$ & $98.6 \msmall{\pm 0.1}$ \\
         & \apbar & CTSketch & $99.1 \msmall{\pm 0.1}$ & $99.1 \msmall{\pm 0.1}$ & $99.0 \msmall{\pm 0.1}$ & $99.0 \msmall{\pm 0.1}$ & $98.1 \msmall{\pm 0.2}$ & $98.6 \msmall{\pm 0.1}$ \\
        \cmidrule(l){2-9}
         &  & {\layer{}} (\textit{cvxpy}) & $\mathbf{99.2 \msmall{\pm 0.1}}$ & $\mathbf{99.2 \msmall{\pm 0.1}}$ & $\mathbf{99.2 \msmall{\pm 0.2}}$ & $\mathbf{99.2 \msmall{\pm 0.2}}$ & $\mathbf{98.5 \msmall{\pm 0.1}}$ & $\mathbf{98.9 \msmall{\pm 0.1}}$ \\
        \multirow{-9}{*}{\textbf{add$_{1}$}} &  & {\layer{} (Ours)} & $\mathbf{99.2 \msmall{\pm 0.1}}$ & $\mathbf{99.2 \msmall{\pm 0.1}}$ & $\mathbf{99.2 \msmall{\pm 0.1}}$ & $\mathbf{99.2 \msmall{\pm 0.1}}$ & $98.4 \msmall{\pm 0.1}$ & $98.8 \msmall{\pm 0.0}$ \\
        \midrule
         & \exbar & DPL & $94.9 \msmall{\pm 0.4}$ & $97.4 \msmall{\pm 0.2}$ & $94.2 \msmall{\pm 0.5}$ & $97.0 \msmall{\pm 0.2}$ & $89.5 \msmall{\pm 0.6}$ & $95.2 \msmall{\pm 0.3}$ \\
         & \apbar & A-NeSI & $98.0 \msmall{\pm 0.3}$ & $99.0 \msmall{\pm 0.1}$ & $98.2 \msmall{\pm 0.3}$ & $99.1 \msmall{\pm 0.1}$ & $96.3 \msmall{\pm 0.4}$ & $98.3 \msmall{\pm 0.2}$ \\
         & \apbar & Scallop & $98.2 \msmall{\pm 0.3}$ & $99.1 \msmall{\pm 0.1}$ & $97.6 \msmall{\pm 0.3}$ & $98.8 \msmall{\pm 0.2}$ & $95.9 \msmall{\pm 0.4}$ & $98.2 \msmall{\pm 0.2}$ \\
         & \apbar & DSL & $98.4 \msmall{\pm 0.2}$ & $99.2 \msmall{\pm 0.1}$ & $98.2 \msmall{\pm 0.3}$ & $99.1 \msmall{\pm 0.1}$ & $96.6 \msmall{\pm 0.4}$ & $98.6 \msmall{\pm 0.2}$ \\
         & \apbar & CTSketch & $98.2 \msmall{\pm 0.1}$ & $99.1 \msmall{\pm 0.1}$ & $98.6 \msmall{\pm 0.2}$ & $\mathbf{99.3 \msmall{\pm 0.1}}$ & $96.8 \msmall{\pm 0.2}$ & $98.4 \msmall{\pm 0.1}$ \\
        \cmidrule(l){2-9}
         &  & {\layer{}} (\textit{cvxpy}) & $98.5 \msmall{\pm 0.2}$ & $\mathbf{99.3 \msmall{\pm 0.1}}$ & $98.4 \msmall{\pm 0.1}$ & $99.2 \msmall{\pm 0.1}$ & $97.2 \msmall{\pm 0.2}$ & $\mathbf{98.8 \msmall{\pm 0.1}}$ \\
        \multirow{-7}{*}{\textbf{add$_{2}$}} &  & {\layer{} (Ours)} & $\mathbf{98.6 \msmall{\pm 0.2}}$ & $\mathbf{99.3 \msmall{\pm 0.1}}$ & $\mathbf{98.7 \msmall{\pm 0.1}}$ & $\mathbf{99.3 \msmall{\pm 0.1}}$ & $\mathbf{97.3 \msmall{\pm 0.2}}$ & $98.7 \msmall{\pm 0.1}$ \\
        \midrule
         & \apbar & A-NeSI & $96.2 \msmall{\pm 0.5}$ & $99.0 \msmall{\pm 0.1}$ & $96.7 \msmall{\pm 0.5}$ & $\mathbf{99.2 \msmall{\pm 0.1}}$ & $93.1 \msmall{\pm 0.7}$ & $98.0 \msmall{\pm 0.2}$ \\
         & \apbar & DSL & $95.8 \msmall{\pm 0.6}$ & $98.9 \msmall{\pm 0.1}$ & $96.6 \msmall{\pm 0.5}$ & $\mathbf{99.2 \msmall{\pm 0.1}}$ & $92.6 \msmall{\pm 0.7}$ & $97.8 \msmall{\pm 0.2}$ \\
         & \apbar & CTSketch & $96.0 \msmall{\pm 1.1}$ & $99.0 \msmall{\pm 0.3}$ & $96.5 \msmall{\pm 0.6}$ & $99.1 \msmall{\pm 0.1}$ & $92.6 \msmall{\pm 1.8}$ & $97.9 \msmall{\pm 0.4}$ \\
        \cmidrule(l){2-9}
         &  & {\layer{}} (\textit{cvxpy}) & $96.2 \msmall{\pm 0.6}$ & $99.0 \msmall{\pm 0.2}$ & $96.1 \msmall{\pm 0.8}$ & $99.0 \msmall{\pm 0.2}$ & $93.3 \msmall{\pm 0.3}$ & $98.0 \msmall{\pm 0.0}$ \\
        \multirow{-5}{*}{\textbf{add$_{4}$}} &  & {\layer{} (Ours)} & $\mathbf{96.6 \msmall{\pm 0.3}}$ & $\mathbf{99.1 \msmall{\pm 0.1}}$ & $\mathbf{97.0 \msmall{\pm 0.7}}$ & $\mathbf{99.2 \msmall{\pm 0.2}}$ & $\mathbf{93.7 \msmall{\pm 0.7}}$ & $\mathbf{98.1 \msmall{\pm 0.3}}$ \\
        \midrule
         & \apbar & A-NeSI & $87.1 \msmall{\pm 1.9}$ & $99.0 \msmall{\pm 0.1}$ & $\mathbf{87.4 \msmall{\pm 1.8}}$ & $\mathbf{99.1 \msmall{\pm 0.1}}$ & $76.9 \msmall{\pm 2.3}$ & $97.4 \msmall{\pm 0.3}$ \\
         & \apbar & DSL & $\mathbf{88.9 \msmall{\pm 1.7}}$ & $\mathbf{99.2 \msmall{\pm 0.1}}$ & $86.5 \msmall{\pm 1.9}$ & $99.0 \msmall{\pm 0.1}$ & $\mathbf{77.8 \msmall{\pm 2.2}}$ & $\mathbf{97.7 \msmall{\pm 0.3}}$ \\
         & \apbar & CTSketch & $84.8 \msmall{\pm 3.2}$ & $98.9 \msmall{\pm 0.2}$ & $85.4 \msmall{\pm 0.2}$ & $99.0 \msmall{\pm 0.1}$ & $72.5 \msmall{\pm 2.3}$ & $97.2 \msmall{\pm 0.2}$ \\
        \cmidrule(l){2-9}
         &  & {\layer{}} (\textit{cvxpy}) & $85.8 \msmall{\pm 0.7}$ & $99.0 \msmall{\pm 0.1}$ & $82.7 \msmall{\pm 1.2}$ & $98.7 \msmall{\pm 0.1}$ & $71.8 \msmall{\pm 1.6}$ & $97.0 \msmall{\pm 0.2}$ \\
        \multirow{-5}{*}{\textbf{add$_{15}$}} &  & {\layer{} (Ours)} & $88.6 \msmall{\pm 1.0}$ & $\mathbf{99.2 \msmall{\pm 0.0}}$ & $86.0 \msmall{\pm 0.6}$ & $99.0 \msmall{\pm 0.0}$ & $73.7 \msmall{\pm 0.3}$ & $97.6 \msmall{\pm 0.1}$ \\
        \midrule
         & \apbar & DSL & $\mathbf{38.0 \msmall{\pm 6.8}}$ & $98.7 \msmall{\pm 0.2}$ & $\mathbf{46.0 \msmall{\pm 7.1}}$ & $98.7 \msmall{\pm 0.2}$ & $\mathbf{28.0 \msmall{\pm 6.4}}$ & $96.6 \msmall{\pm 0.4}$ \\
         & \apbar & CTSketch & $30.5 \msmall{\pm 0.7}$ & $98.9 \msmall{\pm 0.0}$ & $30.8 \msmall{\pm 4.6}$ & $98.9 \msmall{\pm 0.1}$ & $8.5 \msmall{\pm 4.9}$ & $96.9 \msmall{\pm 0.3}$ \\
        \cmidrule(l){2-9}
        \multirow{-3}{*}{\textbf{add$_{100}$}} &  & {\layer{} (Ours)} & $37.9 \msmall{\pm 2.6}$ & $\mathbf{99.0 \msmall{\pm 0.0}}$ & $39.4 \msmall{\pm 1.8}$ & $\mathbf{99.0 \msmall{\pm 0.1}}$ & $13.8 \msmall{\pm 1.9}$ & $\mathbf{97.2 \msmall{\pm 0.2}}$ \\
        \midrule
         & \apbar & CTSketch & $4.6 \msmall{\pm 2.2}$ & $\mathbf{98.8 \msmall{\pm 0.0}}$ & $5.7 \msmall{\pm 0.7}$ & $\mathbf{98.9 \msmall{\pm 0.0}}$ & $0.0 \msmall{\pm 0.0}$ & $\mathbf{96.9 \msmall{\pm 0.1}}$ \\
        \cmidrule(l){2-9}
        \multirow{-2}{*}{\textbf{add$_{256}$}} &  & {\layer{} (Ours)} & $\mathbf{8.9 \msmall{\pm 3.0}}$ & $\mathbf{98.8 \msmall{\pm 0.2}}$ & $\mathbf{6.1 \msmall{\pm 0.6}}$ & $98.8 \msmall{\pm 0.2}$ & $\mathbf{0.6 \msmall{\pm 0.5}}$ & $96.6 \msmall{\pm 0.6}$ \\
        \bottomrule
    \end{tabular}
\end{table}

\begin{table}[t]
    \centering
    \caption{Computational efficiency on MNIST-Sum across methods and scales.}
    \label{tab:mnist_sum_efficiency}
    \setlength{\tabcolsep}{3pt}
    \definecolor{exactshade}{RGB}{252, 228, 230}%
    \definecolor{approxshade}{RGB}{223, 237, 247}%
    \begin{tabular}{l ccccc ccccc}
        \toprule
        & \multicolumn{5}{c}{\textbf{Peak Memory (MB)}} & \multicolumn{5}{c}{\textbf{Epoch Time (s)}} \\
        \cmidrule(lr){2-6} \cmidrule(lr){7-11}
        \textbf{Method / $n$} & $2$ & $16$ & $64$ & $256$ & $1024$ & $2$ & $16$ & $64$ & $256$ & $1024$ \\
        \midrule
        \addlinespace[2pt]
        \rowcolor{exactshade}\multicolumn{11}{l}{\textit{Exact Methods}} \\
        \addlinespace[2pt]
        DPL       & 20.1  & T/O   & T/O    & T/O    & T/O    & 58.4  & T/O   & T/O   & T/O   & T/O   \\
        SPL       & 39.8  & T/O   & T/O    & T/O    & T/O    & 4.1   & T/O   & T/O   & T/O   & T/O   \\
        CCN+      & 56.7  & T/O   & T/O    & T/O    & T/O    & 26.9  & T/O   & T/O   & T/O   & T/O   \\
        \addlinespace[6pt]
        \rowcolor{approxshade}\multicolumn{11}{l}{\textit{Approximate Methods}} \\
        \addlinespace[2pt]
        A-NeSI    & 80.0  & 209.5 & 1014.1 & 5680.1 & 13157.0 & 11.8 & 24.7 & 66.4 & 232.3 & 995.8 \\
        Scallop $^{\dagger}$   & \textbf{18.9} & 57.1 & T/O & T/O & T/O & 3100.0 & 3400.0 & T/O & T/O & T/O \\
        DSL       & 20.6  & 35.4  & T/O    & T/O    & T/O    & 33.3  & 386.6 & T/O   & T/O   & T/O   \\
        CTSketch  & 19.0  & \textbf{28.6}  & \textbf{106.1}  & \textbf{367.7}  & \textbf{1421.0} & \textbf{0.6}   & \textbf{4.2}   & \textbf{14.9}  & \textbf{60.9}  & \textbf{239.2} \\
        \midrule
        {\layer{}} (\textit{cvxpy})$^\ddagger$ & 20.3 & 31.7 & 123.7 & 529.4 & T/O & 33.7 & 94.8 & 257.6 & 815.2 & T/O \\
        {\layer{}(Ours)} & 20.3 & 31.7 & 123.7 & 529.4 & 1866.5 & 1.5 & 5.9 & 17.5 & 62.3 & 240.3 \\
        \bottomrule
    \end{tabular}%
    \par\vspace{2pt}
    {\footnotesize
    $^\dagger$\,Scallop runs on CPU. \quad
    $^\ddagger$\,The \textit{cvxpy} projection runs on CPU\par}
\end{table}

\subsection{Sudoku solving.}\label{app:experiments_sudoku}   

\paragraph{Task Details} We consider the sudoku dataset from SatNet~\citep{wang2019satnet},
consisting of $10{,}000$ nine-by-nine Sudoku boards whose hint cells are
rendered as $28\!\times\!28$ grayscale MNIST digits
and whose blank cells are represented as uniformly grey images.                            
The average number of given hints per board is $36.22$ (with a minimum of $31$, and maximum $42$).                              
The task is to predict the complete $9\!\times\!9$ solution, inserting one digit from                
$\{1,\ldots,9\}$ per cell---from the $81$-cell image grid, while abiding by the standard Sudoku rules:                                                           
a board is accepted as correct only when every row, every column, and every
$3\!\times\!3$ sub-grid contains each digit exactly once.
The dataset can be downloaded from the~\href{https://powei.tw/sudoku.zip }{official repository}.
No explicit license for the dataset was provided.

\paragraph{Experimental Setup}                                                                                                                                                      
  The $10{,}000$ boards are split into $9{,}000$ training, $500$ validation,                                
  and $500$ test examples following the standard SatNet partition.                                                    
  All hyperparameter and checkpoint selection is performed on the validation                                                     
  split; final results are reported on the fixed test split. Every method is run with $10$ random seeds and we report mean and standard
  deviation across seeds for each metric we consider.
  The primary metric is \textit{PipeAcc}: the fraction of test boards for                                                                                
  which the complete pipeline output exactly matches the ground-truth solution.
  We additionally report the \textit{NeuroAcc}: board accuracy of the                                                                                            
  argmax of the soft board distribution $P\!\in\!\mathbb{R}^{81\times 9}$                                                                                          
  produced by component~(ii) \emph{before} masking and symbolic completion,
  isolating the contribution of the learned upstream components from that of the
  Prolog solver. Finally, we report the percentage of CNN errors, that the board solver is able to detect and recover from, out of the total number of
  CNN errors across the test set (\textit{RecRate}). Results are shown in Table~\ref{tab:sudoku}.

  At evaluation time, methods that produce a scalar mask-logit bias (\layer{}                                                                                      
  and its LLM variant) select the bias from the grid $\{-2.0,-1.0,0.0,1.0,2.0\}$
  that maximises board accuracy on the validation set; for the remaining methods
  (CTSketch and A-NesI) the bias is swept over $\{-1.0,0.0,1.0\}$ and the results
  for the best performing evaluation run is reported.

  \paragraph{Implementation Details}                                                                                                                               
  \subparagraph{Shared backbones.}                                                                                                                           
  All six methods share the same three-component backbone, initialised from the
  publicly released NASR checkpoint~\citep{cornelio2023learning}. Specifically these are as follows:                                                                                                    
  \begin{compactitem}
    \item The \emph{perception module} is a convolutional network that processes each of                                                                           
    the $81$ cell images independently and outputs per-cell log-posteriors over
    the ten classes $\{0,1,\ldots,9\}$ (class $0$ denotes a blank cell).                                                                                                         
    \item The \emph{board-solver} is a Vision Transformer                                                                                             
    with embedding dimension $192$, depth $4$, $3$ attention heads, and MLP                                                                                     
    expansion ratio $4$; it maps the $81$ per-cell ten-dimensional inputs to a                                                                                     
    soft board distribution $P\!\in\!\mathbb{R}^{81\times 9}$ over $\{1,\ldots,9\}$.                                                                                                                                                    
    \item The \emph{masking network} shares the same Transformer configuration with a
    scalar output per cell; sigmoid-activating those logits yields the per-cell
    keep probability used to gate Prolog.
  \end{compactitem} 

\subparagraph{LLM fine-tuning.}                                                                                                                                          
  For the LLM-backed variants, we fine-tune Qwen2.5-14B-Instruct on the $9{,}000$ SatNet training boards using                                        
  LoRA~\citep{lora} with rank $r=16$, scaling $\alpha=16$, no dropout, applied to all attention and MLP layers.
  The base model is loaded in 4-bit NormalFloat quantisation with bfloat16 activations (via Unsloth). 
  We train for $3$ epochs with AdamW-8bit, learning rate $2\!\times\!10^{-4}$, cosine decay schedule, warmup ratio
  $0.05$, weight decay $0.01$, and an effective batch size of $16$ (providing two examples per device
  with eight gradient-accumulation steps). Each board is formatted under the Qwen2.5 chat template: the system message
  states the \textssc{'AllDifferent'} rules and specifies a structured output format; the                                                                          
  user message provides the full $9\!\times\!9$ grid with zeros marking blank cells.
  The model is trained to output a \texttt{<WORKING>/\allowbreak<SOLUTION>}
  block containing exactly nine lines of nine space-separated digits. At inference time, outputs that violate the format, contain out-of-range                                                                                    
  digits, or alter any given cell are treated as failures.

  The fine-tuned LLM produces one-hot-like logits---high confidence on the
  predicted digit and zero elsewhere---carrying no per-cell uncertainty.                                                                                         
  To recover calibrated soft probabilities compatible with downstream masking,
  the frozen neuro-solver from~\cite{cornelio2023learning} is used as an auxiliary
  uncertainty prior for \layer{}. Concretely, we introduce a new \emph{fusion head}, 
  implemented as a residual MLP (with a total of ${\approx}21{,}700$ parameters)
  operating on the individual cells receiving both the LLM-solved boards and the 
  frozen standard neuro-solver soft probabilities, modelling prior prediction uncertainty.
  Across all LLM-backed methods the CNN and neuro-solver are frozen throughout;           
  only the fusion head (and, where applicable, the masking network) receive gradient updates.

  \subparagraph{\layer{}.}                                                                                                                                                  
  \textit{Neuro-board solver from~\citep{cornelio2023learning}.}
  Only the masking network is trained; the CNN and board solver remain at their pre-trained weights throughout.
  At each training step, the solver distribution is projected through $\layer{}$                                                                      
  constraint layer. The masking network is trained with a per-cell weighted binary cross-entropy
  loss: the target for cell $j$ is $1$ if and only if the projected digit equals                                                                      
  the ground-truth, and $0$ otherwise.                                                            
  Incorrectly projected cells (the minority class, typically ${<}10\%$) receive
  loss weight $20$ to compensate for the severe class imbalance.                                                                                        
  We use AdamW with learning rate $10^{-4}$, weight decay $0.3$, cosine LR                                                                                    
  schedule, gradient clipping at norm $1.0$, batch size $200$, up to $10$
  epochs, and early stopping on validation board accuracy at patience $5$.
  
  \textit{LLM board solver.}                                                                                                          
  The fusion head and masking network are trained jointly.
  Training is cache-only: a full-data generation pass is run once to populate an on-disk cache of LLM outputs. 
  The loss combines the same weighted BCE mask term ($\lambda_{\text{mask}}=1.0$) 
  and a label-smoothed NLL calibration loss on the fused distribution       
  ($\lambda_{\text{temp}}=0.5$, smoothing $\alpha=0.3$).                                                                                                    
  Label smoothing is necessary because the LLM is already highly accurate:
  without it, the NLL drives the fusion temperature toward zero, collapsing the
  soft distribution. With $\alpha=0.3$ the analytical optimum lies at a finite temperature,
  preserving uncertainty. All other hyperparameters match the previous variant, with the training budget         
  extended to $20$ epochs and early stopping patience $7$.                                                            
  The mask logit bias is tuned at evaluation on the grid $\{-2.0,-1.0,0.0,1.0,2.0\}$.                                                                                                                                                 
  \subparagraph{CTSketch.}                                                                                                                                    
  \textit{Neuro-board solver from~\citep{cornelio2023learning}.}
  The full model (CNN, board-solver, masking network) is trained end-to-end.
  The training objective is the CTS tensor-sketch loss, following the code from~\citep{choi2025ctsketch}:
  a differentiable rank-$2$ Tucker approximation of the nine-ary \textsc{AllDifferent}
  constraint-satisfaction probability, summed over all $27$ groups (rows,
  columns, and boxes) and combined with a Prolog-based per-board reward signal.
  We use AdamW with learning rate $5\!\times\!10^{-6}$, weight decay $0.3$, a
  linear learning-rate ramp over all $9$ training epochs (warmup covering the full training duration), and batch size $200$.                                       
    
  \textit{LLM board solver.}                   
  Only the fusion head and masking network are trained; the CNN, board-solver,
  The CTS sketch loss and Prolog reward are applied directly to the fused
  distribution output by the fusion head.                                                                                                       
  We use AdamW with learning rate $10^{-4}$, weight decay $0.3$, cosine LR
  schedule, gradient clipping (norm $1.0$), batch size $200$, $20$ epochs, and
  early stopping on validation reward at patience $7$.                                                                                                       
  
  \subparagraph{A-NesI.}                   
  \textit{Neuro-board solver from~\citep{cornelio2023learning}.}                                                                                                                                                     
  We apply the A-NesI framework~\citep{van2023nesi} with pairwise encoding of
  the \textsc{AllDifferent} constraints.                                         
  The inference model $q$ is a two-layer MLP with hidden size $128$ that
  predicts binary constraint satisfaction for each cell pair; it is optimised
  with Adam at learning rate $10^{-3}$ (MSE loss).                                                                                                              
  The perception backbone (CNN and board-solver) is updated via the
  log-$q$ perception loss with Adam at learning rate $5\!\times\!10^{-5}$.
  Each training step draws $100$ samples from the board distribution~$P$.
  The masking network is frozen during A-NesI training and is used solely at
  evaluation time. We train for $3$ epochs with batch size $50$.
  
  \textit{LLM board solver.}             
  The fusion head takes the role of the perception backbone in the A-NesI
  training loop: it is updated via the log-$q$ perception loss with Adam at
  learning rate $5\!\times\!10^{-5}$, while $q$ is trained identically to the
  neuro counterpart (Adam, lr $10^{-3}$, MSE).                                                                                                         
  The masking network is trained jointly from epoch $1$ via a separate AdamW
  optimiser (learning rate $10^{-4}$, weight decay $0.3$, gradient clipping at
  norm $1.0$). The checkpoint selection uses A-NesI inference success (fraction of validation                  
  batches for which $q$ predicts all pairwise constraints satisfied) as the
  criterion. Training runs for $20$ epochs with early stopping patience $7$.

\section{Choice of the Encoding}
\label{sec:trimmed-steps}
Theorem~\ref{thm:complexity}
bounds the cost of \textsc{Compile} by
\[
\text{time } \mathcal{O}\big((r+n) \cdot k^{w^\star(d)+1}\big), \qquad
\text{space } \mathcal{O}\big(n \cdot k^{w^\star(d)}\big),
\]
with $r$ the number of clauses, $n$ the number of variables,
$k=\max_i h_i$ the largest domain, and $w^\star(d)$ the conflict width of $\Pi$ along the ordering $d$, which coincides with the induced width of the
interaction graph of $\Pi$ along $d$ \citep{Dechter2013}. For highly interconnected problems, $w^\star(d)$ grows with the problem size $N$, which is intractable for compilation, so we instead consider alternative problem settings that keep $w^\star(d)$ bounded. 

In Sudoku solving, compiling the joint constraint set over the whole board is intractable, because every cell shares a group with many others, so the interaction graph is highly interconnected. We instead enforce the all-different constraint on each of the 27 groups (9 rows, 9 columns and 9 boxes) individually, so that each constraint set involves only the 9 cells of one group.

In Warcraft, specifying the task as path-finding with local constraints between neighbouring cells gives an interaction graph where $w^\star(d) \propto N$, making intractable to compile. We instead formulate it as walk-finding, describing the route as a sequence of cells, which gives $w^\star(d)=1$ for every $N$. This is a different problem-setting since walks may revisit a cell, so a simple path is recovered by loop erasure. All baselines are run in the same walk-finding setting (excl. CTSketch), with loop erasure applied to every method's output before computing the metrics.

\paragraph{Warcraft: walk-finding formulation.}
Let $G=(V,E)$ be the four-neighbour grid graph on a $g\times g$ board,
$n=g^2=|V|$, with source $s$ (top-left) and target $t$ (bottom-right).
Write $d(u,v)$ for the graph (Manhattan) distance. Recall that $G$ is
bipartite under the checkerboard two-colouring: every edge joins cells of
different colours, hence a walk of length $\ell$ from $s$ can only end on
a cell $v$ whose distance $d(s,v)$ has the same parity as $\ell$
($\ell \equiv d(s,v) \pmod 2$; e.g.\ a cell at even distance is reachable
only in an even number of moves).

\paragraph{Variables and transitions.}
Fix the horizon $L = n-1$, the maximum number of edges of any simple
path. Introduce multi-valued variables $X_0,\dots,X_L$ with intended
reading ``$X_i$ = the cell occupied at time $i$''. The \emph{absorbing
transition relation} is
\[
  T(u,v) \iff \bigl(u\neq t \wedge \{u,v\}\in E\bigr) \;\lor\; u=v=t ,
\]
i.e.\ every step is a real move, except that the target, once reached,
repeats forever (and can never be left). (There is no general waiting move; this is what makes
the parity argument below available.) The formula is the chain
\[
  \Phi \;=\; \bigwedge_{i=0}^{L-1} T(X_i, X_{i+1}),
\]
endpoint pinning being absorbed into the domains below.

\paragraph{From the relation to signed clauses.}
$T$ is not a primitive of the constraint language; it is unrolled value
by value. Writing $[S]{:}X$ for the signed literal ``$X\in S$'', each
conjunct $T(X_i,X_{i+1})$ becomes one clause per value $u\in D_i$,
\[
  [\,D_i\setminus\{u\}\,]{:}X_i \;\vee\;
  [\,\mathrm{succ}(u)\cap D_{i+1}\,]{:}X_{i+1},
  \qquad
  \mathrm{succ}(u)=\begin{cases}
    N(u) & u\neq t,\\
    \{t\} & u=t,
  \end{cases}
\]
i.e.\ the implication $X_i{=}u \Rightarrow X_{i+1}\in\mathrm{succ}(u)$
written as a disjunction. The edge set $E$ never appears as an object:
adjacency ($N(u)$ = the up/down/left/right cells of $u$) is materialized
inside the enumerated successor lists. A guard $[\emptyset]{:}X_i$
(empty when $D_i=\{u\}$ is a singleton) is dropped, leaving a unary
clause. For the trimmed $2\times 2$ instance the whole formula is four
clauses:
\[
  [1,2]{:}X_1, \qquad
  [2]{:}X_1 \vee [3]{:}X_2, \qquad
  [1]{:}X_1 \vee [3]{:}X_2, \qquad
  [3]{:}X_3,
\]
(step $0$: $D_0=\{0\}$ forces $X_1\in N(0)\cap D_1=\{1,2\}$; step $1$:
from $u=1$ or $u=2$ the only admissible successor in $D_2=\{3\}$ is the
target; step $2$: absorption). Optionally,
\textsc{AllDifferent} clauses over the non-target positions restrict the
models to simple paths; we omit them (\texttt{allow\_cycles}) because
they connect every pair of step variables and destroy the chain
structure — simplicity is instead recovered downstream (loop erasure) or
left to the learner.

\paragraph{Domain trimming.}
The untrimmed encoding gives every $X_i$ the full domain $V$. The
trimmed encoding declares instead
\[
  D_i \;=\;
  \bigl\{\, v \in V\setminus\{t\} \;:\;
      \underbrace{d(s,v)\le i}_{\text{forward}},\;
      \underbrace{i \equiv d(s,v) \!\!\pmod 2}_{\text{parity}},\;
      \underbrace{d(v,t)\le L-i}_{\text{backward}} \,\bigr\}
  \;\cup\;
  \bigl\{\, t \;:\; i \ge d(s,t) \,\bigr\}.
\]
In particular $D_0=\{s\}$ and $D_L=\{t\}$: the endpoint constraints are
domain facts, not clauses.

\paragraph{Reading the three tests.}
Since $D_L=\{t\}$, the walk must have \emph{arrived by the deadline}
$L$; at time $i$ it therefore has exactly $L-i$ transitions left. The
three tests bound where it can possibly stand at that moment:
\begin{itemize}
\item \emph{Forward} ($d(s,v)\le i$): the past constrains the present —
  in $i$ moves from $s$ no walk outruns the graph distance, so cells
  farther than $i$ from the source are unoccupiable.
\item \emph{Backward} ($d(v,t)\le L-i$): the future constrains the
  present — from $v$ even the most direct route to $t$ costs $d(v,t)$
  moves, so a cell with $d(v,t)>L-i$ admits \emph{no} completion by the
  deadline, whatever the walk does afterwards. (At $i=L$ this test alone
  yields $D_L=\{t\}$; one step earlier only $t$ and its neighbours
  survive.)
\item \emph{Parity} ($i\equiv d(s,v) \bmod 2$): within reach is not
  enough — every move flips the checkerboard colour, so the step count
  and the distance must also agree modulo $2$.
\end{itemize}
The forward and backward tests are the two cones of a discrete
space--time ``light cone'' — reachable-from-$s$ and
can-still-reach-$t$ — and $D_i$ is their intersection restricted to the
correct colour class. Domains are accordingly narrow near both ends of
the horizon and widest in the middle. Deleting a cell that fails any
test removes no walk, since every walk satisfies all three by the
prefix/suffix argument of Proposition~\ref{prop:trim-exact}; the
backward family is also precisely what bucket elimination would
otherwise have to derive as resolvents (see below), while the forward
and parity families are stronger facts that the target-first elimination
would never produce.

\begin{proposition}[Exactness of the trimming]
\label{prop:trim-exact}
An assignment $(x_0,\dots,x_L)\in D_0\times\cdots\times D_L$ satisfies
$\Phi$ if and only if it is an absorbing $s$--$t$ walk: a four-neighbour
walk that starts at $s$, first reaches $t$ at some time
$\ell \le L$, and repeats $t$ thereafter. Moreover the trimming deletes
no such walk: the models of the trimmed formula and of the untrimmed
formula (with absorbing transitions and pinned endpoints) coincide.
Consequently \emph{every} simple $s$--$t$ path — of every length, the
maximum being $L=n-1$ — is a model, obtained by padding it with
repetitions of $t$; the remaining models are exactly the walks that
revisit a cell before reaching $t$.
\end{proposition}

\begin{proof}[Proof sketch]
($\Leftarrow$) Let $(x_i)$ be an absorbing walk with arrival time
$\ell$. For $i<\ell$ the prefix $x_0\cdots x_i$ is a genuine walk of
length $i$ from $s$, so $d(s,x_i)\le i$ and, by bipartiteness,
$i\equiv d(s,x_i) \pmod 2$; the suffix $x_i\cdots x_\ell$ reaches $t$ in
$\ell-i\le L-i$ moves, so $d(x_i,t)\le L-i$. For $i\ge\ell$, $x_i=t$ and
$i \ge \ell \ge d(s,t)$. Hence every position lies in its trimmed
domain, and consecutive positions satisfy $T$ by definition of a walk.
($\Rightarrow$) Conversely, a satisfying assignment moves along edges
until its first visit to $t$ (self-repetition is licensed only at $t$),
and $x_L=t$ because $D_L=\{t\}$; it is therefore an absorbing $s$--$t$
walk. Both directions never invoke the trimming tests negatively, which
proves the coincidence of model sets. Finally, a simple path padded at
$t$ is an absorbing walk, and an absorbing walk with pairwise-distinct
non-target positions is a padded simple path.
\end{proof}

\paragraph{Why this compiles with zero resolvents.}
The primal graph of $\Phi$ is the chain
$X_0 - X_1 - \cdots - X_L$ (induced width $1$), so bucket elimination
never couples more than two step variables. Its only derivations would
be the backward-reachability constraints
``$X_i$ must lie within distance $L-i$ of $t$'' propagated from the
pinned endpoint — precisely the \emph{backward} trimming test, which the
domains already encode; the forward and parity tests are stronger facts
that resolution would not derive at all. The trimmed input is therefore
its own compilation fixpoint. Measured compilation costs (single
worker, AMD EPYC 9335):

\begin{center}
\begin{tabular}{llrrrr}
\hline
grid & variant & clauses & resolvents & time & artifact \\
\hline
$12\times 12$ & untrimmed (waiting) & $20{,}448$ & $1{,}441$ & $38.2$\,s & $11$\,MB \\
$12\times 12$ & \textbf{trimmed}    & $8{,}839$  & $\mathbf{0}$ & $\mathbf{2.0}$\,\textbf{s} & $1.9$\,MB \\
$12\times 12$ & trimmed, dataset horizon $L{=}32$ & $864$ & $0$ & $0.13$\,s & $113$\,KB \\
$30\times 30$ & untrimmed (waiting) & $808{,}200$ & $25{,}201$ & $32.1$\,h$^{*}$ & $2.8$\,GB \\
$30\times 30$ & \textbf{trimmed}    & $379{,}315$ & $\mathbf{0}$ & $\mathbf{968}$\,\textbf{s} & $612$\,MB \\
$30\times 30$ & trimmed, dataset horizon $L{=}80$ & $10{,}806$ & $0$ & $9.1$\,s & $8.1$\,MB \\
\hline
\end{tabular}
\end{center}

\noindent
$^{*}$Of the $32.1$ hours, essentially all is spent in the
$2(g{-}1)=58$-bucket backward-reachability zone at $\approx 30$ minutes
per bucket; the remaining $842$ buckets produce no resolvents and pass
in a fraction of the time. The $25{,}201$ resolvents are precisely the
growing balls around the target, derived one family at a time in
width-$900$ literal arithmetic — exactly the information the backward
trimming test precomputes for free. The trimmed encoding
eliminates this zone entirely at every grid size (zero resolvents at
$12\times 12$ and $30\times 30$ alike), leaving compilation dominated by
parsing and subsumption bookkeeping.

\paragraph{Worked example: $2\times 2$.}
Cells row-major: $s=0=(0,0)$, $1=(0,1)$, $2=(1,0)$, $t=3=(1,1)$;
$d(s,t)=2$, $L=3$. The trimmed domains are
\[
  D_0=\{0\},\qquad
  D_1=\{1,2\},\qquad
  D_2=\{3\},\qquad
  D_3=\{3\}.
\]
(E.g.\ $0\notin D_1$ by parity; $0\notin D_2$ because
$d(0,t)=2>L-2=1$; $t\notin D_1$ because $1<d(s,t)$.)
The formula has exactly two models,
\[
  (0,1,3,3) \qquad\text{and}\qquad (0,2,3,3),
\]
i.e.\ precisely the two simple $s$--$t$ paths — at this size the walk
relaxation is already exact, and the model count equals the path count.

\paragraph{Worked example: $3\times 3$.}
Cells $0..8$ row-major, $s=0$, $t=8$, $d(s,t)=4$, $L=8$. Distances from
$s$ are $(0,1,2,1,2,3,2,3,4)$ and to $t$ are $(4,3,2,3,2,1,2,1,0)$.
The trimmed domains are
\[
\begin{array}{ll}
  D_0=\{0\} &
  D_1=\{1,3\}\\
  D_2=\{0,2,4,6\} &
  D_3=\{1,3,5,7\}\\
  D_4=\{0,2,4,6,8\} &
  D_5=\{1,3,5,7,8\}\\
  D_6=\{2,4,6,8\} &
  D_7=\{5,7,8\}\\
  D_8=\{8\}
\end{array}
\]
— $29$ values in total against $81$ untrimmed (the parity test alone
halves each interior domain; the backward test additionally removes
$0$ from $D_6$ and $1,3$ from $D_7$; $t$ enters only from step
$d(s,t)=4$ on). Sample models:
\begin{itemize}
\item a geodesic simple path, padded at the target:
      $(0,1,2,5,8,8,8,8,8)$;
\item a maximal simple path using the full horizon (no padding):
      $(0,1,2,5,4,3,6,7,8)$;
\item a cyclic walk retained by \texttt{allow\_cycles} — it revisits
      $0$ and $3$ before arriving:
      $(0,1,4,3,0,3,6,7,8)$; loop erasure maps it to $(0,3,6,7,8)$.
\end{itemize}
All three satisfy every domain membership and transition. Excluded by
construction are: assignments that leave the target after reaching it
(absorption), assignments placing any cell off its parity class (e.g.\
$x_2=1$), jumps between non-adjacent cells, and prefixes that stray too
far to return (e.g.\ $x_6=0$: $d(0,t)=4>L-6$).

\paragraph{Sequence-level supervision and head design.}
Feeding every step variable the \emph{same} network tile distribution
leaves the sequence dimension unsupervised: the greedy walk then has no
reason to progress, and empirically it oscillates inside the predicted
corridor and reaches $t$ only when the backward cone forces it (arrival
at the last admissible step for $100\%$ of test samples), leaving path
quality intact but the raw sequence degenerate. The remedy is to let the
network output \emph{one categorical distribution per step variable} and
to supervise it with the ground-truth optimal path \emph{as a sequence},
the absorbing target repeated after arrival; the training loss applies
the cross-entropy \emph{after} the projection (on the layer-adjusted
distributions), plus the mean-normalized $\ell_2$ projection
regularizer. This has one structural cost: a dense head needs one logit
per (step, admissible cell) pair, i.e.\ $\sum_i |D_i|$ outputs —
benign at $12\times 12$ ($6{,}234$ logits at the provable horizon
$L{=}106$) but $O(Ln)$ in general ($379$k logits, a $194$M-parameter
final layer, at $30\times 30$ with $L{=}899$). Two mitigations: (i)
shrink the horizon — either to the provable chordless bound of
Section~\ref{sec:trimmed-steps} or, more aggressively, to the longest
optimal path \emph{observed in the dataset} (at $12\times 12$: $L=32$
versus the provable $106$; the artifact shrinks to $869$ logits and
$864$ clauses); (ii) factorize the head as
$\mathrm{logit}(i,v) = \langle u_i, e_v(x)\rangle$ with per-cell
embeddings $e_v(x)$ taken from the stride-8 feature map (one feature
per grid tile) plus broadcast global context, and learned per-step
queries $u_i$ — $O((L{+}n)d)$ parameters in place of $O(Ln)$.
Empirically, sequence supervision removes the procrastination outright
(mean arrival drops from the horizon to the geodesic within two
epochs).
Measured at $12\times 12$ (CE after projection, $\lambda=1$
mean-normalized, paper recipe). We report the \emph{walk-cost} metric —
every entered tile is paid at every visit, so wandering is penalized
directly and a walk counts as optimal only if it is a wander-free
optimal simple path — alongside the loop-erased strict metrics.
``simple'' = fraction of emitted sequences with no revisit. Dense-head
rows are mean $\pm$ std over six seeds; factorized rows are single-seed:

\begin{center}
\begin{tabular}{llrrrrr}
\hline
head & $L$ & walk-opt.\% & simple & erased-opt.\% & exact & ms/sample \\
\hline
dense & 32  & $68.8\pm2.1$ & $97.9\%$ & $69.5\pm2.0$ & $0.599\pm0.024$ & $0.09$ \\
dense & 40  & $68.5\pm2.1$ & $97.5\%$ & $69.1\pm2.0$ & $0.608\pm0.044$ & $0.11$ \\
dense & 106 & $66.2\pm4.8$ & $98.3\%$ & $66.7\pm4.9$ & $0.579\pm0.059$ & $0.15$ \\
factorized & 106 & $58.1$ & --- & $58.1$ & $0.532$ & $0.30$ \\
factorized & 32  & $59.1$ & --- & $59.1$ & $0.497$ & $0.11$ \\
\hline
\end{tabular}
\end{center}

\noindent
Sequence supervision removes the procrastination outright: $97$--$99\%$
of emitted sequences are already simple, median arrival equals the
geodesic, and the wandering-penalized and loop-erased metrics coincide
to within one point — the layer outputs clean optimal paths with no
post-processing credit. Across horizons the differences are within
seed noise (single-seed comparisons at this task overstate contrasts by
up to $7$ points), so the \emph{dataset} horizon is preferred purely on
cost: the smallest head and the cheapest projection at equal accuracy. The factorized head trades roughly
$10$ points for its $O((L{+}n)d)$ parameter budget — the price of
input-independent step queries — making it the scalable but not yet
accuracy-matched option for larger grids.

\paragraph{Sequence supervision at $30\times 30$.}
The same protocol at $30\times 30$ (three seeds per configuration;
horizons: dataset maximum $L{=}80$, dataset maximum plus slack
$L{=}100$, provable chordless bound $L{=}674$):

\begin{center}
\begin{tabular}{llrrrrrr}
\hline
head & $L$ & walk-opt.\% & ratio med & simple & exact & params & s/epoch \\
\hline
dense & 80  & $7.0\pm0.6$ & $1.235$ & $94.1\%$ & $0.055$ & $16.7$M & $11$ \\
dense & 100 & $7.0\pm1.1$ & $1.263$ & $95.0\%$ & $0.053$ & $21.3$M & $13$ \\
dense & 674 & $6.6\pm0.2$ & $1.349$ & $98.4\%$ & $0.058$ & $154$M & $48$ \\
factorized & 80  & $0.0\pm0.0$ & $1.539$ & $0.0\%$ & $0.046$ & $11.2$M & $15$ \\
factorized & 100 & $0.0\pm0.0$ & $2.485$ & $0.0\%$ & $0.004$ & $11.2$M & $21$ \\
\hline
\end{tabular}
\end{center}

\noindent
Three conclusions. \emph{First}, the dense head scales: even the
$154$M-parameter output layer at the chordless bound trains stably
($48$\,s/epoch, $2$\,ms projections), and accuracy is horizon-invariant
— the dataset horizon again wins on cost alone. \emph{Second}, the
factorized head fails outright at this scale ($0\%$ wander-free
optimal, no simple sequences): input-independent step queries cannot
drive $900$-way step decisions, so the affordable head is the one that
does not work, while the ``expensive'' dense head is entirely
affordable. \emph{Third}, the $\approx 7\%$ walk-opt plateau, shared by the greedy oracle walks and every horizon, is a \emph{perception} ceiling, not a decoding one: an
\emph{exact} dynamic-programming decode over the down/right-monotone
DAG on the same network's probabilities attains only $6.3\%$ (median
ratio $1.23$), despite searching its restricted space optimally, while
the masks contain a true-weight optimal path $38\%$ of the time.
The network's coarse corridor is right; its fine cost ranking between
neighbouring corridors is not, and no decoder over the same
probabilities — restricted or not — can recover information the scores
do not carry. Improving the $30\times 30$ operating point is therefore
a backbone problem (e.g.\ cost regression in place of membership
probabilities), orthogonal to the layer's guarantees.

\section{Broader Impacts}\label{app:broader_impacts}

The guarantees provided by \layer{} are highly relevant in safety-critical
domains, such as clinical decision support tools. However, while \layer{}
guarantees that its outputs satisfy the given constraints, it does not
guarantee that they are correct. First, the constraints themselves may contain
errors, which \layer{} would enforce just as reliably as correct knowledge.
Second, a prediction can satisfy every constraint and still be wrong; since
all outputs appear valid, such constraint-compliant errors may go unnoticed.

\clearpage
\input{checklist.tex}

\end{document}

%% file: checklist.tex
\section*{NeurIPS Paper Checklist}

\begin{enumerate}

\item {\bf Claims}
    \item[] Question: Do the main claims made in the abstract and introduction accurately reflect the paper's contributions and scope?
    \item[] Answer: \answerYes{} % Replace by \answerYes{}, \answerNo{}, or \answerNA{}.
    \item[] Justification: The abstract and introduction claim CASPER (i) guarantees constraint
    satisfaction by construction, (ii) scales at inference time, (iii) requires no problem-specific
    decomposition and (iv) is fully differentiable. These properties are substantiated by
    Theorem~\ref{thm:correctness} (correctness) and verified by the experimental analysis detailed in Section 4.

\item {\bf Limitations}
    \item[] Question: Does the paper discuss the limitations of the work performed by the authors?
    \item[] Answer: \answerYes{} % Replace by \answerYes{}, \answerNo{}, or \answerNA{}.
    \item[] Justification: In Appendix A we discuss the limitations of CASPER namely we consider the problem of enforcing hard constraints over categorical variables. We do not address several orthogonal problems to ours: constraint satisfaction over continuous variables, reasoning shortcuts, modelling dependencies between predicted symbols and concept base interpretability. 

\item {\bf Theory assumptions and proofs}
    \item[] Question: For each theoretical result, does the paper provide the full set of assumptions and a complete (and correct) proof?
    \item[] Answer: \answerYes{} % Replace by \answerYes{}, \answerNo{}, or \answerNA{}.
    \item[] Justification: All theoretical results are stated with explicit assumptions in Section 3 and
    proven in Appendices B–M. Theorems 1–2 and Lemma 1 establish the single-variable
    projection and its satisfaction guarantee; Theorem~\ref{thm:argmax} shows that the surrogate and exact projections select the same value; Theorem~\ref{thm:exact_error}, Corollary 1, and Propositions 1–2 characterise the surrogate error and gradient behaviour; Lemma 2 and Theorem~\ref{thm:correctness} extend correctness to the multi-variable case via the compilation step, whose cost is bounded by Theorem 5; and Corollary~\ref{cor:snowball_effect} shows that the surrogate does not alter the sequence of assignments.
    Each result is numbered and cross-referenced from the main text.

    \item {\bf Experimental result reproducibility}
    \item[] Question: Does the paper fully disclose all the information needed to reproduce the main experimental results of the paper to the extent that it affects the main claims and/or conclusions of the paper (regardless of whether the code and data are provided or not)?
    \item[] Answer: \answerYes{} % Replace by \answerYes{}, \answerNo{}, or \answerNA{}.
    \item[] Justification: We provide a detailed description of our method in Section~\ref{sec:casper}, our key experimental results in Section~\ref{sec:experiments} and all remaining details (task-formulation, experimental setup and implementation details) specific to each task in Appendix ~\ref{app:full-results}.

\item {\bf Open access to data and code}
    \item[] Question: Does the paper provide open access to the data and code, with sufficient instructions to faithfully reproduce the main experimental results, as described in supplemental material?
    \item[] Answer: \answerYes % Replace by \answerYes{}, \answerNo{}, or \answerNA{}.
    \item[] Justification: Code and instructions to reproduce all experimental results are publicly available at \url{https://github.com/lkonathala25/casper}.

\item {\bf Experimental setting/details}
    \item[] Question: Does the paper specify all the training and test details (e.g., data splits, hyperparameters, how they were chosen, type of optimizer) necessary to understand the results?
    \item[] Answer: \answerYes{} % Replace by \answerYes{}, \answerNo{}, or \answerNA{}.
    \item[] Justification: All relevant details for our experimental analysis are covered, for each task, in Section~\ref{sec:experiments}.

\item {\bf Experiment statistical significance}
    \item[] Question: Does the paper report error bars suitably and correctly defined or other appropriate information about the statistical significance of the experiments?
    \item[] Answer: \answerYes{} % Replace by \answerYes{}, \answerNo{}, or \answerNA{}.
    \item[] Justification: We report the error bars in all the tables and charts present in our paper. 

\item {\bf Experiments compute resources}
    \item[] Question: For each experiment, does the paper provide sufficient information on the computer resources (type of compute workers, memory, time of execution) needed to reproduce the experiments?
    \item[] Answer: \answerYes{} % Replace by \answerYes{}, \answerNo{}, or \answerNA{}.
    \item[] Justification: We cover the compute resources used in our experiment analysis in Section~~\ref{sec:experiments}.
    
\item {\bf Code of ethics}
    \item[] Question: Does the research conducted in the paper conform, in every respect, with the NeurIPS Code of Ethics \url{https://neurips.cc/public/EthicsGuidelines}?
    \item[] Answer: \answerYes{} % Replace by \answerYes{}, \answerNo{}, or \answerNA{}.
    \item[] Justification: We confirm we have read the NeurIPS code of ethic and we comply with it.

\item {\bf Broader impacts}
    \item[] Question: Does the paper discuss both potential positive societal impacts and negative societal impacts of the work performed?
    \item[] Answer: \answerYes{} % Replace by \answerYes{}, \answerNo{}, or \answerNA{}.
    \item[] Justification: We discuss the potential positive and negative
    societal impacts of \layer{}, particularly in safety-critical domains such as clinical decision support, in
    Appendix~\ref{app:broader_impacts}.
    
\item {\bf Safeguards}
    \item[] Question: Does the paper describe safeguards that have been put in place for responsible release of data or models that have a high risk for misuse (e.g., pre-trained language models, image generators, or scraped datasets)?
    \item[] Answer: \answerNA{} % Replace by \answerYes{}, \answerNo{}, or \answerNA{}.
    \item[] Justification: This paper proposes a method that does not present any particular risk of misuse.

\item {\bf Licenses for existing assets}
    \item[] Question: Are the creators or original owners of assets (e.g., code, data, models), used in the paper, properly credited and are the license and terms of use explicitly mentioned and properly respected?
    \item[] Answer: \answerYes{} % Replace by \answerYes{}, \answerNo{}, or \answerNA{}.
    \item[] Justification: All datasets used in the experimental analysis are cited: MNIST-Arithmetic
    [32], Warcraft Pathfinding [40] and SatNet (Sudoku Solving) [49] where the associated licenses are outlined in Appendix N. 

\item {\bf New assets}
    \item[] Question: Are new assets introduced in the paper well documented and is the documentation provided alongside the assets?
    \item[] Answer: \answerNA{} % Replace by \answerYes{}, \answerNo{}, or \answerNA{}.
    \item[] Justification: This paper does not release any new assets.

\item {\bf Crowdsourcing and research with human subjects}
    \item[] Question: For crowdsourcing experiments and research with human subjects, does the paper include the full text of instructions given to participants and screenshots, if applicable, as well as details about compensation (if any)? 
    \item[] Answer: \answerNA{} % Replace by \answerYes{}, \answerNo{}, or \answerNA{}.
    \item[] Justification: This paper  does not involve crowdsourcing nor research with human subjects

\item {\bf Institutional review board (IRB) approvals or equivalent for research with human subjects}
    \item[] Question: Does the paper describe potential risks incurred by study participants, whether such risks were disclosed to the subjects, and whether Institutional Review Board (IRB) approvals (or an equivalent approval/review based on the requirements of your country or institution) were obtained?
    \item[] Answer: \answerNA{} % Replace by \answerYes{}, \answerNo{}, or \answerNA{}.
    \item[] Justification: This paper does not involve crowdsourcing nor research with human subjects.

\item {\bf Declaration of LLM usage}
    \item[] Question: Does the paper describe the usage of LLMs if it is an important, original, or non-standard component of the core methods in this research? Note that if the LLM is used only for writing, editing, or formatting purposes and does \emph{not} impact the core methodology, scientific rigor, or originality of the research, declaration is not required.
    %this research? 
    \item[] Answer: \answerNA{} % Replace by \answerYes{}, \answerNo{}, or \answerNA{}.
    \item[] Justification: The development of the core methodology of this paper does not involve LLMs as any important, original, or non-standard component. 

\end{enumerate}